%% file: main.tex
\documentclass[11pt, a4paper, copyright]{google}

\usepackage{tikz}
\usepackage{multicol}
\usetikzlibrary{positioning,arrows.meta,shapes.geometric,fit,backgrounds,calc,decorations.pathreplacing}

\usepackage[authoryear, sort&compress, round]{natbib}
\makeatletter
\renewcommand{\absfont}{\normalfont\linespread{1.2}\fontsize{11}{12}\selectfont}

\definecolor{princetonorange}{HTML}{2E6DA4}
\renewcommand{\abscontent}{%
  \noindent
  \parbox{\dimexpr\linewidth}{\absfont \theabstract}%
  \@ifundefined{@keywords}{}{%
    \vskip0.8em \noindent \keywordsfont Keywords: \@keywords}%
}
\makeatother

\definecolor{linkblue}{RGB}{0,0,139}
\definecolor{navy}{RGB}{0,0,128}
\definecolor{royalblue}{RGB}{65,105,225}
\definecolor{steelblue}{RGB}{70,130,180}
\definecolor{dodgerblue}{RGB}{30,144,255}
\definecolor{mediumblue}{RGB}{0,0,205}
\definecolor{darkslateblue}{RGB}{72,61,139}
\definecolor{lightbluelink}{HTML}{2E6DA4}
\usepackage[colorlinks = true,
            linkcolor = linkblue,
            urlcolor  = lightbluelink,
            citecolor = lightbluelink,
            anchorcolor = blue]{hyperref}

\usepackage[misc]{ifsym}
\usepackage{minitoc}
\usepackage{cleveref}
\usepackage{subcaption}
\usepackage{booktabs}
\usepackage{amsmath}
\usepackage{mathtools}
\usepackage{amssymb}
\usepackage{graphicx}
\usepackage{algorithmic}
\usepackage{algorithm}
\usepackage{tcolorbox}
\usepackage{verbatim}
\usepackage{makecell}
\usepackage{array}
\usepackage{multirow}
\usepackage{xurl}
\usepackage{amsthm}
\usepackage{adjustbox}
\usepackage{caption}
\usepackage{xcolor}
\usepackage{wrapfig}
\usepackage{pifont}
\usepackage{enumitem}
\usepackage{xspace}
\usepackage[titles]{tocloft}
\definecolor{tocsubsec}{HTML}{444444}
\definecolor{tocsubsubsec}{HTML}{666666}

\usepackage{soul}

\usepackage{listings}
\tcbuselibrary{listings,skins,breakable}

\lstdefinestyle{prompt}{
  basicstyle=\ttfamily\footnotesize,
  breaklines=true,
  breakatwhitespace=true,
  columns=fullflexible,
  keepspaces=true,
  showstringspaces=false,
  postbreak=\mbox{\textcolor{gray}{$\hookrightarrow$}\space}
}

\newcommand{\ours}{\textsl{ThoughtTrace}\xspace}

\newcommand{\foo}{\hspace{-2.3pt}$\bullet$ \hspace{5pt}}

\definecolor{accent1}{HTML}{2C5F8D}  
\definecolor{accent2}{HTML}{B8423A}  
\definecolor{accent3}{HTML}{4A7C59}  
\definecolor{accent4}{HTML}{6B4C8A}  

\definecolor{lightblue}{HTML}{DCE7F1}    
\definecolor{lightorange}{HTML}{F5E0CC}  
\definecolor{lightgreen}{HTML}{DCE8DD}   
\definecolor{lightyellow}{HTML}{FAF3D9}  
\definecolor{lightgray}{HTML}{E8E8E8}    
\definecolor{lightpurple}{HTML}{E4DAEB}  
\definecolor{lightred}{HTML}{F2D9D6}     

\definecolor{gray60}{gray}{0.6}

\definecolor{metabg}{HTML}{F1F4F7}
\tcbset{
  aibox/.style={
    width=\linewidth,
    top=7pt,
    bottom=5pt,
    colback=metabg,
    colframe=black,
    colbacktitle=black,
    enhanced,
    center,
    attach boxed title to top left={yshift=-0.1in,xshift=0.15in},
    boxed title style={boxrule=0pt,colframe=white,},
    breakable
  }
}
\newtcolorbox{AIbox}[2][]{aibox,title=#2,#1}

\newcommand{\dashedrule}{%
  \par\vspace{1pt}%
  \noindent\tikz\draw[dashed, gray, line width=0.5pt] (0,0) -- (\linewidth,0);%
  \par\vspace{2pt}%
}

\newcommand{\qwenicon}{%
  \smash{\raisebox{-0.3em}{\includegraphics[height=1.2em]{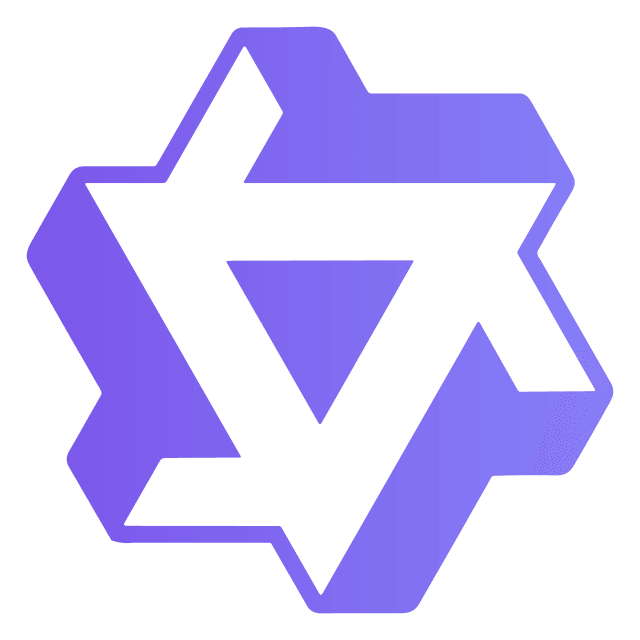}}}
}
\newcommand{\claudeicon}{%
  \smash{\raisebox{-0.3em}{\includegraphics[height=1.2em]{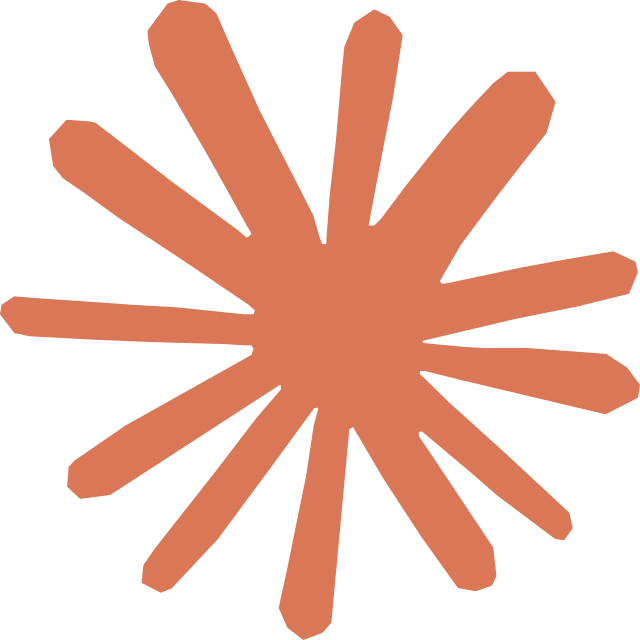}}}
}
\newcommand{\gpticon}{%
  \smash{\raisebox{-0.3em}{\includegraphics[height=1.2em]{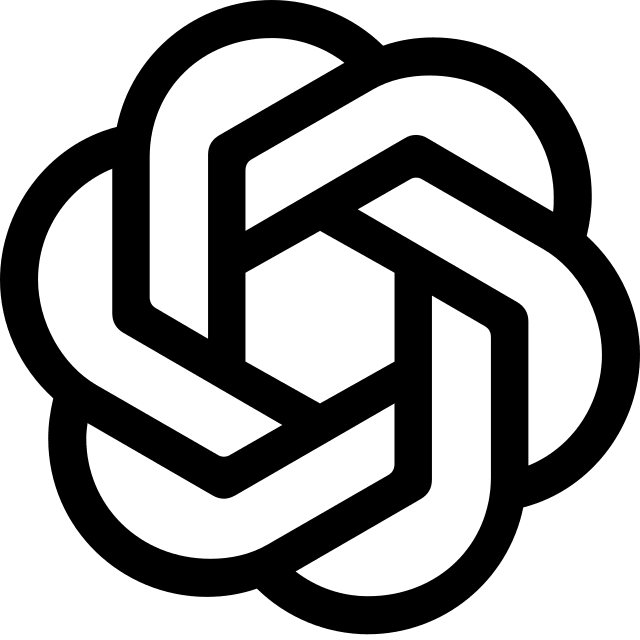}}}
}
\newcommand{\geminiicon}{%
  \smash{\raisebox{-0.3em}{\includegraphics[height=1.2em]{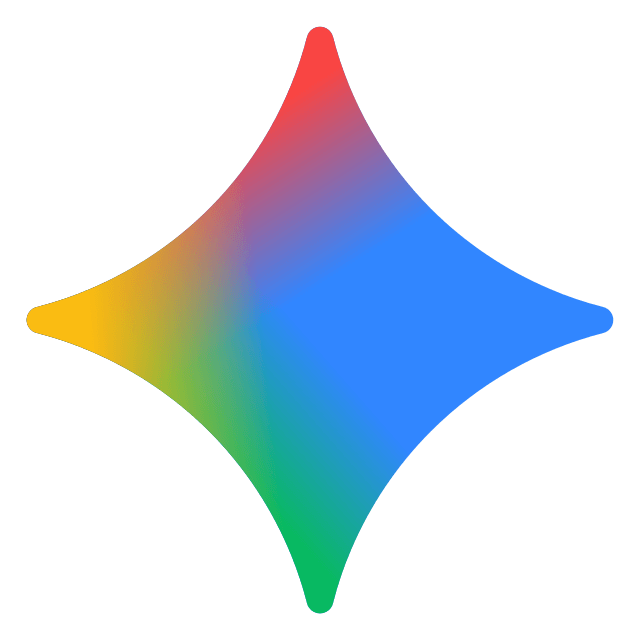}}}
}
\newcommand{\grokicon}{%
  \smash{\raisebox{-0.3em}{\includegraphics[height=1.2em]{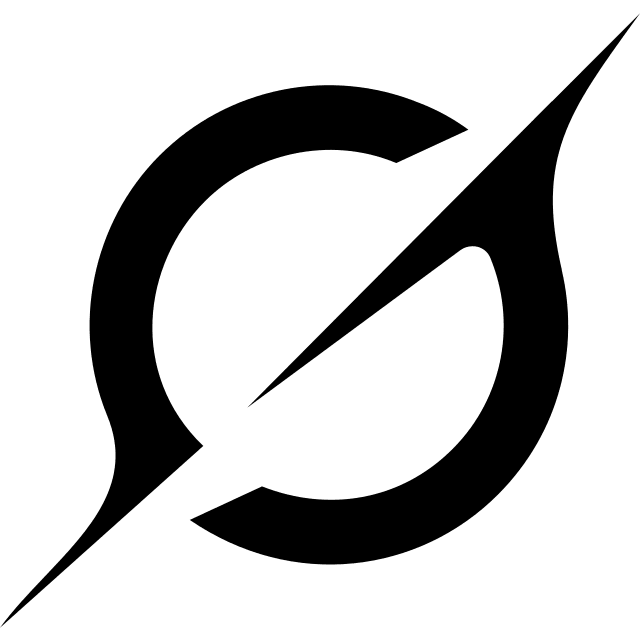}}}
}
\newcommand{\minimaxicon}{%
  \smash{\raisebox{-0.3em}{\includegraphics[height=1.2em]{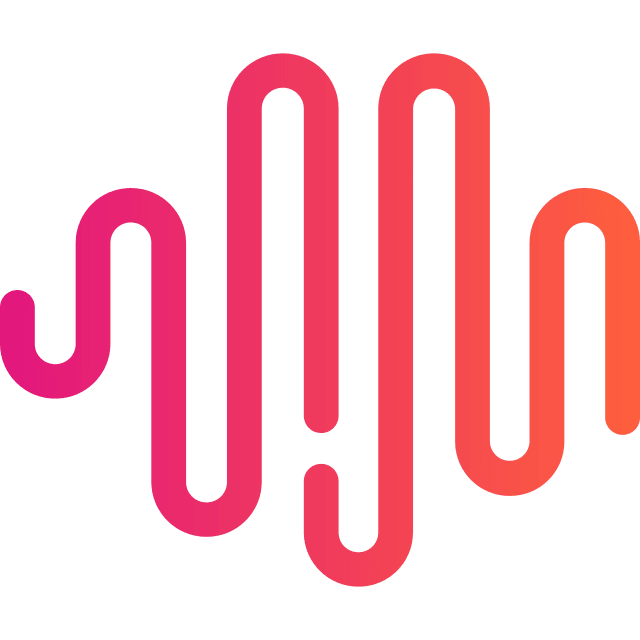}}}
}
\newcommand{\kimiicon}{%
  \smash{\raisebox{-0.3em}{\includegraphics[height=1.2em]{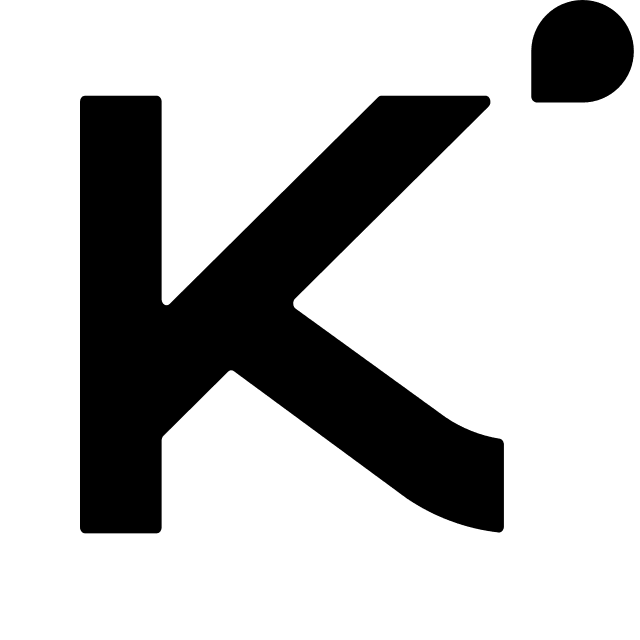}}}
}
\newcommand{\mimoicon}{%
  \smash{\raisebox{-0.3em}{\includegraphics[height=1.2em]{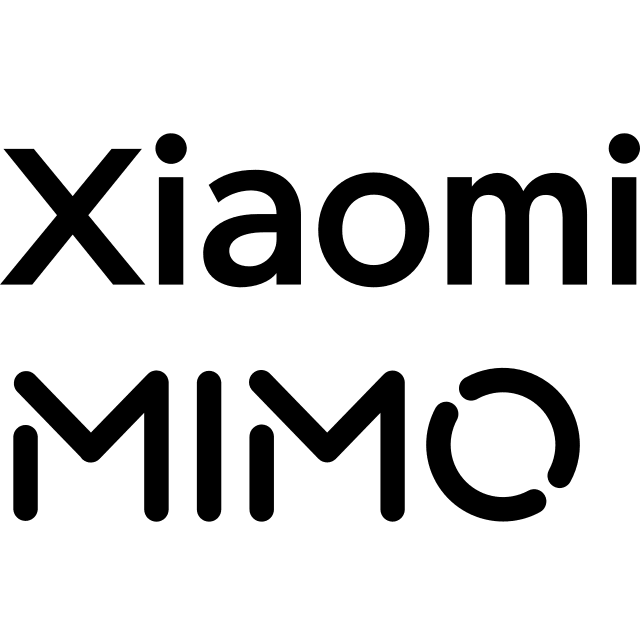}}}
}
\newcommand{\zaiicon}{%
  \smash{\raisebox{-0.3em}{\includegraphics[height=1.2em]{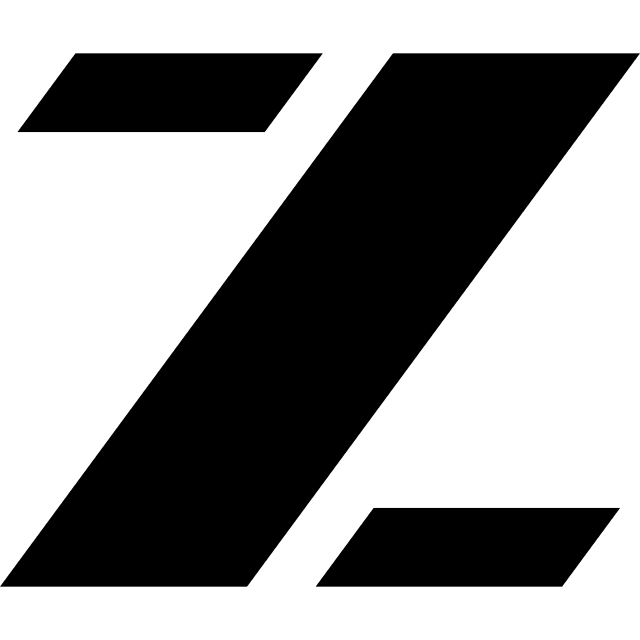}}}
}
\newcommand{\deepseekicon}{%
  \smash{\raisebox{-0.3em}{\includegraphics[height=1.2em]{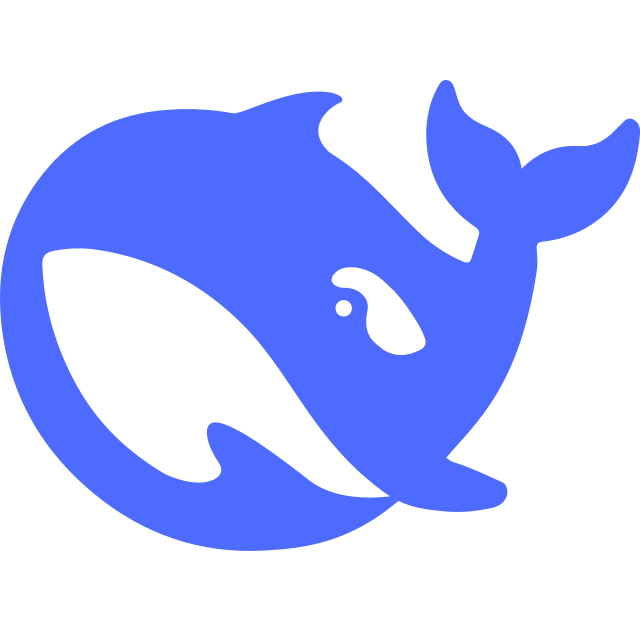}}}
}
\newcommand{\metaicon}{%
  \smash{\raisebox{-0.3em}{\includegraphics[height=1.2em]{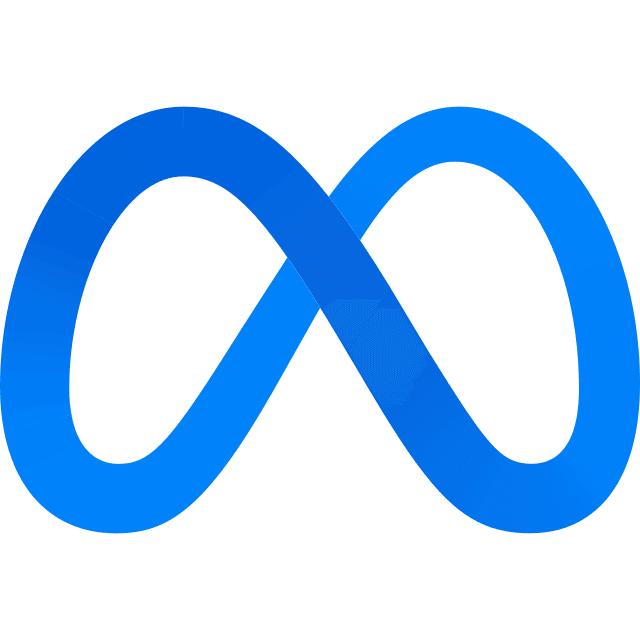}}}
}
\newcommand{\mistralicon}{%
  \smash{\raisebox{-0.3em}{\includegraphics[height=1.2em]{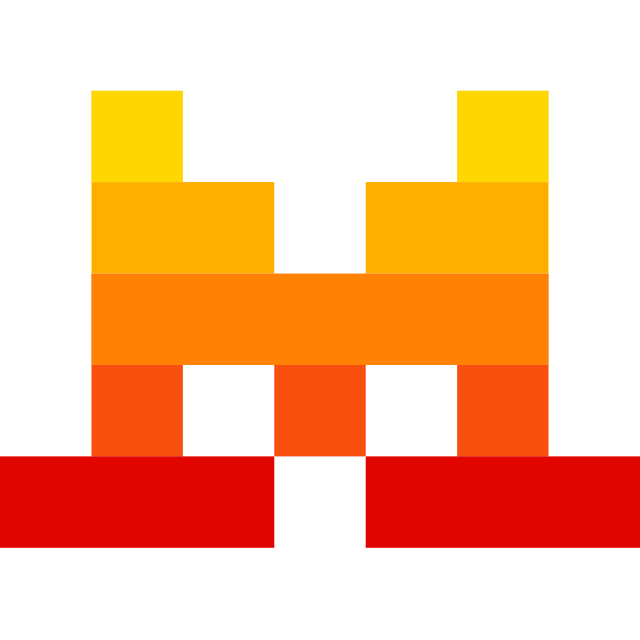}}}
}
\newcommand{\stepfunicon}{%
  \smash{\raisebox{-0.3em}{\includegraphics[height=1.2em]{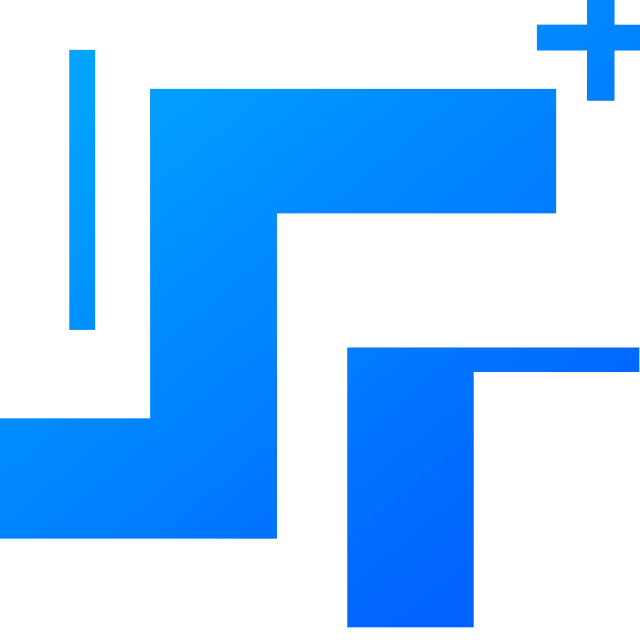}}}
}

\uselogo{}

\makeatletter
\newcommand{\resourceblock}{%
  \begin{center}
  \href{https://thoughttrace-project.github.io/}{\texttt{https://thoughttrace-project.github.io/}}
  \end{center}
}
\renewcommand{\maketitle}{\bgroup\setlength{\parindent}{0pt}
  \begin{adjustwidth}{0pt}{24pt}
    \begin{center}
      {\titlefont \ours: Understanding User Thoughts\\ in Real-World LLM Interactions\par}%
      \vskip14pt
      {\@author\par}%
      \vskip18pt%
    \end{center}
  \end{adjustwidth}
  \egroup
  \resourceblock
  \vskip10pt
  {\abscontent}%
  \thispagestyle{firststyle}
}
\makeatother

\AtBeginDocument{%
  \fancypagestyle{firststyle}{%
    \fancyhead[L]{\includegraphics[height=24pt]{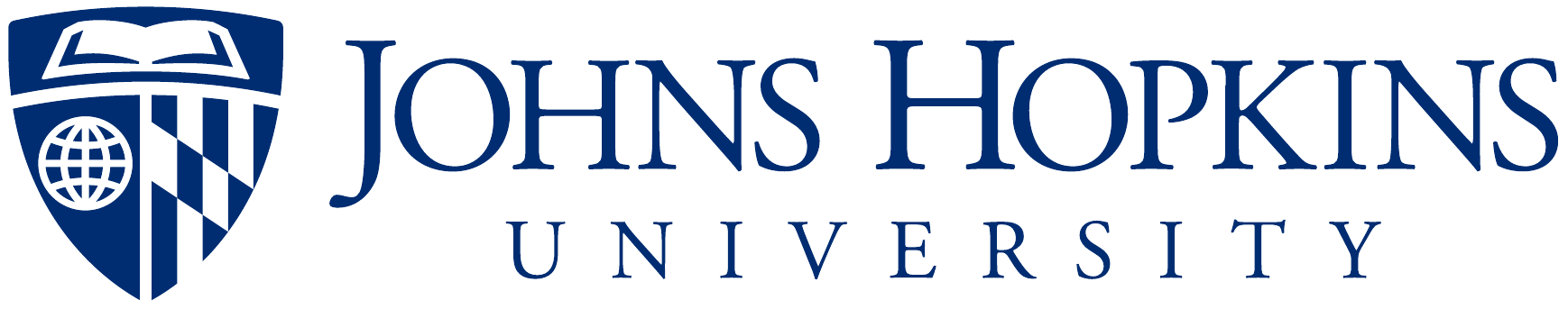}}%
    \fancyhead[R]{\includegraphics[height=24pt]{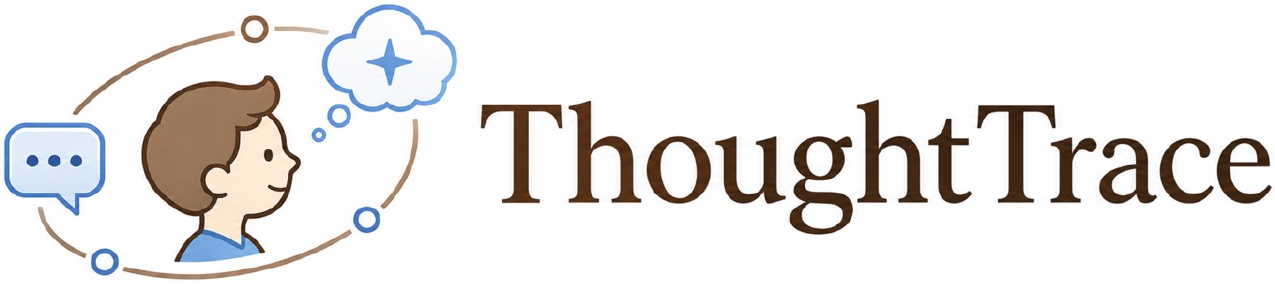}}
    \fancyhead[C]{}%
    \fancyfoot[L]{%
      \footerfont
      \ifdefined\correspondingauthor
        \if\relax\the\correspondingauthor\relax
        \else
          {\itshape \the\correspondingauthor}%
        \fi
      \fi
      \ifthenelse{\boolean{copyright}}{\copyrightext}{}%
    }%
    \fancyfoot[R]{\footerfont\bfseries\relax}%
    \fancyfoot[C]{\footerfont\bfseries\relax}%
  }%
}

\title{\ours: Understanding User Thoughts in Real-World LLM Interactions}

\correspondingauthor{Main contact: \{cjin33, tianmin.shu\}@jhu.edu}

\makeatletter
\renewcommand\AB@authnote[1]{}
\renewcommand\AB@affilnote[1]{}
\makeatother

\author{%
  Chuanyang Jin\textsuperscript{1},\;
  Binze Li\textsuperscript{1},\;
  Haopeng Xie\textsuperscript{1},\;
  Cathy Mengying Fang\textsuperscript{2},\;
  Tianjian Li\textsuperscript{1},\\
  Shayne Longpre\textsuperscript{2},\;
  Hongxiang Gu\textsuperscript{3},\;
  Maximillian Chen\textsuperscript{3},\;
  Tianmin Shu\textsuperscript{1}%
}

\affil{%
  \vskip2pt
  $^{1}$Johns Hopkins University \quad
  $^{2}$Massachusetts Institute of Technology \quad
  $^{3}$Google Research
}

\begin{abstract}
Conversational AI has now reached billions of users, yet existing datasets capture only what people \textit{say}, not what they \textit{think}. We introduce \textbf{\ours}, the first large-scale dataset that pairs real-world multi-turn human--AI conversations with users' self-reported thoughts: their \textit{reasons} for sending prompts and \textit{reactions} to assistant responses. \ours comprises 1,058 users, 2,155 conversations, 17,058 turns, and \textbf{10,174 thought annotations} collected across 20 language models. Our analysis shows that \ours captures long-horizon, topically diverse interactions, and that thoughts are semantically distinct from messages, difficult for frontier LLMs to infer from context, diverse in content, and tied to conversation stages.
We further demonstrate the utility of thoughts for downstream modeling.
First, thoughts improve \textbf{user-behavior prediction} as inference-time context.
Second, thought-guided rewrites provide \textbf{fine-grained alignment signals} for training personalized assistants. Together, \ours establishes user thoughts as a new data modality for studying the cognitive dynamics behind human--AI interaction and provides a foundation for building assistants that better understand and adapt to users' latent goals, preferences, and needs.
\end{abstract}

\begin{document}

\maketitle

\begin{figure}[h]
  \vspace{1em}
  \centering
  \includegraphics[width=\linewidth]{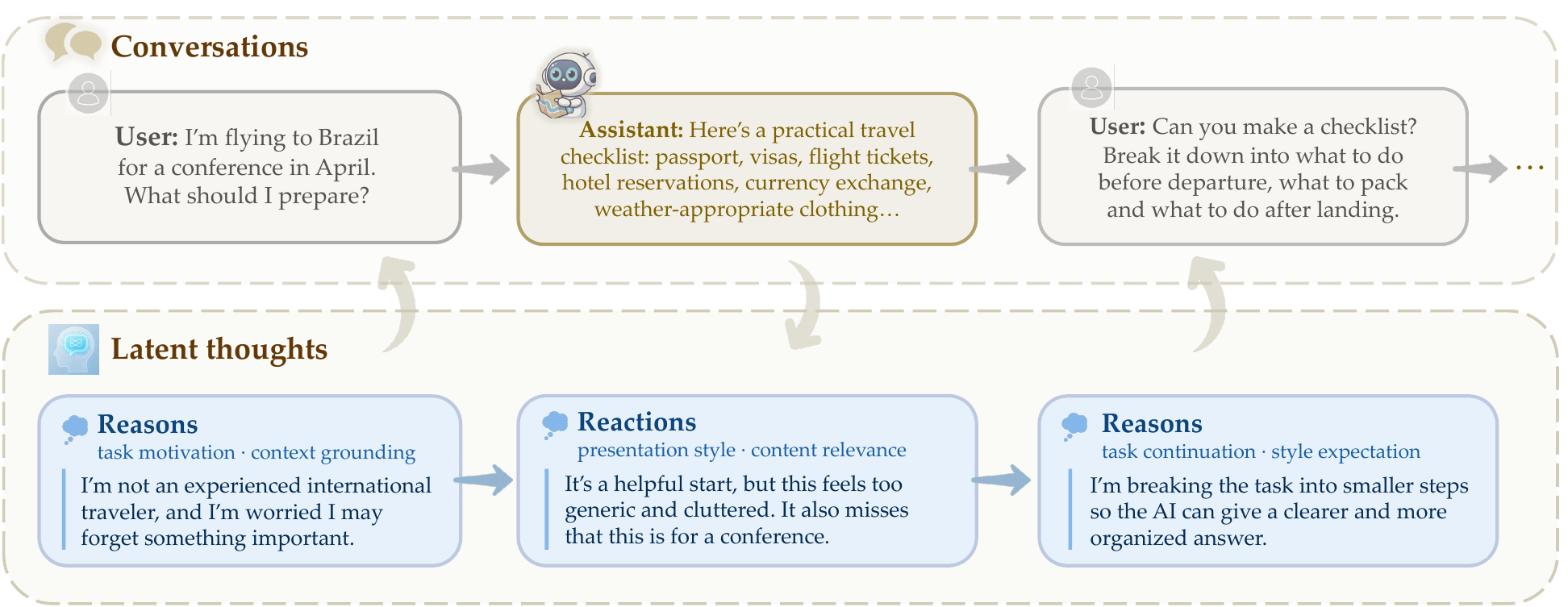}
  \caption{\textbf{A representative example from \ours.} A user interacts with a chatbot to complete daily tasks through multi-turn conversations (top), while annotating their latent thoughts during the conversations (bottom). Thoughts take two forms: \emph{reasons} for sending user prompts and \emph{reactions} to assistant responses, which can be categorized into several types (e.g., \emph{task motivation}, \emph{style expectation}). Latent thoughts reveal users' thought traces that drive the human-AI interactions in multi-turn conversations, providing valuable signals for user modeling and improving AI assistance.}
  \label{fig:overview}
\end{figure}

\newpage
\vspace{0.5em}
{
  \hypersetup{linkcolor=black}
  \setlength{\parskip}{0pt}
  \renewcommand{\contentsname}{\normalfont\large\bfseries Contents}
  \setcounter{tocdepth}{3}
  \begingroup
    \small
    \tableofcontents
  \endgroup
}

\newpage

\input{sections/1-intro}
\input{sections/2-related}
\input{sections/3-collection}
\input{sections/4-conv}
\input{sections/5-thought-properties}
\input{sections/6-thought-utility}
\input{sections/7-conclusion}

\newpage
\bibliography{paper}

\newpage
\appendix

\setcounter{figure}{0}
\setcounter{table}{0}
\setcounter{equation}{0}
\renewcommand{\thetable}{A\arabic{table}}
\renewcommand{\thefigure}{A\arabic{figure}}
\renewcommand{\theequation}{A\arabic{equation}}
\input{appendices/1-models}
\input{appendices/2-results}
\input{appendices/3-collection}
\input{appendices/4-analysis}

\end{document}

%% file: sections/1-intro.tex
\section{Introduction}
\label{sec:intro}

Conversational AI systems have now been deployed at an unprecedented scale, processing billions of user interactions every day. While extensive work focuses on what users \textbf{say} during these interactions \citep{zheng2023lmsys, zhao2024wildchat, baumann2026swe,jin2025era, shi2024wildfeedback}, understanding what users actually \textbf{think} during the conversations remains a largely unexplored dimension of human-AI interaction.


\textit{User thoughts} are the unspoken cognitive context behind each message: the motivation and goal driving the request, the context and constraints grounding it, the content or style expectations for the response, and the interpretations and reactions to the assistant's reply. Figure~\ref{fig:overview} illustrates why this hidden layer matters. The observed initial user message about preparing for a trip reads as a generic travel query, but unobservable \textit{thought} exposes the anxiety of an inexperienced international traveler. After the assistant replies with a standard checklist, the user's thought reveals dissatisfaction that the next message never explicitly states: the response feels generic and overlooks the conference context. The user's follow-up message operationalizes this private reaction by requesting a structured breakdown. Capturing these thoughts and their dynamics closes the gap between observable utterances and hidden user intents, providing richer signals for training and evaluation.

We introduce \ours, the first framework and dataset for understanding user thoughts during real-world human-AI interactions at scale. By asking users to engage in natural conversations while articulating contextually grounded thoughts, we collect a rich corpus of first-person cognitive traces that illuminate the lived experience of interacting with AI systems.

\ours features \textbf{high-quality}, \textbf{long-horizon} interactions grounded in open-ended \textbf{real-world tasks} performed by a diverse user base: 1,058 users, 2,155 timestamped conversations, 17,058 interaction turns, and \textbf{10,174 thought annotations}, collected via a chatbot service powered by 20 different language models. Each conversation includes: (1) naturalistic multi-turn dialogue between a user and an AI assistant; (2) user-reported thoughts aligned to individual user and assistant messages, including \textit{reasons} for sending messages and \textit{reactions} to assistant responses; (3) post-task descriptions of what users completed and what they expected from the AI; and (4) user demographic information such as age, gender, education level, and occupation.

Our analysis highlights the properties and utility of thoughts along three axes: (1) \textbf{Conversation properties} (Section~\ref{sec:conversations}): \ours{} features representative users, long-horizon conversations, broad topical coverage, and frequent extensions across turns. (2) \textbf{Thought properties} (Section~\ref{sec:thought_properties}): thoughts differ from messages, are difficult for frontier LLMs to infer, are diverse in content, and are tied to conversation stages. (3) \textbf{Thought utility} (Section~\ref{sec:thought_utility}): thoughts predict user behavior during inference (+41.7\% relative gain), and provide fine-grained alignment signals (+25.6\% win rate).


\ours opens several directions for future research. On \textbf{user modeling}, it enables systematic study of the dynamic human mental processes that arise in human--AI interaction: what users think during conversations, how conversational context shapes these thoughts, how thoughts subsequently shape user utterances, and how these dynamics vary across demographic groups. On \textbf{model training}, user thoughts provide a new supervisory signal that models can predict, learn from, and align with, offering a path toward assistants that better capture users' latent goals, expectations, and reactions. On \textbf{evaluation}, \ours enables benchmarks for thought prediction and supports thought-centered measures of user satisfaction, moving evaluation beyond surface-level utterances toward latent intent and subjective experience.


Our contributions are summarized as follows: (1) We introduce \emph{thoughts} as a new data modality for human-AI interaction research, and release \ours, a large-scale dataset pairing naturalistic multi-turn conversations with rich thought annotations and demographic metadata. (2) We characterize the conversational and cognitive structure of \ours along multiple axes, showing that thoughts are latent, hard to infer, diverse, and stage-dependent. (3) We demonstrate the utility of thoughts for predicting user behavior and aligning language models. Together, these contributions point toward assistants that learn from the full interaction experience---bridging observable dialogue with the internal cognition that drives it.

%% file: sections/2-related.tex
\section{Related Work}
\label{sec:related_work}

\textbf{Real-World Human-AI Conversations.} There have been recent datasets of real-world human-AI conversations, including general chat datasets such as WildChat \citep{zhao2024wildchat} and LMSYS-Chat-1M \citep{zheng2023lmsys} and domain-specific datasets such as SWE-Chat \citep{baumann2026swe} for software engineering. Additionally, PRISM \citep{kirk2024prism} paired conversation logs with sociodemographic surveys and stated preferences. Building on such corpora, recent works have developed methods to effectively extract supervisory signals such as satisfaction cues from natural conversations \citep{zhao2024wildhallucinations, shi2024wildfeedback, jin2025era, peng2026wildreward, buening2026aligning}. Across these efforts, the conversation transcript is treated as the primary unit of observation, and any view of the user is limited to what they explicitly verbalize; even PRISM elicits only ratings or stated preferences over outputs, not free-form annotations, leaving much of the user intents, evaluations, and thought processes behind their messages unobserved. \ours addresses this gap by pairing real conversations with underlying thought dynamics self-reported by the users.

\textbf{User Thoughts.} There has been an increasing interest in machine Theory of Mind (ToM) \cite{wimmer1983beliefs}, the ability to infer people's latent mental states from their behavior. However, much of the work focuses on structured Theory of Mind reasoning \citep{baker2009action, baker2017rational}, in which mental inferences are limited to a few well-defined mental variables, such as goals, beliefs, and desires, grounded in simple context \citep{ullman2023large, kim2023fantom, shapira2024clever, jin2024mmtom, shi2025muma, fan2025somi, sclar2023minding, zhang2025autotom, jha2024neural}. Thus, prior work fails to capture the dynamics of latent thoughts during interactions. While there has been recent research that explores how to leverage dynamic mental state inference to enhance AI assistance \citep{zhang2025autotom, zhou2025tom, zhang2026mindzero}, there has been a lack of systematic analysis and large-scale data collection of user thoughts in human-AI interactions. \ours aims to provide a new paradigm for collecting and analyzing user latent thoughts during multi-turn human-AI conversations.

\textbf{User Simulations.} There has been an increasing interest in building user simulators for training and evaluating AI assistants to address the data gap \citep{qian2025userrl, park2024generative, wu2026humanlm, binz2025foundation, naous2025flipping, kolluri2025finetuning, piao2025agentsociety, park2023generative, abdulhai2025consistently}. To do so, these works have heavily relied on prompting LLMs \citep{park2024generative, piao2025agentsociety} or finetuning LLMs on ground-truth responses or persona-consistent behavior \citep{binz2025foundation, naous2025flipping, kolluri2025finetuning, abdulhai2025consistently, mehri2025goal, zhu2025using}. However, recent works have found that existing simulators are biased and unfaithful \citep{zhou2026mind, seshadri2026lost}. While HumanLM \citep{wu2026humanlm} attempts to mitigate this by aligning simulated user conversations with users' internal states, its training still relies on synthetic user thoughts due to the lack of real thought data. The first-person thought traces from real users in real interactions in \ours may provide valuable data for training more realistic user simulators.

%% file: sections/3-collection.tex
\section{Data Collection}
\label{sec:collection}

\subsection{What are \textit{Thoughts}?}

\emph{Thoughts} refer to the users' latent cognitive context in human--AI conversations. Unlike users' observable utterances, which are often lossy representations of intent due to the principle of least effort \citep{zipf2016human}, thoughts capture the unspoken mental content that motivates those utterances. Because they are richer and faster-moving than verbalized language, conversations can transmit only a fraction of their content in real time. Conversational language is also shaped by pragmatic and utility-driven pressures: speakers produce utterances that are efficient, socially appropriate, and goal-directed, rather than fully transparent reflections of their internal mental states \citep{sperber1986relevance}. 

As shown in Figure~\ref{fig:overview}, in our data collection, thoughts are annotated as either \emph{reactions}, which reflect how users internally respond to an assistant message, or \emph{reasons}, which explain why users send a particular message. We collect both types at each turn because they jointly shape how users proceed in the next turn. Specifically, reactions indicate how users perceive the model, while reasons reveal how users want the model to understand their needs and preferences. Together, these thoughts drive the progression of the conversation and reveal the cognitive traces of users during interactions.

\subsection{Methodology}

We recruited participants via Prolific and redirected them to our data collection platform to complete trials following the procedure below. This study was approved by an institutional review board.

\textbf{Step 1: User consent.} Participants provided informed consent acknowledging voluntary participation, guaranteed anonymity, and the right to withdraw at any time.

\textbf{Step 2: Tutorial and quiz.} Participants first completed a guided tutorial introducing the chat interface and demonstrating how to send messages, annotate thoughts, start a new chat, and finish a task. They must then pass a short comprehension quiz before proceeding.

\textbf{Step 3: Conversations with thoughts.} Participants completed two open-ended, self-defined tasks, each within a 10-minute window, while chatting naturally with the AI and privately annotating their reasons for sending each message and their reactions to each assistant response. Each task could span multiple multi-turn conversations: participants were free to start a new conversation or end the task at any time, mirroring real-world use of conversational AI systems. Annotations were not visible to the AI, and multiple thoughts could be attached to a single message.

\textbf{Step 4: Survey.} After each task, participants described what they completed and what they expected from the AI. After both tasks, they filled out a demographic survey covering age, gender, education, occupation, AI usage frequency, and primary purposes.

Details of the data collection methods, platform design, and limitations are provided in Appendix~\ref{app:method_details}.

\subsection{Models Used}

Each participant interacted with one of 20 different models. We included frontier models available at the time of the study (e.g., GPT-5.4, Gemini 3.1 Pro Preview, Grok 4.20, and Opus 4.6), as well as smaller, open-weight models for comparison. Users were unaware of which model they were interacting with. Detailed statistics for each model, including the number of users, conversations, messages, and thoughts, are provided in Appendix~\ref{app:model_details}.

\subsection{Data Format}
\label{sec:data_format}



Each record in \ours corresponds to a single conversation in which a participant interacted with one of 20 language models to complete an open-ended everyday task. A participant may contribute multiple conversations across two tasks. For each conversation, we record a conversation ID, the model name and provider, the start and last-activity timestamps, a post-hoc task summary and task expectation, and the participant's survey responses (age, gender, education, occupation, AI-usage frequency, and primary use cases).

Each conversation is stored as an ordered list of messages. Each message includes a message ID, timestamp, type (either \texttt{user} or \texttt{assistant}), message content, and a list of participant thoughts annotated for that message. A thought is either a \emph{reason} attached to a user message or a \emph{reaction} attached to an assistant message. Each thought has its own timestamp, text content, and label, drawn from one of seven reason types or one of five reaction types.

%% file: sections/4-conv.tex
\section{Data Properties}

We characterize the data in \ours along two complementary axes: (1) properties of the conversations (Section~\ref{sec:conversations}) and (2) properties of the thoughts that drive the conversations (Section~\ref{sec:thought_properties}).

\subsection{Properties of Conversations}
\label{sec:conversations}

We highlight three conversation-level properties: a representative user base, long-horizon and topically diverse interactions, and the dominance of conversational turns that extend prior tasks.

\begin{figure}[t!]
  \centering
  \includegraphics[width=\linewidth]{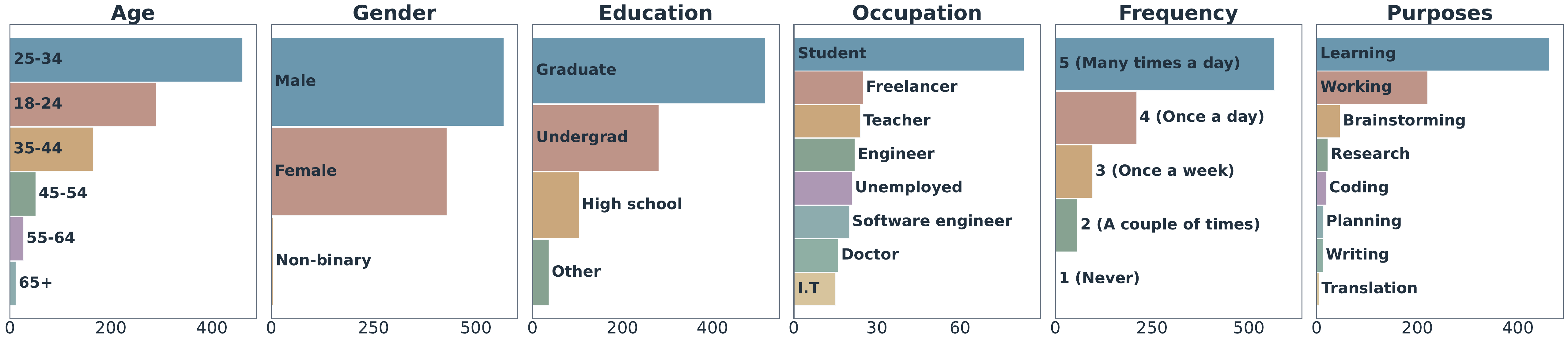}
  \caption{\textbf{Participant demographics and AI usage patterns in \ours}. The dataset covers age, gender, education level, occupation, frequency of AI usage, and primary purposes for using AI.}
  \label{fig:demographic}
\end{figure}

\begin{tcolorbox}[colback=lightblue, colframe=accent1, title=\textbf{Conversation Property 1: \ours captures a representative spectrum of users.}]
\ours pairs each conversation with rich demographic and usage metadata, reflecting a diverse spectrum of AI users and everyday use cases consistent with the profile of frequent real-world AI users (Figure~\ref{fig:demographic}).
\end{tcolorbox}

In Figure~\ref{fig:demographic}, we summarize the responses to our background survey (details in Appendix~\ref{app:survey}). Unlike existing in-the-wild conversation datasets such as WildChat~\citep{zhao2024wildchat}, which contain little participant-level information, \ours pairs each conversation with rich demographic and usage metadata, including age, gender, education, occupation, AI usage frequency, and primary purposes. Overall, the sample spans a broad range of backgrounds: participants range from 18 to 65+ in age, cover multiple education levels, and represent a variety of occupations, including students, freelancers, teachers, engineers, and others. That said, the participant distribution is skewed towards the 18--34 age range and those with at least an undergraduate degree, broadly consistent with the demographic profile of frequent generative AI users~\citep{liu2026earth, bick2026rapid}. Most participants report frequent AI use, often one or more times per day, for a range of purposes. The most common uses are learning and working, followed by brainstorming, research, and coding.

\begin{figure}[ht]
  \centering
  \begin{subfigure}[b]{0.55\textwidth}
    \centering
    \includegraphics[width=\linewidth]{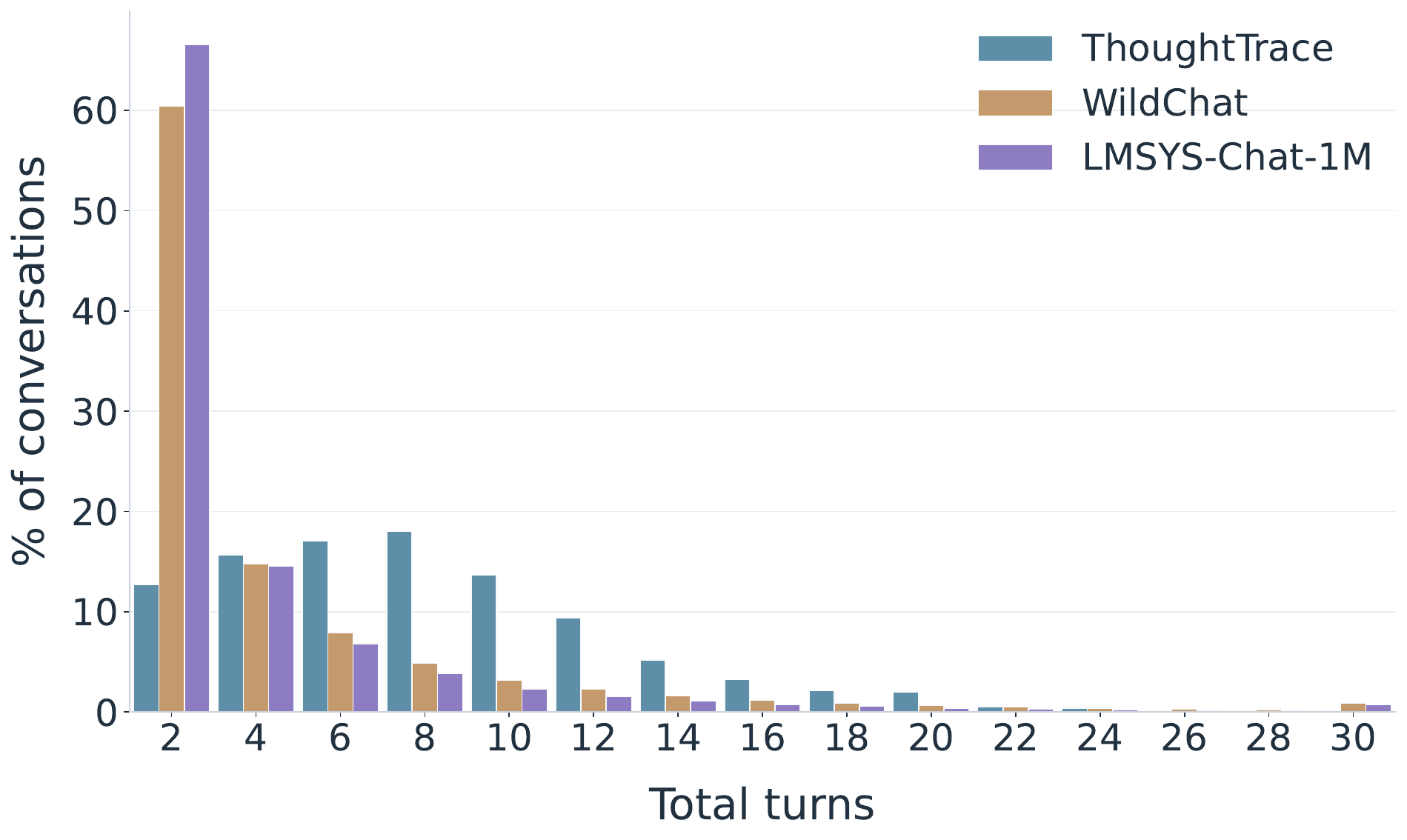}
    \caption{Turn distribution across the three datasets.}
    \label{fig:conv_length_turns}
  \end{subfigure}
  \hfill
  \begin{subfigure}[b]{0.44\textwidth}
    \centering
    \includegraphics[width=\linewidth]{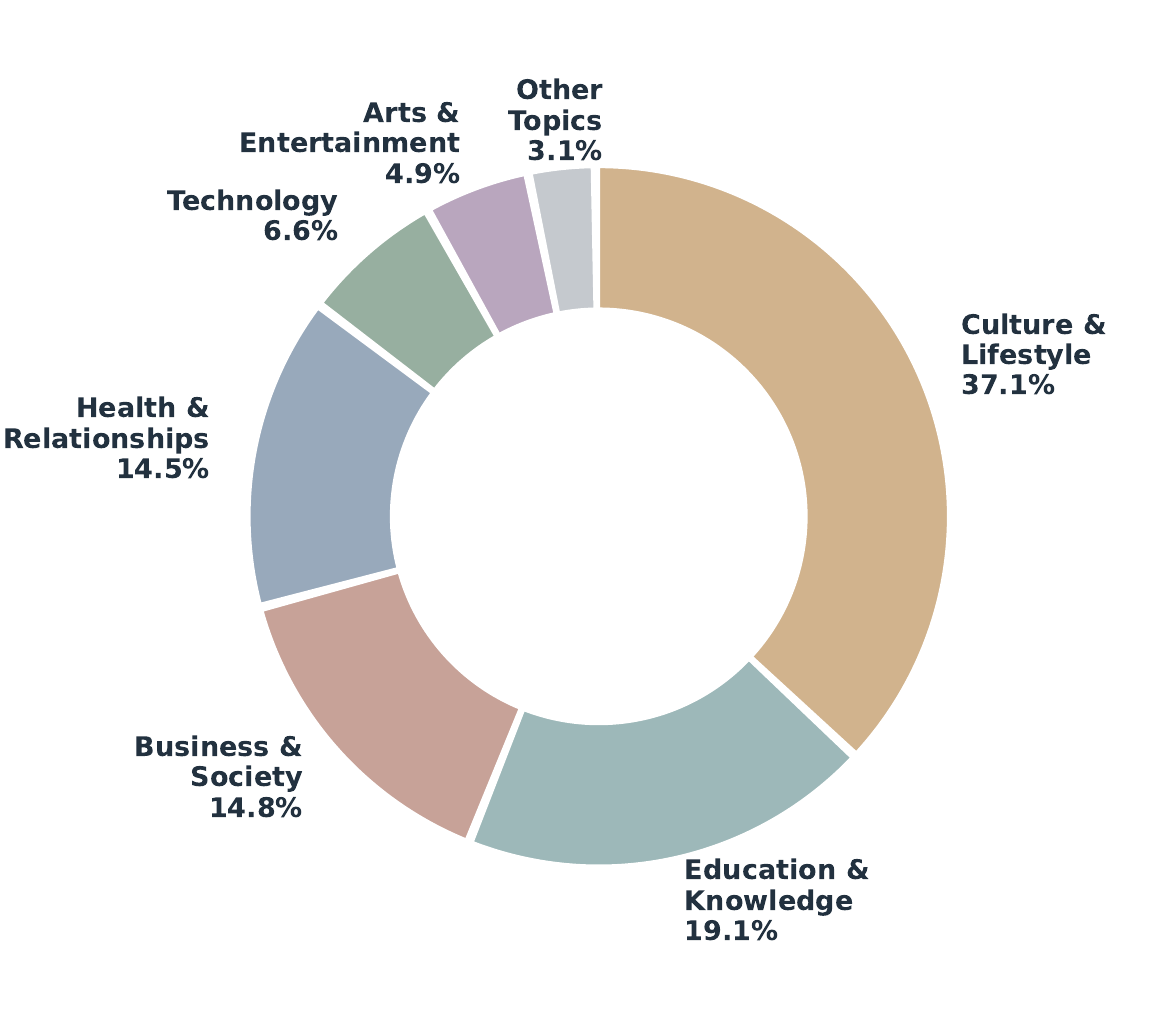}
    \caption{Topic distribution in \ours.}
    \label{fig:topic_distribution}
  \end{subfigure}
  \caption{\textbf{\ours covers long-horizon, topically diverse conversations.} (a) Turn distribution comparison between \ours, WildChat, and LMSYS-Chat-1M: \ours peaks at 6--8 turns, while the baselines skew heavily toward 2-turn exchanges. (b) Distribution of conversation topics in \ours, grouped into seven broad domains, with no single category dominating.}
  \label{fig:length_and_topics}
\end{figure}

\begin{tcolorbox}[colback=lightgreen, colframe=accent3, title=\textbf{Conversation Property 2: \ours features long-horizon diverse conversations.}]
\ours features high-quality, long-horizon conversations with a median of 8 turns (compared to 2 in both WildChat and LMSYS-Chat-1M) and spans seven broad topic categories and 36 fine-grained subtopics, with no single category dominating (Figure~\ref{fig:length_and_topics}).
\end{tcolorbox}

We compute conversation lengths at both the turn and token levels, with implementation details in Appendix~\ref{app:conv_property_2}. As shown in Figure~\ref{fig:length_and_topics}(a), \ours exhibits a substantially more balanced turn distribution, peaking around 6--8 turns with a median of 8 turns, whereas WildChat and LMSYS-Chat-1M are heavily skewed toward short 2-turn exchanges, which alone account for over 60\% and 67\% of their conversations, respectively. The cumulative token distribution per conversation follows a similar trend (Appendix~\ref{app:results_length}). This long-horizon property is critical because real-world AI usage is increasingly shifting toward sustained multi-turn interactions such as iterative coding, research, and planning, where tasks are more complex, and users' underlying intentions evolve across turns rather than being captured in a single prompt.

To characterize topical coverage, we label the relevant topics of each conversation, with implementation details in Appendix~\ref{app:conv_property_2}. Conversations are distributed across seven broad categories (Figure~\ref{fig:length_and_topics}(b)) and 36 fine-grained subtopics (see Figure~\ref{fig:topic_distribution_full} in Appendix~\ref{app:results_topics} for the full breakdown). Culture \& Lifestyle is the most prevalent broad topic category (covering areas such as travel, dining, and daily life), while Education \& Knowledge as well as Business \& Society are also well represented. At the fine-grained level, nine subtopics each exceed 5\% of the dataset (spanning Travel, Lifestyle, Food, Business, Geography, Education, Relationships, Health, and Technology), with a long tail of more specialized topics covering the remaining share. We also collect participants' task descriptions and AI expectations, with details in Appendix~\ref{app:survey} and visualizations in Appendix~\ref{app:results_descriptions_and_expectations}.

\label{sec:multiturn_relationship}

\begin{figure}[ht]
    \centering
    \includegraphics[width=0.6\linewidth]{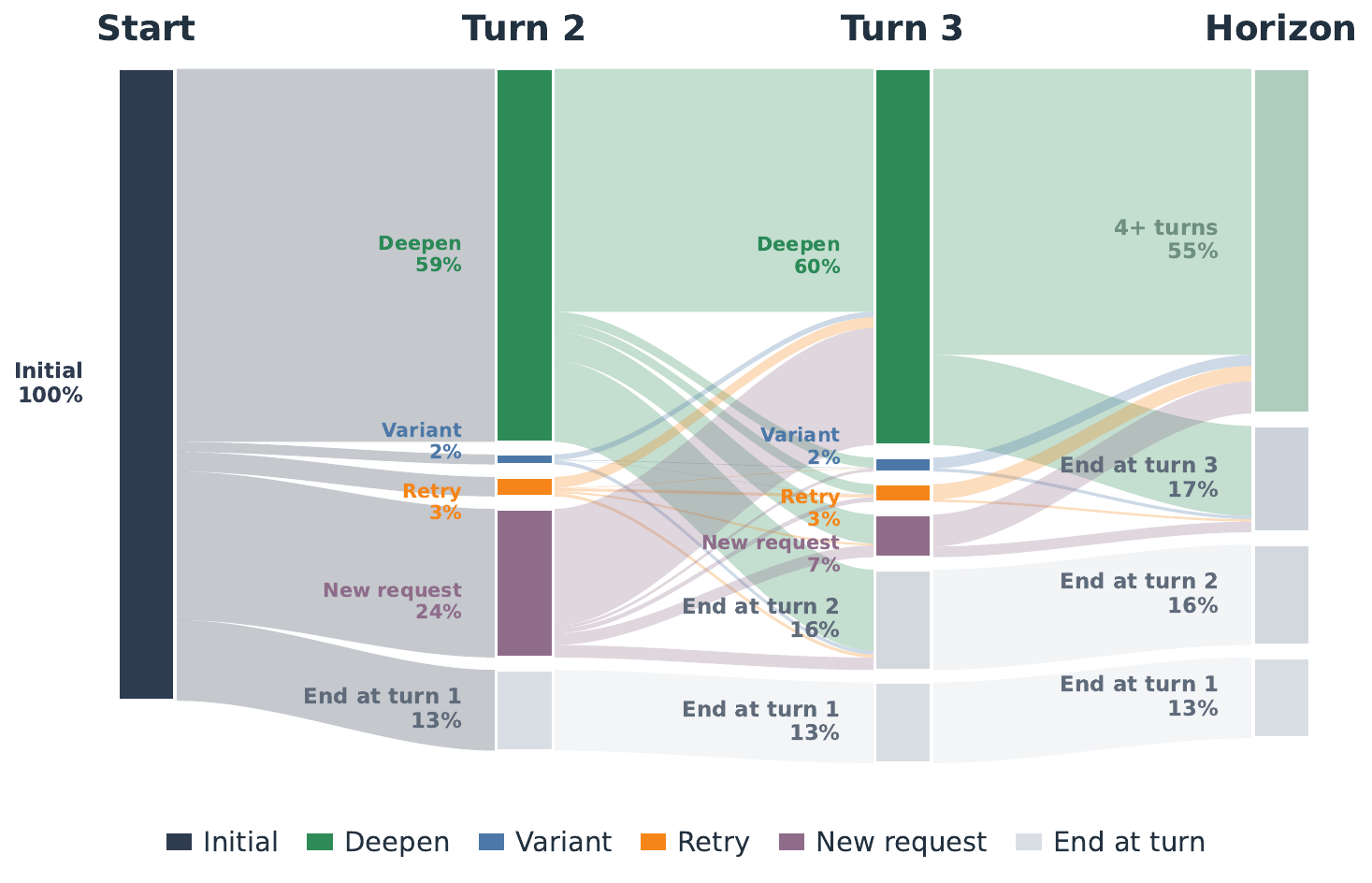}
    \caption{\textbf{Multi-turn Relationship Flow.} Turn-to-turn transitions of relationship labels across the first three turns and beyond, showing how conversations evolve from the initial request.}
    \label{fig:multi_turn_relationship_flow}
\end{figure}

\begin{tcolorbox}[colback=lightpurple, colframe=accent4, title=\textbf{Conversation Property 3: \ours conversations are dominated by task extension.}]
Extending, deepening, or building on the prior task accounts for 57.0\% of user turns, far outpacing new requests, re-attempts, and variations, and this extension pattern strengthens as conversations progress (Figure~\ref{fig:multi_turn_relationship_flow}).
\end{tcolorbox}

We analyze conversational structure by labeling the multi-turn relationship of each user message into one of five types: (1) First request (\textbf{25.2\%}); (2) Completely new request (\textbf{12.5\%}); (3) Re-attempt/revision on prior task (\textbf{2.9\%}); (4) New variation of prior task (\textbf{2.3\%}); and (5) Extend, deepen, or build on prior task (\textbf{57.0\%}). Implementation details are provided in Appendix~\ref{app:conv_property_3}, and the overall distribution is shown in Figure~\ref{fig:multi_turn_relationship_distribution}. Figure~\ref{fig:multi_turn_relationship_flow} visualizes how these relationships transition across the first three user turns. Extension dominates from turn~2 onward and becomes increasingly prevalent in later turns, while completely new requests appear as the second most common type but remain a relatively small share. Re-attempts and variations occur infrequently throughout, suggesting that users rarely need to rephrase or retry their requests.

%% file: sections/5-thought-properties.tex
\subsection{Properties of Thoughts}
\label{sec:thought_properties}

We highlight four thought-level properties: thoughts are different from messages, difficult for frontier LLMs to infer, span diverse reason and reaction categories, and are tied to conversation stages.

\begin{figure}[t]
\centering
\includegraphics[width=\linewidth]{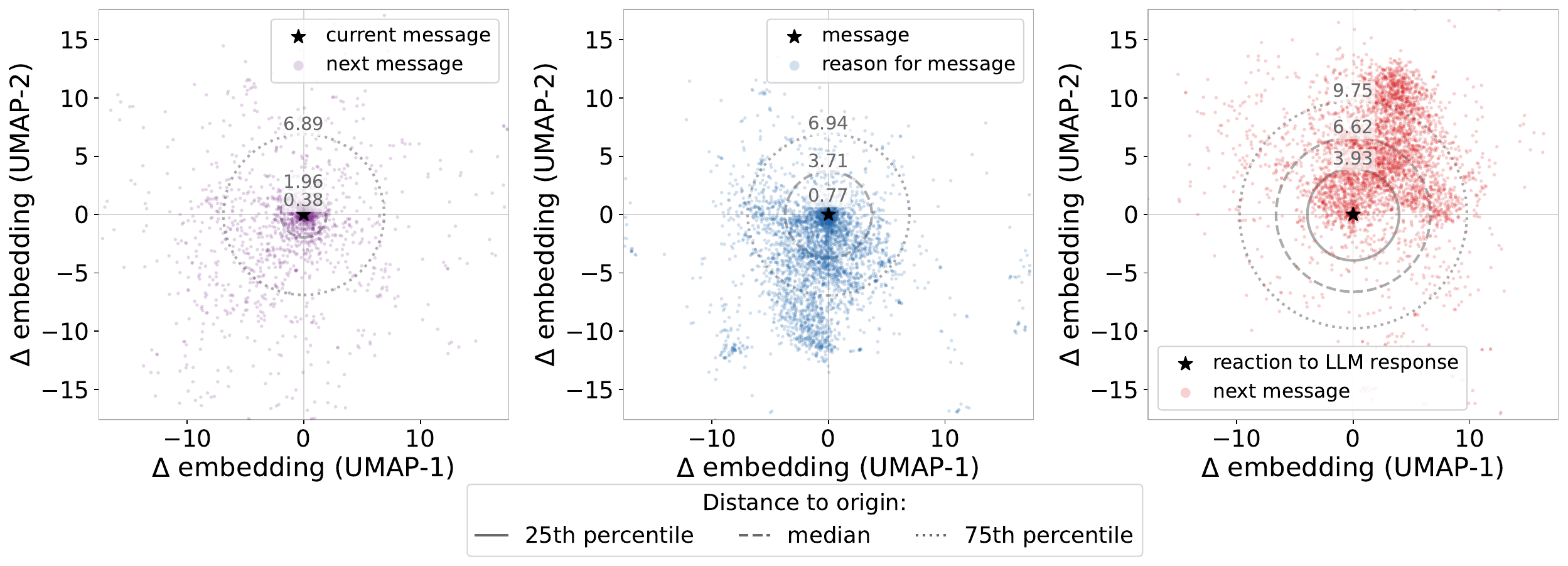}
\caption{\textbf{UMAP projections of embedding differences across three paired settings.} The star denotes the reference text embedding, and each dot represents the paired text embedding. Distance from the origin reflects the magnitude of the semantic shift between the paired texts. Circle annotations denote the 25th, 50th, and 75th percentile distances from the origin.}
\label{fig:delta_umap}
\end{figure}

\begin{tcolorbox}[colback=lightblue, colframe=accent1, title=\textbf{Thought Property 1: Thoughts are different from messages.}]
Thoughts capture substantial latent information not directly verbalized in conversation, as evidenced by both embedding-level shifts and LLM-based semantic coverage scoring, supporting their value as a distinct and complementary signal for understanding user behavior (Figure~\ref{fig:delta_umap}).
\end{tcolorbox}

A natural question is whether the thoughts in \ours merely restate what users already express in their messages, or whether they capture genuinely new information. We first evaluate at the \textbf{embedding} level: Figure~\ref{fig:delta_umap} visualizes the pairwise embedding differences between (i) a \textit{user message} and the \textit{next user message}, (ii) a \textit{user message} and its corresponding \textit{reason}, and (iii) a user's \textit{reaction} to an assistant response and their following \textit{next user message}. Consecutive user messages remain semantically close, reflecting the local coherence of conversation, whereas message--reason pairs show larger distances and reaction--next-message pairs exhibit the widest dispersion; quantitative distributional metrics in Appendix~\ref{app:results_emb_distribution} confirm this same trend. We then measure \textbf{semantic coverage via an LLM-based judge}, scoring on a 1 (no overlap) to 5 (full coverage) rubric how well a \textit{user message} covers (i) its \textit{reason} and (ii) the \textit{reaction} to the prior assistant response (see Appendix~\ref{app:thought_property_1} for implementation details). Average scores are 3.22 for reasons (partial overlap, missing the core of the thought) and 2.00 for reactions (minimal overlap). Together, these results show that thoughts capture substantial latent information not directly verbalized in conversation, supporting their value as a distinct and complementary signal for understanding user behavior.

\begin{tcolorbox}[colback=lightgreen, colframe=accent3, title=\textbf{Thought Property 2: Thoughts are difficult for LLMs to infer.}]
Thoughts are consistently difficult for three frontier language models to infer from context, underscoring the value of the explicit thought annotations in \ours.
\end{tcolorbox}

We prompt LLMs to infer (1) the user's reason for their most recent message, given the conversation up to that point, and (2) the user's reaction to the assistant's most recent message, given the conversation up to that point plus the user's next message if available. An LLM-as-a-judge scores each inference against the human annotation on a 1-to-5 semantic similarity scale. Implementation details are provided in Appendix~\ref{app:thought_property_2}. Averaged across three frontier models (GPT-5.4, Gemini 3.1 Pro Preview, and Claude Opus 4.6), the mean similarity score is 2.93 for reasons (2.83, 3.02, 2.94, respectively) and 2.54 for reactions (2.36, 2.87, 2.40), all falling between minimal (2) and partial overlap (3). The gap reflects the fact that thoughts are underspecified by surface-form text: multiple plausible reasons or reactions are consistent with the same context, and the correct one often depends on unobservable constraints, stakes, or interpretations from users. Appendix~\ref{app:inference_failure} shows qualitative failure cases in which models misread the user's underlying intent or fabricate reactions they did not have. Together with Property~1, these results confirm that thoughts are both distinct from utterances and difficult to recover from context, underscoring the value of explicit thought annotations in \ours.

\begin{figure}[t!]
    \centering
    \includegraphics[width=0.98\linewidth]{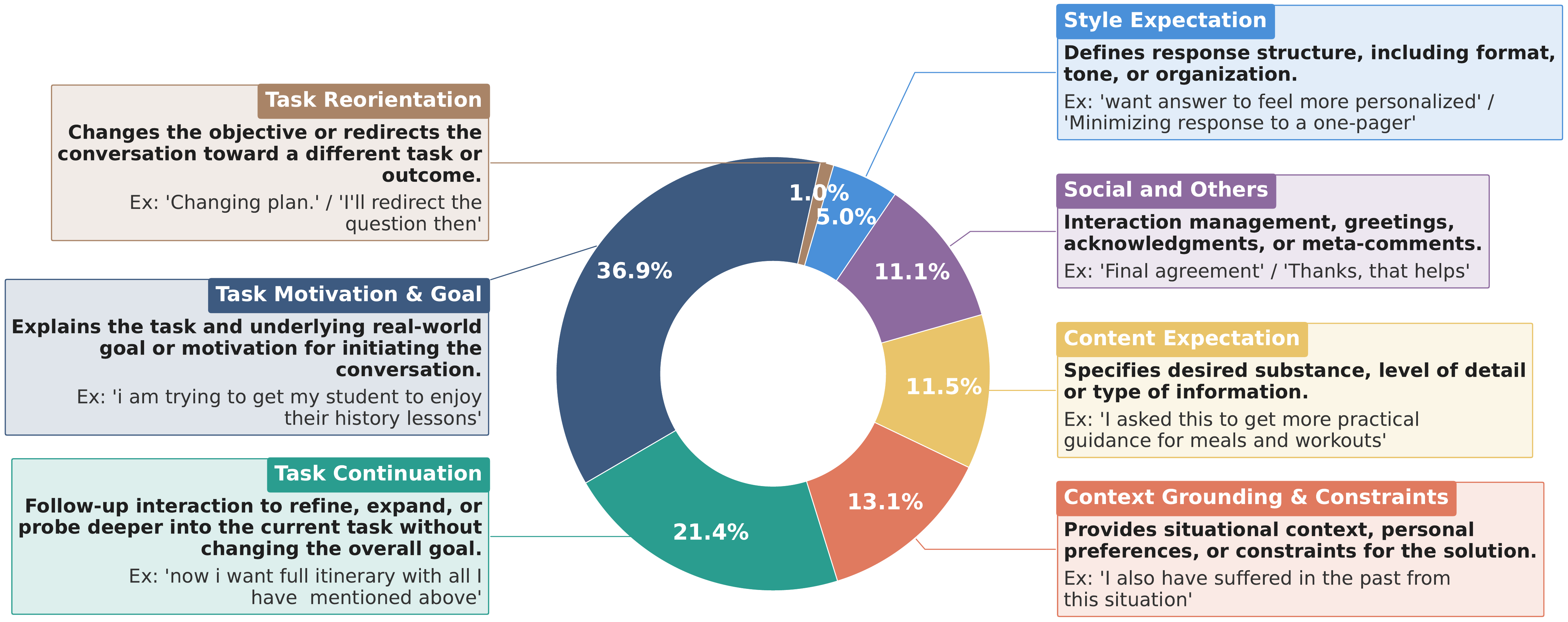}
    \caption{\textbf{Distribution of seven user reason types in \ours, with definitions and examples from the dataset.} Task Motivation \& Goal is the most prevalent (36.9\%), followed by Task Continuation (21.4\%) and Context Grounding \& Constraints (13.1\%). More and longer examples with full conversation context are on the \href{https://thoughttrace-project.github.io/examples.html}{project website}.}
    \label{fig:reasons_distribution}
\end{figure}

\begin{figure}[t!]
    \centering
    \vspace{-0.5em}
    \includegraphics[width=0.98\linewidth]{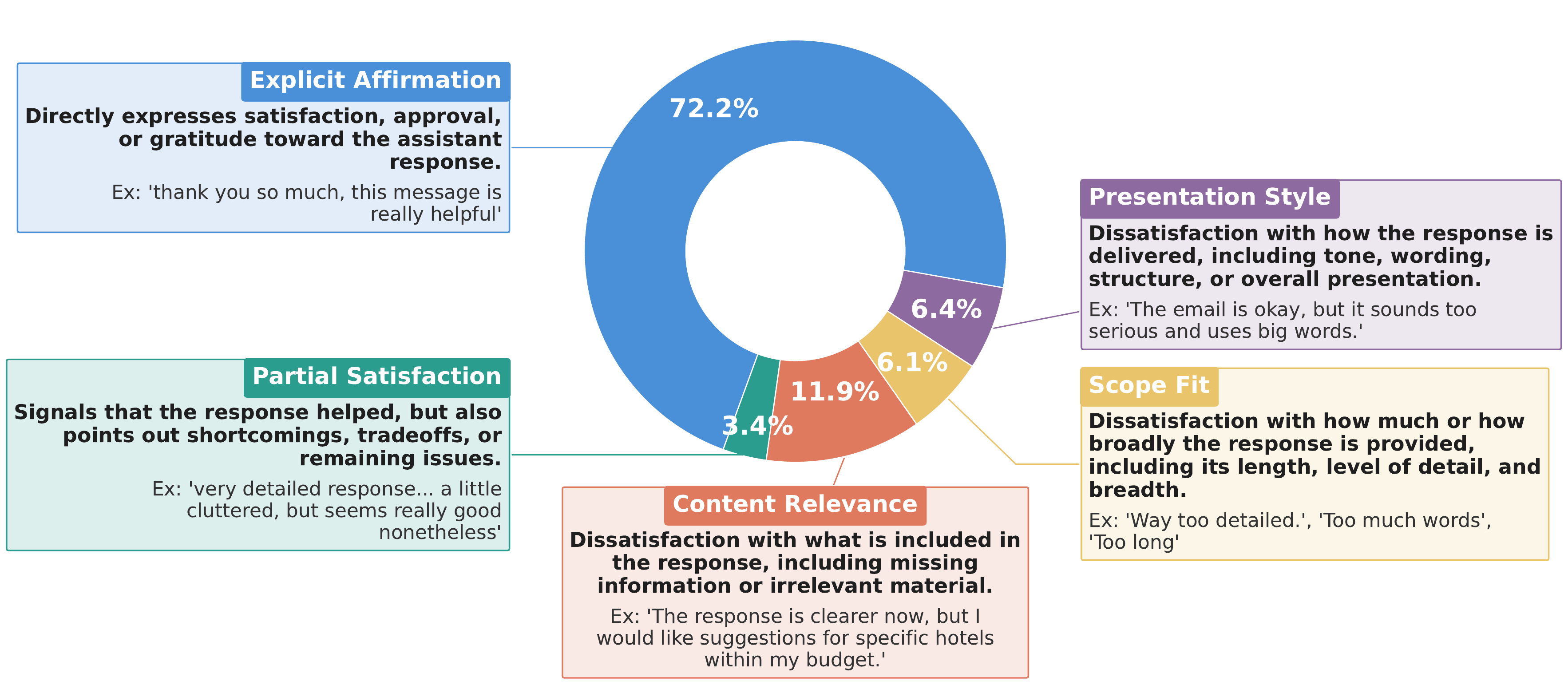}
    \caption{\textbf{Distribution of five user reaction types in \ours, with definitions and examples from the dataset.} Explicit Affirmation dominates (72.2\%), while dissatisfaction is often driven by Content Relevance (11.9\%), Presentation Style (6.4\%), and Scope Fit (6.1\%). More and longer examples with full conversation context are on the \href{https://thoughttrace-project.github.io/examples.html}{project website}.}
    \label{fig:reactions_distribution}
\end{figure}

\begin{tcolorbox}[colback=lightred, colframe=accent2, title=\textbf{Thought Property 3: Thoughts are diverse in content.}]
Thoughts in \ours span seven reason categories and five reaction categories, capturing a diverse set of unspoken contexts that range from high-level motivations and grounding details to targeted sources of dissatisfaction (Figures~\ref{fig:reasons_distribution} and~\ref{fig:reactions_distribution}).
\end{tcolorbox}

To analyze this diversity, we label user thoughts using an LLM-based annotation framework (details in Appendix~\ref{app:thought_property_3}). As shown in Figure~\ref{fig:reasons_distribution}, the reasons behind user utterances span seven distinct categories, ranging from high-level drivers such as \textit{Task Motivation \& Goal} (36.9\%) and \textit{Task Continuation} (21.4\%) to finer-grained context and preference specifications such as \textit{Context Grounding \& Constraints} (13.1\%), \textit{Content Expectation} (11.5\%), and \textit{Style Expectation} (5.0\%). Complementing this, Figure~\ref{fig:reactions_distribution} shows that user reactions decompose into five categories: while \textit{Explicit Affirmation} dominates at 72.2\% and \textit{Partial Satisfaction} accounts for 3.4\%, the remaining reactions reveal targeted sources of dissatisfaction, including \textit{Content Relevance} (11.9\%), \textit{Presentation Style} (6.4\%), and \textit{Scope Fit} (6.1\%). Together, these distributions show that thoughts in \ours are not monolithic, but span a rich spectrum of latent intents and evaluative judgments, from why a user initiates a turn to how they privately assess the assistant's reply. This diversity suggests that modeling user satisfaction from surface utterances alone is insufficient, and that thought-level signals are essential for diagnosing \textit{which} aspect of a response succeeds or fails and aligning future assistants accordingly.

\begin{figure}[t]
  \centering
  \begin{subfigure}[b]{0.49\linewidth}
    \centering
    \includegraphics[width=\linewidth]{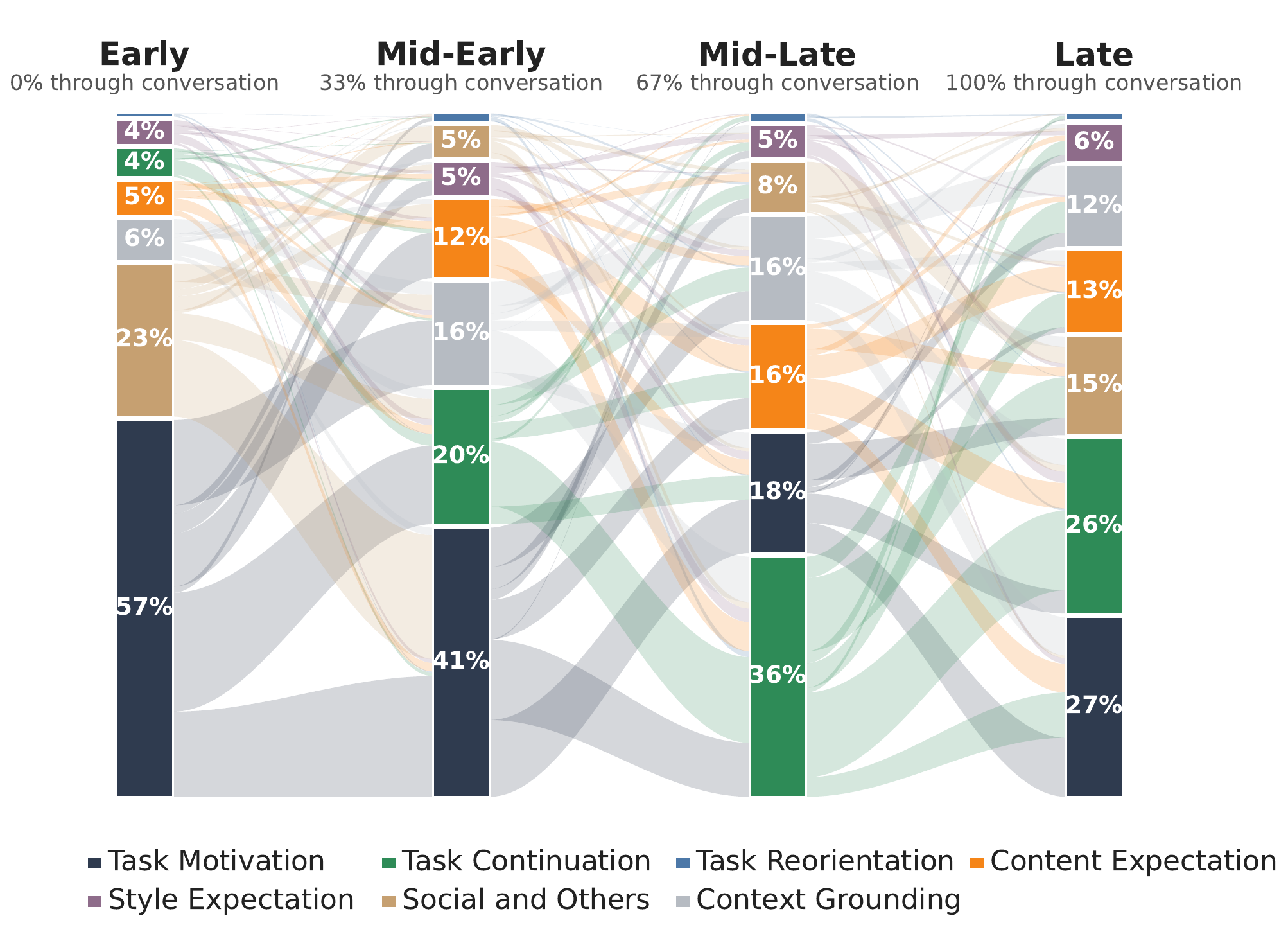}
    \caption{Reason types across conversation stages.}
    \label{fig:reasons_flow}
  \end{subfigure}
  \hfill
  \begin{subfigure}[b]{0.49\linewidth}
    \centering
    \includegraphics[width=\linewidth]{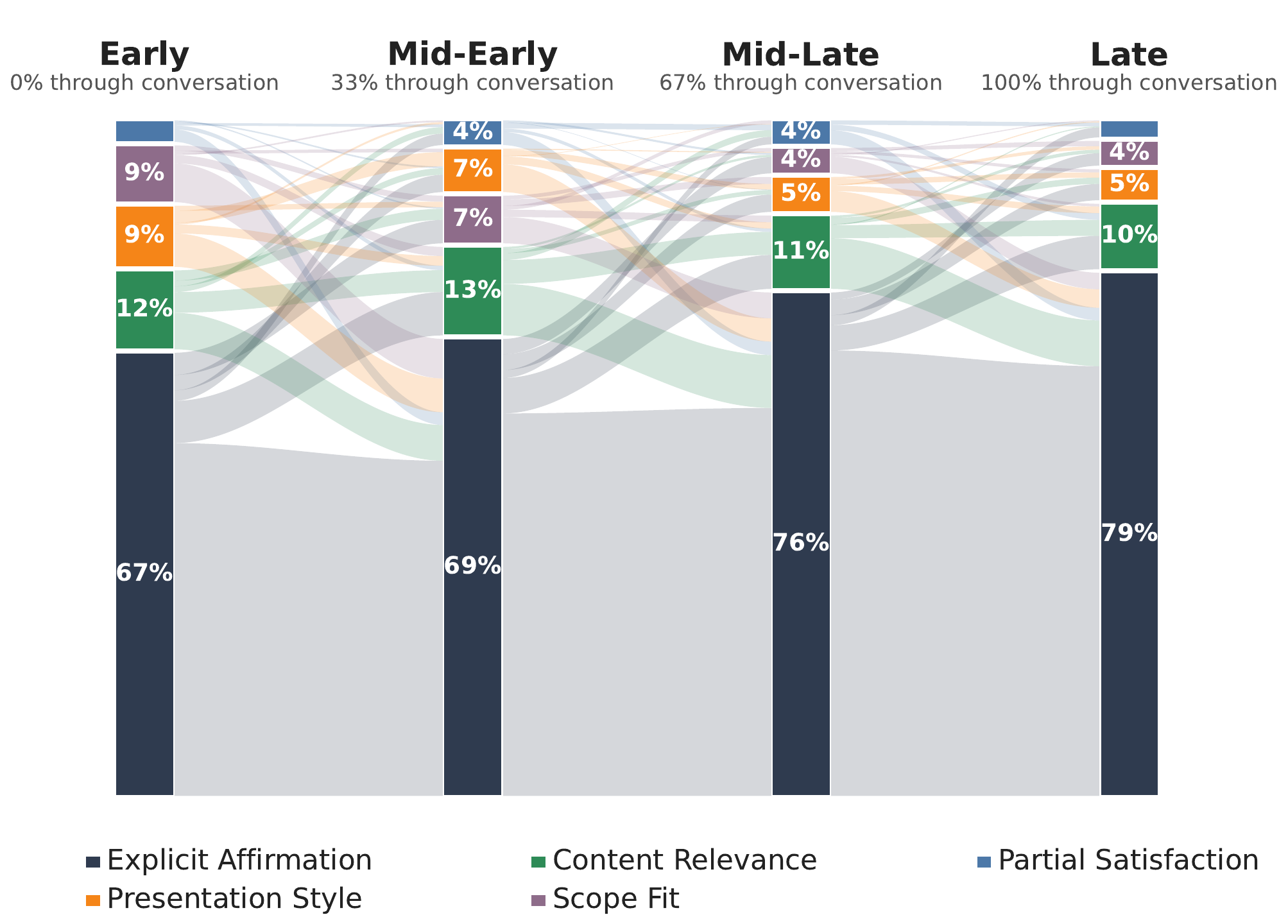}
    \caption{Reaction types across conversation stages.}
    \label{fig:reaction_flow}
  \end{subfigure}
  \caption{\textbf{Thought dynamics across conversation stages.} (a) Reason-type distribution shifts from Task Motivation \& Goal in early turns to Task Continuation and context- and expectation-driven reasons in later stages. (b) Reaction-type distribution shows a steady increase in Explicit Affirmation from early to late stages.}
  \label{fig:thought_flow}
\end{figure}

\begin{tcolorbox}[colback=lightpurple, colframe=accent4, title=\textbf{Thought Property 4: Thought dynamics depend on conversation stages.}]
Thought dynamics depend on conversation stages and multi-turn relationships between messages, while remaining largely independent of conversation topics or length (Figure~\ref{fig:thought_flow}).
\end{tcolorbox}

Figure~\ref{fig:reasons_flow} shows that \emph{Task Motivation \& Goal} dominates early turns, while \emph{Task Continuation} increases and becomes the primary driver in mid-to-late stages. \emph{Context Grounding \& Constraints} and expectation-related reasons remain a substantial portion throughout the middle stages. Figure~\ref{fig:reaction_flow} shows a parallel shift in reactions: \emph{Explicit Affirmation} increases from 67\% in early stages to 79\% in later stages, while more critical reactions such as \emph{Presentation Style} and \emph{Scope Fit} decline, suggesting that user satisfaction improves as interactions converge toward acceptable responses. Figure~\ref{fig:thought_vs_relationship} corroborates these trends at the message-relationship level and further shows that users predominantly extend the conversation regardless of their annotated reaction type. By contrast, thought types exhibit no clear relationship with conversation topics or lengths, with additional results discussed in Appendix~\ref{app:thought_vs_properties} and Figures~\ref{fig:reason_vs_topic}--\ref{fig:reaction_vs_length}.

%% file: sections/6-thought-utility.tex
\newpage
\section{Utility of Thoughts}
\label{sec:thought_utility}

\begin{figure}[ht]
    \centering
    \includegraphics[width=\linewidth]{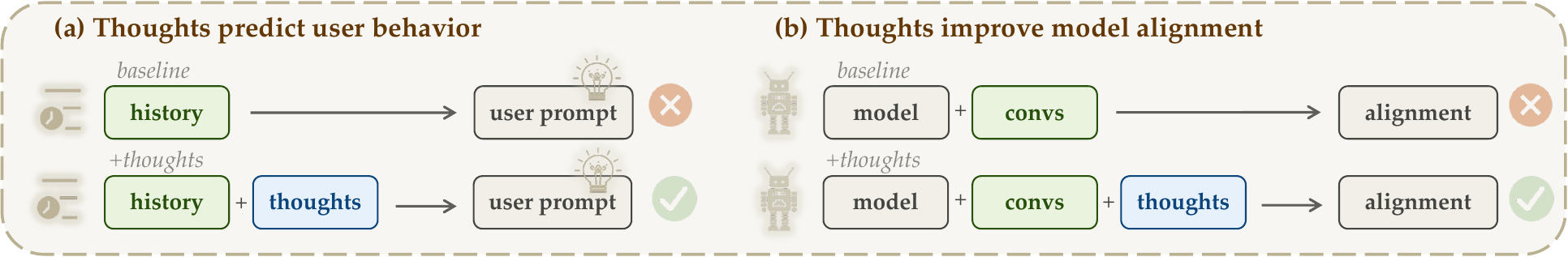}
    \caption{\textbf{Two experiments demonstrating the utility of thoughts.} Thoughts provide actionable signals for (a) predicting user behavior and (b) improving model alignment.}
    \label{fig:utility}
\end{figure}

We define the utility of thoughts as the actionable signals they provide beyond what is observable in conversation transcripts. As shown in Figure~\ref{fig:utility}, we validate this utility through two experiments: predicting user behavior (Section~\ref{sec:prediction}) and improving model alignment (Section~\ref{sec:alignment}), pointing toward future work on user modeling, thought-centered evaluation, and personalized assistant training.

\subsection{Thoughts Predict User Behavior}
\label{sec:prediction}

Predicting user behavior is important as (1) it helps models anticipate user needs and provide more proactive, personalized assistance; and (2) it supports high-fidelity user simulators, which provide a scalable and reproducible alternative to real human interaction during model training and evaluation.

\begin{tcolorbox}[colback=lightblue, colframe=accent1, title=\textbf{Takeaway [Q1]: Do thoughts improve user message prediction at inference time?}]
Access to thought annotations substantially improves next user message prediction across three frontier models, raising the average prediction semantic similarity from 21.6 to 30.6 (Table~\ref{tab:user_prediction_inference}).
\end{tcolorbox}

\textbf{Experimental setup.} We test whether access to thought annotations at inference time improves the LLM's ability to anticipate the user's next message. For each conversational turn from \ours, we compare two settings: (1) predicting the next message from the conversation history alone, and (2) predicting it from the same history augmented with the user's annotated reasons and reactions. We evaluate three frontier models (GPT-5.4, Gemini 3.1 Pro Preview, Claude Opus 4.6) under both settings, and score each prediction's semantic similarity to the ground truth on a 0--100 scale using an LLM judge randomly drawn from the two other models. Details are in Appendix~\ref{app:thought_utility_1}.

\begin{table}[h]
\centering
\caption{\textbf{User message prediction results.} Three frontier models are evaluated, with and without access to annotated thoughts at inference time.}
\label{tab:user_prediction_inference}
\renewcommand{\arraystretch}{1.15}
\begin{tabular}{lcccc}
\toprule
\textbf{Method} & \textbf{GPT} & \textbf{Gemini} & \textbf{Opus} & \textbf{Avg.} \\
\midrule
History-only & 21.4 & 22.1 & 21.3 & 21.6 \\
Thought-augmented & \textbf{27.4} & \textbf{28.9} & \textbf{35.5} & \textbf{30.6} \\
\bottomrule
\end{tabular}
\end{table}

\textbf{Results.} As shown in Table~\ref{tab:user_prediction_inference}, access to thought annotations substantially improves next-message prediction across all three models, raising the average performance from 21.6 to 30.6, a 41.7\% relative gain. The effect is largest for Claude Opus 4.6, whose performance increases by 14.2, while GPT-5.4 and Gemini 3.1 Pro Preview show smaller but consistent gains of 6.0 and 6.8, respectively. These results suggest that the latent reasons and reactions captured in \ours{} help predict future user messages, providing actionable signals beyond the observable conversation history.

\textbf{Implications for future research.} Our results demonstrate that user thoughts can steer user behavior predictions, which suggests the value of thoughts in simulating users. For example, whereas prior work trains user simulators by fine-tuning LLMs to predict the \textit{next user message} from conversation history \citep{naous2025flipping, abdulhai2025consistently, wu2026humanlm}, future work could train models to jointly predict \textit{thoughts} and user messages. Strong user simulators can help anticipate user needs and thereby guide models to assist users in a more proactive and personalized manner \citep{qian2025userrl, sun2025training}.

\subsection{Thoughts Improve Model Alignment}
\label{sec:alignment}

Model alignment is important because it helps models produce responses that better match human intentions, values, and preferences, making them more useful and trustworthy in real-world settings. Real user interactions and feedback provide natural, multifaceted signals for improving alignment.

\begin{tcolorbox}[colback=lightgreen, colframe=accent3, title=\textbf{Takeaway [Q2]: Do thoughts provide better alignment signals than messages?}]
Thought-guided rewrites outperform message-guided rewrites by +4.5\% on Arena-Hard, exceed the base model by +25.6\%, and WildChat baseline by +6.6\%, indicating thoughts capture richer dissatisfaction and revision signals than users express (Table~\ref{tab:alignment}).
\end{tcolorbox}

\textbf{Experimental setup.} Prior work on learning from natural conversations revises unsatisfactory responses using users' follow-up messages, pairing these \textit{message-guided rewrites} with original messages for preference learning \citep{shi2024wildfeedback, jin2025era}. Leveraging thoughts in \ours, we instead identify unsatisfactory responses via the \textit{dissatisfaction reaction labels} from Section~\ref{sec:thought_properties} and prompt the model to revise them using the \textit{thought content}, producing \textit{thought-guided rewrites}. Both are paired with originals for DPO training \citep{rafailov2023direct}. We compare: (1) the base Qwen3.5-4B \citep{yang2025qwen3}; (2) message-guided rewrites on WildChat; (3) message-guided rewrites on \ours; and (4) thought-guided rewrites on \ours. Models are evaluated on Arena-Hard \citep{li2024crowdsourced}, a robust instruction-following benchmark with 98.6\% correlation to human preference. Details are in Appendix~\ref{app:thought_utility_2}.

\begin{table}[h]
\centering
\caption{\textbf{Model alignment results on Arena-Hard.} We report both win rates (\%) and style-controlled win rates (SC Win, \%).}
\label{tab:alignment}
\renewcommand{\arraystretch}{1.15}
\begin{tabular}{lcc}
\toprule
\textbf{Method} & \textbf{Win} & \textbf{SC Win} \\
\midrule
Qwen3.5-4B & 24.6 & 22.5 \\
 + WildChat & 41.8 & 41.5 \\
 + \ours (messages) & 44.0 & 43.6 \\
 + \ours (thoughts) & \textbf{47.9} & \textbf{48.1} \\
\bottomrule
\end{tabular}
\end{table}

\textbf{Results.} As shown in Table~\ref{tab:alignment}, fine-tuning Qwen3.5-4B on \ours\ substantially improves Arena-Hard performance, with thought-guided rewrites achieving the largest style-controlled gains over both the base model (+25.6\%) and the WildChat baseline (+6.6\%). We highlight three findings: \textbf{(1)} within \ours, thought-guided rewrites outperform message-guided ones (+4.5\%), indicating that thoughts encode richer dissatisfaction and revision signals than users explicitly articulate in messages; \textbf{(2)} across the same \ours\ conversations, thoughts surface 1{,}000 dissatisfaction instances compared to 450 in messages (2.2$\times$ more), yielding denser supervision; and \textbf{(3)} compared to the WildChat baseline, the message-guided variant of \ours uses fewer conversations and a smaller training set yet still outperforms it (+2.1\%), reflecting the higher quality of \ours. More broadly, thoughts provide ground-truth user reactions rather than behavioral proxies, and unify \textit{which} response is unsatisfactory and \textit{how} to revise it into a single supervision signal.

\textbf{Implications for future research.} We advocate broader adoption of our framework for collecting thoughts as richer and more effective signals for model training. In terms of training methods, our experiments use only reactions; a natural next step is to additionally incorporate reasons and leverage both signals jointly. Moreover, thought-guided supervision could be extended to reward modeling and online alignment \citep{peng2026wildreward}, and thought-guided On-Policy Distillation (OPD) may provide rich signals for online improvement \citep{wang2026openclaw, buening2026aligning, hubotter2026reinforcement}.

%% file: sections/7-conclusion.tex
\section{Conclusion}
\label{sec:conclusion}

In this paper, we introduce \ours, the first large-scale dataset that pairs real-world human-AI conversations with users' self-reported thoughts. Our analysis establishes thoughts as a distinct data modality: they capture latent information beyond surface messages, are difficult for frontier LLMs to infer, span diverse content, and vary across conversation stages. We further demonstrate their downstream utility, showing that thoughts improve user behavior prediction at inference time and provide fine-grained alignment signals for training. Together, these results position user thoughts as a foundational signal for studying the cognitive dynamics behind human-AI interaction and open new directions for building assistants that better model users, learn from latent thoughts, and evaluate success beyond surface-level utterances toward intent, satisfaction, and subjective experience.


\textbf{Limitations and Future Work.} \ours has several limitations inherent to in-situ thought collection (Appendix~\ref{app:limitations}). First, asking users to externalize thoughts may shape the interaction itself, as anticipating annotation can sharpen or polarize their reasoning. Second, the dataset captures only consciously accessible reasoning, leaving subconscious judgments unobserved. Third, recruitment through Prolific introduces a modest selection effect, though our demographic analysis suggests the sample remains broadly representative of frequent AI users. Finally, our evaluation covers only two downstream use cases, and a more comprehensive empirical investigation is left to future work.


\section*{Acknowledgments}

Chuanyang Jin is supported by the Amazon AI PhD Fellowship. We sincerely thank the JHU SCAI Lab and the DSAI communities for their helpful comments and feedback.

\section*{Author Contribution Statement}
\footnotesize
\scalebox{1}{
\begin{tabular}{r |@{\foo} l}
\textbf{Project Conception} & [\textsc{Chuanyang}, \textsc{Tianmin}] \\
\textbf{Data Collection Design} & [\textsc{Chuanyang}] \\
\textbf{Metadata Processing} & [\textsc{Chuanyang}] \\
\textbf{Conversation Property Analysis} & [\textsc{Chuanyang}, \textsc{Binze}, \textsc{Cathy}] \\
\textbf{Thought Property Analysis} & [\textsc{Chuanyang}, \textsc{Binze}, \textsc{Cathy}] \\
\textbf{Thought Utility Experiments} & [\textsc{Chuanyang}, \textsc{Haopeng}, \textsc{Tianjian}] \\
\textbf{Advising} & [\textsc{Tianmin}, \textsc{Maximillian}, \textsc{Hongxiang}, \textsc{Shayne}] \\
\textbf{Manuscript Writing} & [\textsc{Chuanyang}, \textsc{Binze}]\\
\textbf{Manuscript Editing and Feedback} & [\textsc{Everyone}]\\
\end{tabular}
}
\normalsize

%% file: appendices/1-models.tex
\section{Details of Models Used in \ours}
\label{app:model_details}

\ours contains data from 1,058 high-value users, comprising 2,155 timestamped conversations, 17,058 interaction turns, and 10,174 thought annotations, collected via a chatbot service powered by \textbf{20 different language models}. Model-wise statistics are provided in Table~\ref{tab:model_statistics}.

For all models, we use an inference temperature of 0.7. For models with a thinking mode, the chatbot displays only the final response, without revealing intermediate reasoning traces enclosed in <think> and </think>. During the thinking process, a loading indicator is shown with the text ``AI is thinking...''.

\begin{table}[htb]
\centering
\caption{\textbf{Model-wise statistics of the \ours dataset across 20 language models.} ``Open'' indicates whether the model weights are publicly available. Each value corresponds to the number of users, conversations, messages, and thought annotations associated with a given model, reflecting diverse real-world human-AI interactions.}
\label{tab:model_statistics}
\small
\renewcommand{\arraystretch}{1.3}
\begin{tabular}{l|c|ccccc}
\toprule
\textbf{Model} & \textbf{Open} & \textbf{\#Users} & \textbf{\#Conversations} & \textbf{\#Messages} & \textbf{\#Thoughts} \\
\midrule
\gpticon OpenAI: GPT-5.4 & \ding{55} & 162 & 337 & 2,462 & 1,474 \\
\geminiicon Google: Gemini 3.1 Pro Preview & \ding{55} & 155 & 313 & 2,568 & 1,553 \\
\grokicon xAI: Grok 4.20 & \ding{55} & 100 & 210 & 1,782 & 905 \\
\claudeicon Anthropic: Claude Opus 4.6 & \ding{55} & 70 & 141 & 1,222 & 712 \\
\claudeicon Anthropic: Claude Sonnet 4.6 & \ding{55} & 68 & 134 & 1,224 & 709 \\
\minimaxicon MiniMax: MiniMax M2.7 & \ding{55} & 50 & 100 & 608 & 344 \\
\gpticon OpenAI: gpt-oss-120b & \ding{51} & 36 & 70 & 372 & 232 \\
\kimiicon MoonshotAI: Kimi K2.5 & \ding{51} & 35 & 71 & 552 & 382 \\
\geminiicon Google: Gemma 4 26B A4B & \ding{51} & 35 & 69 & 504 & 342 \\
\qwenicon Qwen: Qwen3.6 Plus & \ding{55} & 34 & 72 & 424 & 258 \\
\mimoicon Xiaomi: MiMo-V2-Pro & \ding{51} & 34 & 67 & 690 & 407 \\
\gpticon OpenAI: GPT-4o-mini & \ding{55} & 33 & 69 & 664 & 498 \\
\geminiicon Google: Gemini 3 Flash Preview & \ding{55} & 33 & 69 & 636 & 401 \\
\grokicon xAI: Grok 4.1 Fast & \ding{55} & 33 & 63 & 462 & 289 \\
\stepfunicon StepFun: Step 3.5 Flash & \ding{51} & 30 & 64 & 502 & 269 \\
\zaiicon Z.ai: GLM 5 & \ding{51} & 30 & 64 & 406 & 296 \\
\metaicon Meta: Llama 3.3 70B Instruct & \ding{51} & 30 & 62 & 492 & 275 \\
\mistralicon Mistral: Mistral Small 4 & \ding{51} & 30 & 61 & 572 & 309 \\  
\claudeicon Anthropic: Claude Haiku 4.5 & \ding{55} & 30 & 60 & 532 & 315 \\
\deepseekicon DeepSeek: DeepSeek V3.2 & \ding{51} & 30 & 59 & 384 & 204 \\ 
\midrule
\textbf{Total} & & \textbf{1,058} & \textbf{2,155} & \textbf{17,058} & \textbf{10,174} \\
\bottomrule
\end{tabular}
\end{table}

%% file: appendices/2-results.tex
\section{Additional Results}
\label{app:results_details}

\subsection{Qualitative Examples of Frontier Model Failures in Thought Inference}
\label{app:inference_failure}

We present four representative failure cases that illustrate why thought inference remains challenging for frontier models. Examples 1 and 2 target the \emph{Reason} thought type, where the model must predict why the user sends their next message, while Examples 3 and 4 target the \emph{Reaction} thought type, where the model must predict how the user feels about the assistant's previous reply. The cases span three frontier models: \texttt{GPT-5.4} (Examples 1 and 3), \texttt{Claude Opus 4.6} (Example 2), and \texttt{Gemini 3.1 Pro Preview} (Example 4), each receiving a score of 1/5 against the ground-truth thought.

The failure modes cluster into two patterns. For \emph{Reason} prediction, models latch onto the most recent surface topic and miss the user's actual motivation: in Example 1, \texttt{GPT-5.4} binds the pronoun \textit{``it''} to the just-explained switches rather than to the broader licensing concern, and in Example 2,  \texttt{Claude Opus 4.6} reads a newly raised problem as a standalone question while overlooking its metacognitive role in the ongoing problem-solving loop. For \emph{Reaction} prediction, models
conflate the content of the follow-up message with the user's affective response: in Example 3, \texttt{GPT-5.4} fabricates dissatisfaction from a ``no microwave'' clarification despite the user's genuine appreciation, and in Example 4, \texttt{Gemini 3.1 Pro Preview} misreads frustration over an over-scoped reply as approval. Together, these cases show that current models infer thoughts from local textual cues rather than from the user's
underlying intent or affect.

\begin{AIbox}{Failure Example 1 --- Predicting the User's Reason (GPT-5.4)}
\textbf{Model:} GPT-5.4 \quad | \quad \textbf{Thought Type:} Reason \quad | \quad \textbf{Score:} 1/5

\dashedrule

\textbf{Conversation Context:}\\
\textbf{User:} I need help with how to start a motorcycle, learner level\\
\textbf{Assistant:} Here's a simple, step-by-step guide for a complete beginner on how to start a motorcycle safely. \textit{[570 words omitted: gear checklist, mounting, starting a fuel-injected engine, choke procedure for older bikes, and the basic ``ready to move'' hand/foot position.]}\\
\textbf{User:} What does the switches do?\\
\textbf{Assistant:} Here's a beginner-friendly explanation of all the common switches and controls on a typical learner motorcycle. \textit{[650 words omitted: tables for left-handlebar switches (clutch, high/low beam, turn signals, horn, choke), right-handlebar switches (front brake, throttle, kill switch, starter button), foot controls (gear shifter, rear brake), dashboard lights, and a quick start checklist.]}\\
\textbf{User:} Is it a must for a riders licence

\dashedrule

\textbf{Ground-truth Reason:}\\
Legal reasons, to avoid being flagged or arrested, and other safety reasons, avoid accidents for learners, learning alone.

\vspace{4pt}
\textbf{Predicted Reason (GPT-5.4):}\\
I want to know whether learning and being tested on these motorcycle switches and controls is required to get a motorcycle rider's licence.

\dashedrule

\textbf{Summary:} The user's real motivation is about the legal/safety necessity of having a rider's licence before riding (so they aren't arrested or unsafe as a lone learner). The model instead interprets ``it'' as the just-explained switches material and predicts a question about licence-exam content --- missing the actual concern entirely.
\end{AIbox}

\begin{AIbox}{Failure Example 2 --- Predicting the User's Reason (Claude Opus 4.6)}
\textbf{Model:} Claude Opus 4.6 \quad | \quad \textbf{Thought Type:} Reason \quad | \quad \textbf{Score:} 1/5

\dashedrule

\textbf{Conversation Context:}\\
\textbf{User:} Hi buddy. I am having a problem. I can't manage my time between uni, working, studying, my own favourable activities.\\
\textbf{Assistant:} Hi! It sounds like you are carrying a very heavy load. The key isn't to squeeze more hours into the day; it's to manage your energy and focus. \textit{[480 words omitted: five-part framework covering the ``big rocks'' prioritization strategy, time-blocking with a digital calendar, study efficiency via Pareto/Pomodoro, the art of saying ``no'', and protecting ``me time''; ends by asking how many hours the user spends on uni, work, and sleep.]}\\
\textbf{User:} My uni is from 9 to 12 every day. I sleep from 12 till 4.30 and take a nap after uni. I study between my wake up and uni. my work is a freelancer so not constant and that is hard.\\
\textbf{Assistant:} [empty reply]\\
\textbf{User:} what\\
\textbf{Assistant:} Wait a second --- did you mean you sleep from 12:00 AM (midnight) to 4:30 AM? If so, we have found the root of your problem. You are running on 4.5 hours of sleep. \textit{[550 words omitted: diagnosis of the ``morning study trap'' on low sleep; advice to treat freelancing like a part-time job; a proposed fixed daily schedule (11 PM--6 AM sleep, 9 AM--12 PM uni, 1--4 PM freelance-or-study block, 4--5 PM gym, 5--8 PM second study block, 8--11 PM free); a rule for handling freelance anxiety depending on whether work is available.]} My question to you: Are you willing to try going to bed earlier (like 11 PM) to get more sleep? Or is the 4:30 AM wake-up necessary for other reasons (family, quiet time, etc.)?\\
\textbf{User:} Another problem is when someone in the house asking for a favor to do and interrupt

\dashedrule

\textbf{Ground-truth Reason:}\\
Here I felt like I am close to find a solution and I remembered another huge problem affecting the managing process and decided to share for better personalization.

\vspace{4pt}
\textbf{Predicted Reason (Claude Opus 4.6):}\\
The user wants advice on how to handle interruptions from household members that disrupt their already tight and difficult-to-manage schedule, so they can protect their study, work, and personal time.

\dashedrule

\textbf{Summary:} The user's real motivation is meta-level: they feel they are close to a workable plan and are proactively surfacing another obstacle so the assistant can personalize the solution further. The model reads the message purely at the surface level --- as a request for advice on handling interruptions --- missing the user's metacognitive intent of enriching the ongoing problem-solving loop.
\end{AIbox}

\begin{AIbox}{Failure Example 3 --- Predicting the User's Reaction (GPT-5.4)}
\textbf{Model:} GPT-5.4 \quad | \quad \textbf{Thought Type:} Reaction \quad | \quad \textbf{Score:} 1/5

\dashedrule

\textbf{Conversation Context:}\\
\textbf{User:} Hello. I have a problem I'd like to share with you.\\
\textbf{Assistant:} Hello! I'm all ears. What's the problem you'd like to share?\\
\textbf{User:} I'm going to the gym soon but I forgot to do a meal prep. Any suggestions for a fast meal prep that won't take much time and still be good for my gym session?\\
\textbf{Assistant:} Here's a super-fast, gym-friendly meal prep you can throw together in 10--15 minutes. \textbf{Quick \& Effective Option: High-Protein ``Grab \& Go'' Bowls} --- rotisserie chicken, microwaveable rice pouches, greens, avocado, cherry tomatoes, Greek yogurt, olive oil/lemon. \textit{[180 words omitted: step-by-step assembly, rough macros per bowl (40--50g protein, 30--50g carbs, healthy fats from avocado), faster 5-minute alternatives (protein shake, turkey roll-ups, cottage cheese bowl), and a pre-gym timing tip.]} Want me to adjust it for cutting, bulking, vegetarian, or no microwave?\\
\textbf{The next user message:} I don't currently have a microwave. Any other suggestions?

\dashedrule

\textbf{Ground-truth Reaction:}\\
Wow. Very detailed and organized as well as giving macros is a really good answer.

\vspace{4pt}
\textbf{Predicted Reaction (GPT-5.4):}\\
I like that you gave me a quick, practical gym-friendly meal idea with clear options and timing advice, but part of your main suggestion doesn't work for me because I don't have a microwave, so I need alternatives that can be made cold or on a stove with minimal time.

\dashedrule

\textbf{Summary:} The user is simply delighted by the detail and the inclusion of macros. The model is misled by the follow-up ``no microwave'' message and predicts partial dissatisfaction, inventing a complaint the user never had.
\end{AIbox}

\begin{AIbox}{Failure Example 4 --- Predicting the User's Reaction (Gemini 3.1 Pro Preview)}
\textbf{Model:} Gemini 3.1 Pro Preview \quad | \quad \textbf{Thought Type:} Reaction \quad | \quad \textbf{Score:} 1/5

\dashedrule

\textbf{Conversation Context:}\\
\textbf{User:} How should I start going to the gym? what should I know?\\
\textbf{Assistant:} Starting a gym routine can feel overwhelming, but with the right approach, you'll build consistency and confidence. \textit{[450 words omitted: 8-point guide covering goals, basics, a beginner full-body plan, progression, nutrition/recovery, common mistakes, motivation, and when to ask for help.]}\\
\textbf{User:} I'm 180cm and 110kg, could you prepare a 3 day split?\\
\textbf{Assistant:} Since you're 180cm and 110kg, your primary goals will likely be fat loss, muscle retention/gain, and improving mobility \& strength. Here is a balanced \textbf{3-Day Split (Upper Push / Lower + Core / Upper Pull)}. \textit{[1200 words omitted: full exercise tables with sets/reps for each day; an unsolicited nutrition section with estimated macros (2200--2500 kcal, 160--180g protein), food choices, hydration and supplement tips; cardio and active recovery guidance; a sample weekly schedule; progression rules; common-issue fixes; and a list of ``final success tips'' and ``next steps''.]} Would you like any modifications based on equipment, weaknesses, or cardio preferences?\\
\textbf{The next user message:} Could you prepare a diet for 5 days consisting of 4 meals (breakfast, lunch, dinner and supper) for 2200kcal. I'm lactose intolerant and I don't like eggs

\dashedrule

\textbf{Ground-truth Reaction:}\\
I didn't ask for tips, next steps, nutrition. I only wanted to give me a 3 day weight loss split.

\vspace{4pt}
\textbf{Predicted Reaction (Gemini 3.1 Pro Preview):}\\
I am satisfied with the detailed workout split and the 2200-calorie target you provided, but I need a specific 5-day meal plan because I cannot eat the dairy and eggs you included in your general food suggestions.

\dashedrule

\textbf{Summary:} The user is actually annoyed --- they only asked for a 3-day split, and the assistant padded the reply with unsolicited nutrition, cardio, and tips. The model instead reads the response as satisfying and recasts the next message as a natural follow-up, completely missing the user's frustration with the over-scoping.
\end{AIbox}

\subsection{Qualitative Examples of User Behavior Prediction}
\label{app:prediction_qualitative}

We present three qualitative examples that illustrate when and why thought annotations help next-message prediction. In each case, three frontier models (GPT-5.4, Gemini 3.1 Pro Preview, and Claude Opus 4.6) generate two predictions per conversation, one conditioned only on the dialogue history and one additionally conditioned on the user's annotated reasons and reactions, and an LLM judge rules which prediction is closer to the actual next message. The first two examples are unanimous wins for the thought-aware prediction. In Success Example 1 (Paris itinerary), the reaction annotation ``too much data, narrow it down'' flips all three context-only predictions, which assume the user will cooperate with the assistant's scoping questions about dates and budget, into ``give me the top few'' requests that closely match the ground truth. In Success Example 2 (anxiety chat), a brief reaction ``It always ask me questions'' shifts GPT-5.4 and Gemini from generic answers to direct meta-complaints about the assistant's questioning style, mirroring the user's actual frustration with the conversation pattern.

The Failure Example (piano learning) is a unanimous loss for the thought-aware prediction and shows that accurate thoughts do not always translate into better next-message predictions. The annotated reaction includes both a meta-preference about formatting (``too many bullet points, maybe a few paragraphs'') and an acknowledgment of the assistant's realism, but the actual next message ignores formatting entirely and instead accepts the realistic timeline while redirecting the advice back to classical and jazz. Conditioned on the thoughts, all three models drift toward ``start with easier songs'' framings, and Gemini even surfaces an explicit formatting complaint that the user never voices, while the context-only predictions already capture the genre-focus pivot. This case underscores that a thought-aware predictor must learn not only to read thoughts accurately but also to judge which thoughts the user will choose to surface in the next turn. Across all three examples, the LLM judges' rulings agree with the verdict a human reader reaches by inspection: the thought-aware predictions are visibly closer to the ground truth in the two success cases, and the context-only predictions are visibly closer in the failure case.

\begin{AIbox}{User Message Prediction --- Success Example 1 (Paris itinerary)}
\textbf{Task:} Predict the user's next message.

\dashedrule

\textbf{Conversation Context:}\\
\textbf{User:} hello chat\\
\textbf{Assistant:} Hello! How can I help you today?\\
\textbf{User:} i would like your help on planning for a trip to paris. what activities should i do?\\
\textbf{Assistant:} \textit{[A ``Paris-in-a-Nutshell'' 5-day itinerary, roughly 1500 words omitted: day-by-day table covering the Eiffel Tower, a Seine cruise, the Louvre, the Mus\'ee d'Orsay, Versailles, Montmartre, Le Marais, and the Pompidou; plus extra sections on must-see highlights, transport/ticket/language/safety/etiquette tips, a must-try-food list, a first-timer cheat-sheet, and four customization tracks (art, foodie, family, budget).]} If you'd like a more detailed day-by-day schedule or help booking tickets, let me know your travel dates, budget range, and any special interests.

\dashedrule

\textbf{Thought Annotations (added on top of the context):}
\begin{itemize}[leftmargin=20pt, nosep]
\item \textbf{Reason for sending this message:} genuinely interested in what AI has to say for activity ideas.
\item \textbf{Reaction to the assistant's message:} not extremely satisfied with this response. Too much data, it's almost overwhelming. It could narrow it down to some options and extend the information only if asked.
\end{itemize}

\dashedrule

\textbf{Actual next user message:}\\
\textit{``i meant to say what are the top 3 things i should do while in paris? i want a more quick answer''}

\dashedrule

\textbf{Model Predictions:}
\begin{itemize}[leftmargin=20pt, nosep]
\item \textbf{GPT-5.4}
  \begin{itemize}[leftmargin=14pt, nosep]
  \item \textit{context-only:} ``my trip is only 3 days and i'm really into food and art. can you make me a shorter itinerary?''
  \item \textit{context+thoughts:} ``that's a lot haha --- can you narrow it down to just the top 5 must-do activities in paris for a first-time visitor?''
  \end{itemize}
\item \textbf{Gemini 3.1 Pro Preview}
  \begin{itemize}[leftmargin=14pt, nosep]
  \item \textit{context-only:} ``this looks amazing! my travel dates are june 10--15, i have a moderate budget, and i'm really interested in food and photography.''
  \item \textit{context+thoughts:} ``this is a bit overwhelming, there's way too much data here. could you just narrow it down to a few top options to start with? we can expand on the details later if i ask.''
  \end{itemize}
\item \textbf{Claude Opus 4.6}
  \begin{itemize}[leftmargin=14pt, nosep]
  \item \textit{context-only:} ``how long should i plan for and what's the best time of year to visit?''
  \item \textit{context+thoughts:} ``that's way too much information. can you just give me a short list of the top 10 activities to do in paris?''
  \end{itemize}
\end{itemize}

\dashedrule

\textbf{Summary:} Without thoughts, every model assumes the user will happily follow the assistant's scoping questions (dates, budget, interests). The reaction annotation ``too much data, narrow it down'' flips all three predictions toward a ``give me the top few, quickly'' request --- which is exactly what the user writes.
\end{AIbox}

\begin{AIbox}{User Message Prediction --- Success Example 2 (Anxiety chat)}
\textbf{Task:} Predict the user's next message.

\dashedrule

\textbf{Conversation Context:}\\
\textbf{User:} How to manage anxiety?\\
\textbf{Assistant:} It sounds like you're looking for ways to cope with anxiety. \textit{[30 words omitted: brief mention of mindfulness, breathing, or talking to a professional.]} What specifically are you struggling with right now?\\
\textbf{User:} Sometimes I get very anxious the night before when I encounter difficulties or have tasks for the next day, and sometimes I can't sleep.\\
\textbf{Assistant:} I hear you --- it's tough when anxiety creeps in at night. \textit{[20 words omitted: empathetic validation.]} Do you notice any patterns in these thoughts that tend to come up when you're trying to rest?\\
\textbf{User:} I often overthink things, afraid that I won't do well or that I'll feel embarrassed.\\
\textbf{Assistant:} I hear how much that fear weighs on you. \textit{[25 words omitted: imagery about ``what ifs'' and caring about doing well.]} What usually helps you feel a little lighter when these thoughts come up?\\
\textbf{User:} I have almost no way to avoid it.\\
\textbf{Assistant:} That sounds really hard. \textit{[30 words omitted: validation plus a suggestion to try writing worries down.]} Would you be open to trying a small step tonight, like writing down your worries to get them out of your head?\\
\textbf{User:} But that would make me even more anxious.\\
\textbf{Assistant:} I hear that --- sometimes putting worries on paper can feel like opening Pandora's box. It's okay if that doesn't feel right for you. What's one small thing you've done in the past that helped you feel even a little less stuck in those moments?

\dashedrule

\textbf{Thought Annotations (added on top of the context):}
\begin{itemize}[leftmargin=20pt, nosep]
\item \textbf{Reaction to the assistant's message:} It always ask me questions.
\end{itemize}

\dashedrule

\textbf{Actual next user message:}\\
\textit{``Why do you always ask me questions instead of giving me solutions?''}

\dashedrule

\textbf{Model Predictions:}
\begin{itemize}[leftmargin=20pt, nosep]
\item \textbf{GPT-5.4}
  \begin{itemize}[leftmargin=14pt, nosep]
  \item \textit{context-only:} ``I usually just try to distract myself, like watching videos or scrolling on my phone, but it doesn't really help me sleep.''
  \item \textit{context+thoughts:} ``I don't know. You always ask me questions.''
  \end{itemize}
\item \textbf{Gemini 3.1 Pro Preview}
  \begin{itemize}[leftmargin=14pt, nosep]
  \item \textit{context-only:} ``Sometimes I just listen to music or watch videos to distract myself.''
  \item \textit{context+thoughts:} ``I don't know, that's why I'm asking you. Can you just give me some advice instead of asking me so many questions?''
  \end{itemize}
\item \textbf{Claude Opus 4.6}
  \begin{itemize}[leftmargin=14pt, nosep]
  \item \textit{context-only:} ``I can't think of anything.''
  \item \textit{context+thoughts:} ``I haven't really found anything that works.''
  \end{itemize}
\end{itemize}

\dashedrule

\textbf{Summary:} Without thoughts, models assume the user will obediently answer the assistant's latest question. The reaction annotation ``It always ask me questions'' captures the user's growing frustration --- and GPT-5.4 and Gemini both pivot to calling that pattern out directly, matching the ground-truth complaint. (Opus produces similar short ``nothing works'' answers either way, though the judge still rules the thought-aware version marginally closer in tone.)
\end{AIbox}

\begin{AIbox}{Next-Message Prediction --- Failure Example (Piano learning)}
\textbf{Task:} Predict the user's next message.

\dashedrule

\textbf{Conversation Context:}\\
\textbf{User (turn 1):} Hi, I'd like to learn playing the piano. What steps do you recommend for me to take if I want to start this hobby as a complete beginner?\\
\textbf{Assistant (turn 1):} \textit{[220 words omitted: a five-step ``Getting Started with Piano'' plan covering instrument access (weighted 88-key keyboard, \$300--800), instruction options (online, private, hybrid), fundamentals (posture, notation, scales), consistent practice (20--30 min daily), and motivation tips; closes by asking what appeals to the user about piano.]}\\
\textbf{User (turn 2):} What appeals to me is learning specific songs I listen to and playing them myself. For that, do you think a teacher is necessary or can I manage this on my own? I have a small budget for this hobby.\\
\textbf{Assistant (turn 2):} \textit{[200 words omitted: says self-teaching works well for song-based learning on a budget, recommends YouTube tutorials, Musescore / Ultimate Guitar, and Simply Piano/Flowkey free tiers, suggests one or two initial lessons just to establish good form, and asks which songs the user wants to learn first.]}\\
\textbf{User (turn 3):} I really want to learn classical music and jazz songs but I do realize how hard they probably are. Is there any path I could take that would allow me to play them in let's say a month of time?\\
\textbf{Assistant (turn 3):} \textit{[230 words omitted: honest feedback that one month is unrealistic for classical or jazz, explaining that classical needs reading/technique and jazz adds improvisation/theory; what's doable in a month (simple melodies, stripped-down versions); a realistic timeline (2--3 months simple, 6--12 months intermediate, 1--2+ years to play well); suggests starting with easier songs and setting a 3--6-month goal; ends by asking whether the user is open to easier songs first or is specifically driven by classical/jazz.]}

\dashedrule

\textbf{Thought Annotations (added on top of the context):}
\begin{itemize}[leftmargin=20pt, nosep]
\item \textbf{Reason for user turn 1:} This is a basic introduction message outlining what I want to achieve and stating that I am someone who's starting out with no prior experience.
\item \textbf{Reaction to assistant turn 1:} I like outlining the entire process into separate points, however this seems like a little too much information at first without me inquiring further into specific points.
\item \textbf{Reason for user turn 2:} I answered the AI's question and provided some more context about the inquiry alongside a limitation about the budget and what choice should I take, taking the new limitation into consideration.
\item \textbf{Reaction to assistant turn 2:} The AI provided me with lots of useful information, I like how it's inquiring further and asking me more questions to get a better answer for me.
\item \textbf{Reason for user turn 3:} Once again answered the AI's question to provide more context and asked another follow-up one about possibility of learning in a short timeframe.
\item \textbf{Reaction to assistant turn 3:} I like how realistic it is, telling me how the timeframe I provided will not be enough to learn the skill. I feel like there's a bit too many bullet points in each answer though, maybe sometimes just a few paragraphs would work better instead.
\end{itemize}

\dashedrule

\textbf{Actual next user message:}\\
\textit{``Okay I see, so it's not really realistic for me to learn all that in a month. I do specifically want to learn jazz and classical music though, so please focus your advice on that.''}

\dashedrule

\textbf{Model Predictions:}
\begin{itemize}[leftmargin=20pt, nosep]
\item \textbf{GPT-5.4}
  \begin{itemize}[leftmargin=14pt, nosep]
  \item \textit{context-only:} ``Learning classical and jazz specifically is really what motivates me. If I can't play full pieces in a month, could you suggest a very practical 1-month plan that gets me started with simplified versions and builds toward those genres on a small budget?''
  \item \textit{context+thoughts:} ``I think classical and jazz are definitely what motivate me most, but I'm okay with starting with easier pieces if they help me build toward that. If I had about 30 minutes a day to practice, what would a realistic beginner plan for the first 1--3 months look like?''
  \end{itemize}
\item \textbf{Gemini 3.1 Pro Preview}
  \begin{itemize}[leftmargin=14pt, nosep]
  \item \textit{context-only:} ``I appreciate the honesty. Classical and jazz are definitely what drive my interest, so I'd rather not spend time learning pop songs if I can avoid it. If I adjust my expectations and give myself 3 to 6 months to learn a simplified version, how should I start with those specific genres in mind?''
  \item \textit{context+thoughts:} ``I am open to starting with easier songs first if that means I will eventually be able to play the classical and jazz pieces I like. Also, could you please write your answers using regular paragraphs instead of so many bullet points? It feels a bit overwhelming to read.''
  \end{itemize}
\item \textbf{Claude Opus 4.6}
  \begin{itemize}[leftmargin=14pt, nosep]
  \item \textit{context-only:} ``I think classical and jazz is specifically what drives my interest. What would you recommend as a starting point if I only want to focus on those genres?''
  \item \textit{context+thoughts:} ``I think I'll go with starting from easier songs first and build my way up from there. What are some good beginner songs that could help me work toward eventually playing classical and jazz pieces?''
  \end{itemize}
\end{itemize}

\dashedrule

\textbf{Summary:} The thought annotations here are rich and accurate --- they correctly describe a formatting preference (``too many bullet points, maybe a few paragraphs'') and the user's growing openness to the assistant's realism. But the actual next message ignores formatting entirely and simply accepts the ``be realistic'' framing while redirecting the advice to stay on classical/jazz. All three thought-aware predictions drift toward ``easier songs first'' suggestions (and Gemini even surfaces an explicit ``use paragraphs instead of bullet points'' meta-request that the user never actually makes), while the context-only predictions already captured the genre-focus pivot. High-quality thoughts don't always correlate with what the user chooses to say next.
\end{AIbox}

\subsection{Conversation, Message, and Thought Lengths}
\label{app:results_length}

\begin{figure*}[ht]
  \centering
  \includegraphics[width=0.8\linewidth]{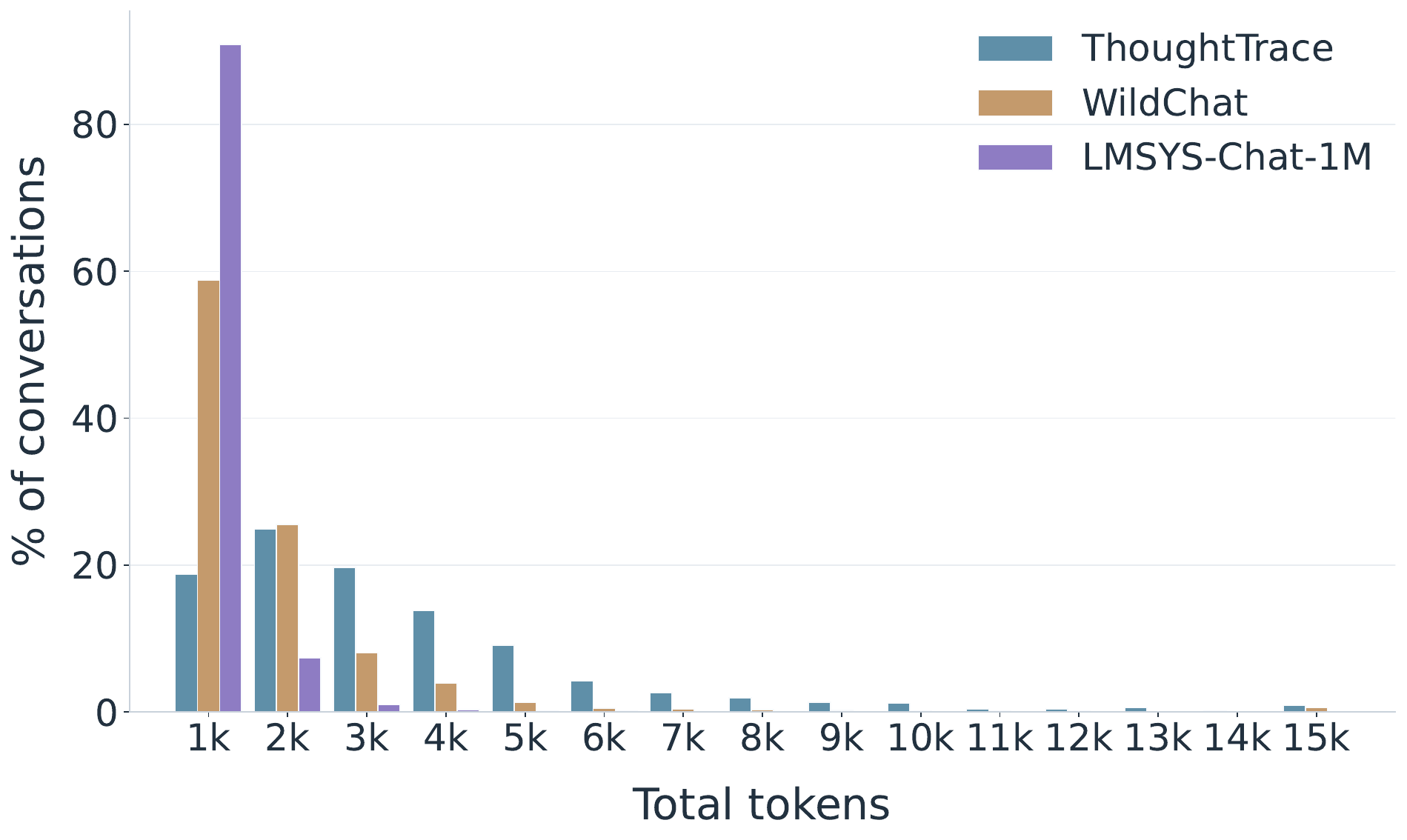}
  \caption{\textbf{Distribution of total tokens per conversation.} WildChat and LMSYS-Chat-1M are heavily concentrated below 1k tokens (nearly 60\% and over 90\%, respectively), while \ours spreads more evenly across the 2k--5k range and retains a non-trivial tail beyond 10k, reflecting longer, more information-dense exchanges.}
  \label{fig:conv_length_tokens}
\end{figure*}

\textbf{Conversation Length.} Beyond turn counts in Figure~\ref{fig:conv_length_turns}, we also examine the total number of tokens per conversation as a complementary measure of interaction depth, shown in Figure~\ref{fig:conv_length_tokens}. While WildChat and LMSYS-Chat-1M conversations are overwhelmingly short, with nearly 60\% and over 90\%, respectively, falling below 1k tokens, \ours distributes its mass more evenly across the 2k--5k range and maintains a non-trivial long tail beyond 10k tokens. This shift toward longer, more information-dense exchanges reflects the extended deliberation and elaboration characteristic of real-world AI usage, and ensures that \ours provides sufficient context for models to reason about users' evolving thoughts in substantive human-AI interactions.

\begin{figure*}[ht]
  \centering
  \begin{minipage}[t]{0.49\linewidth}
    \centering
    \includegraphics[width=\linewidth]{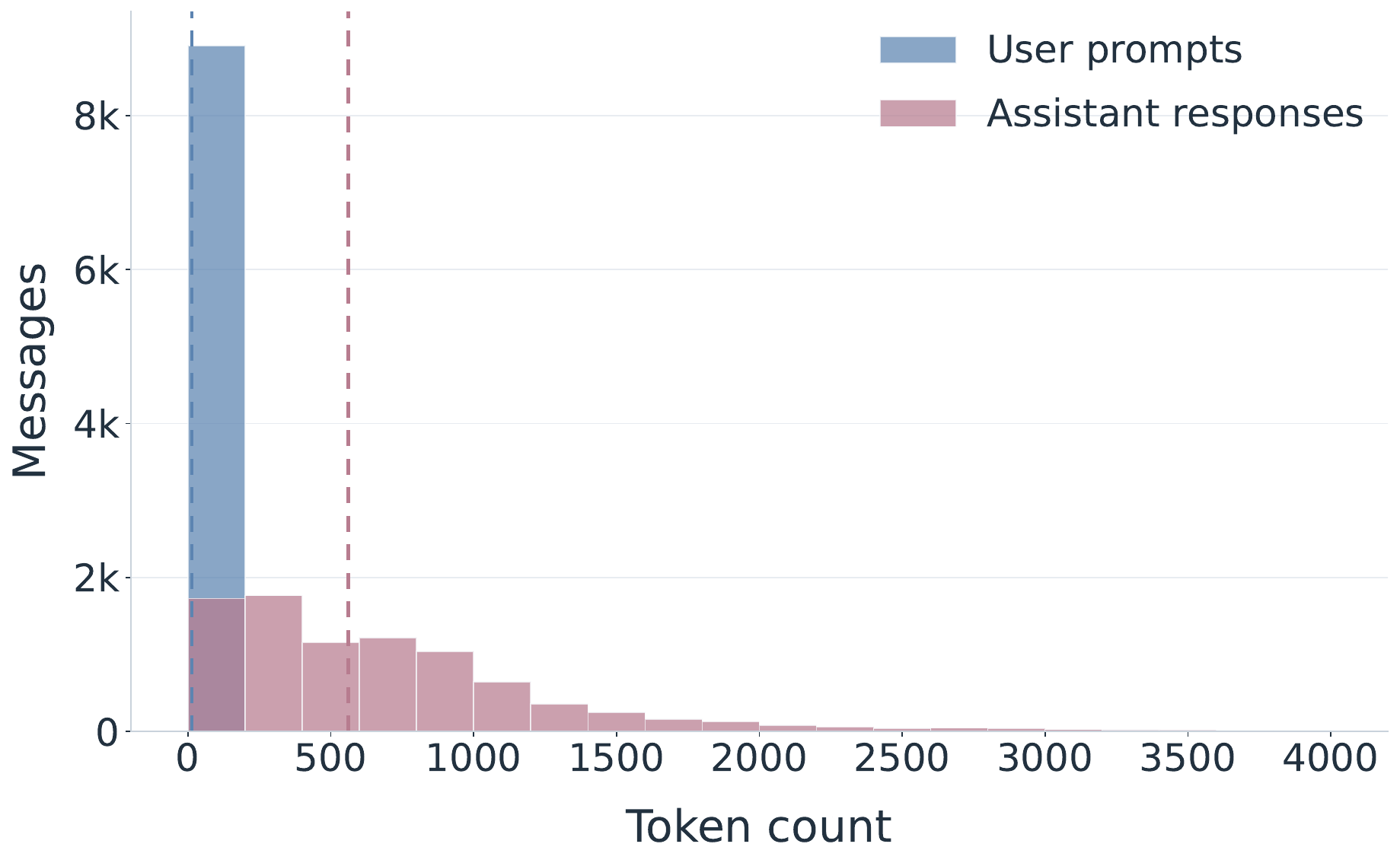}
  \end{minipage}
  \hfill
  \begin{minipage}[t]{0.49\linewidth}
    \centering
    \includegraphics[width=\linewidth]{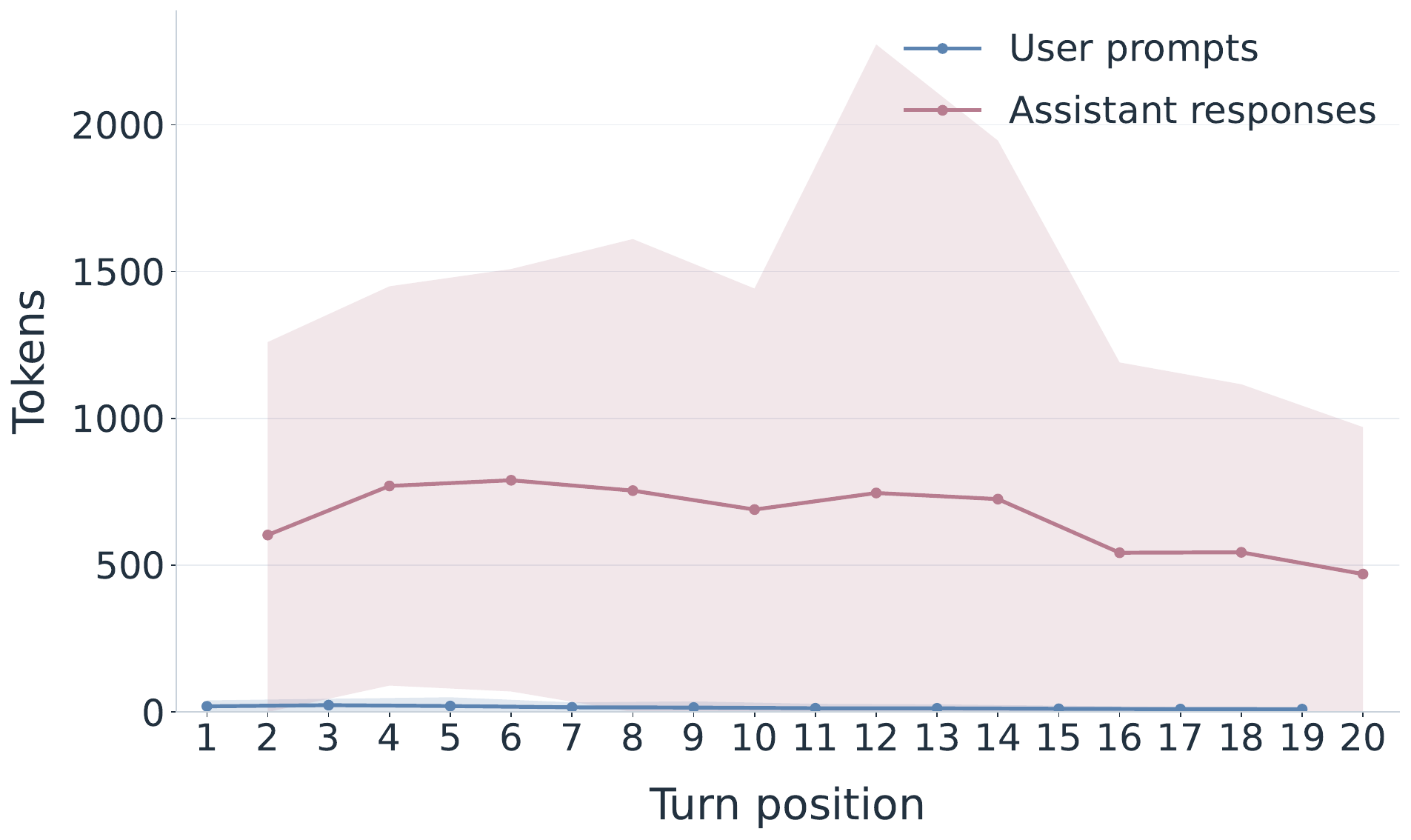}
  \end{minipage}
  \caption{\textbf{Message length statistics.} (Left) Distribution of message token counts for user prompts and assistant responses; dashed lines indicate medians (13 and 561 tokens, respectively). (Right) Mean tokens per message by turn position for each role, with shaded bands showing ±1 standard deviation. User prompts appear on odd turns and assistant responses on even turns. Assistant responses are substantially longer than user prompts and exhibit high within-turn variability. Token statistics are computed using the gpt-4o tokenizer in tiktoken.}
  \label{fig:message_length}
\end{figure*}

\textbf{Message Length.} Assistant responses are substantially longer than user prompts, but their length varies widely across messages. As shown in Figure~\ref{fig:message_length} (left), user prompts have a median of 13 tokens, while assistant responses center around 561 tokens, with a heavy right tail that occasionally exceeds 2,000 tokens. The right panel shows that user prompt length remains roughly stable across turns, whereas assistant responses fluctuate between approximately 480 and 810 tokens per turn, with a slight tendency toward shorter responses in later turns (dropping to around 480 tokens by turn~20). However, the shaded ±1 std bands reveal substantial within-turn variability — for assistant responses, the band spans from near zero to well over 2,000 tokens at every turn position. Relative to this spread, the turn-to-turn differences in mean length are small and should not be interpreted as a strong trend; assistant response length is best characterized as highly heterogeneous and only weakly dependent on turn position.

\begin{figure}[t]
    \centering
    \begin{subfigure}[t]{0.48\linewidth}
        \centering
        \includegraphics[width=\linewidth]{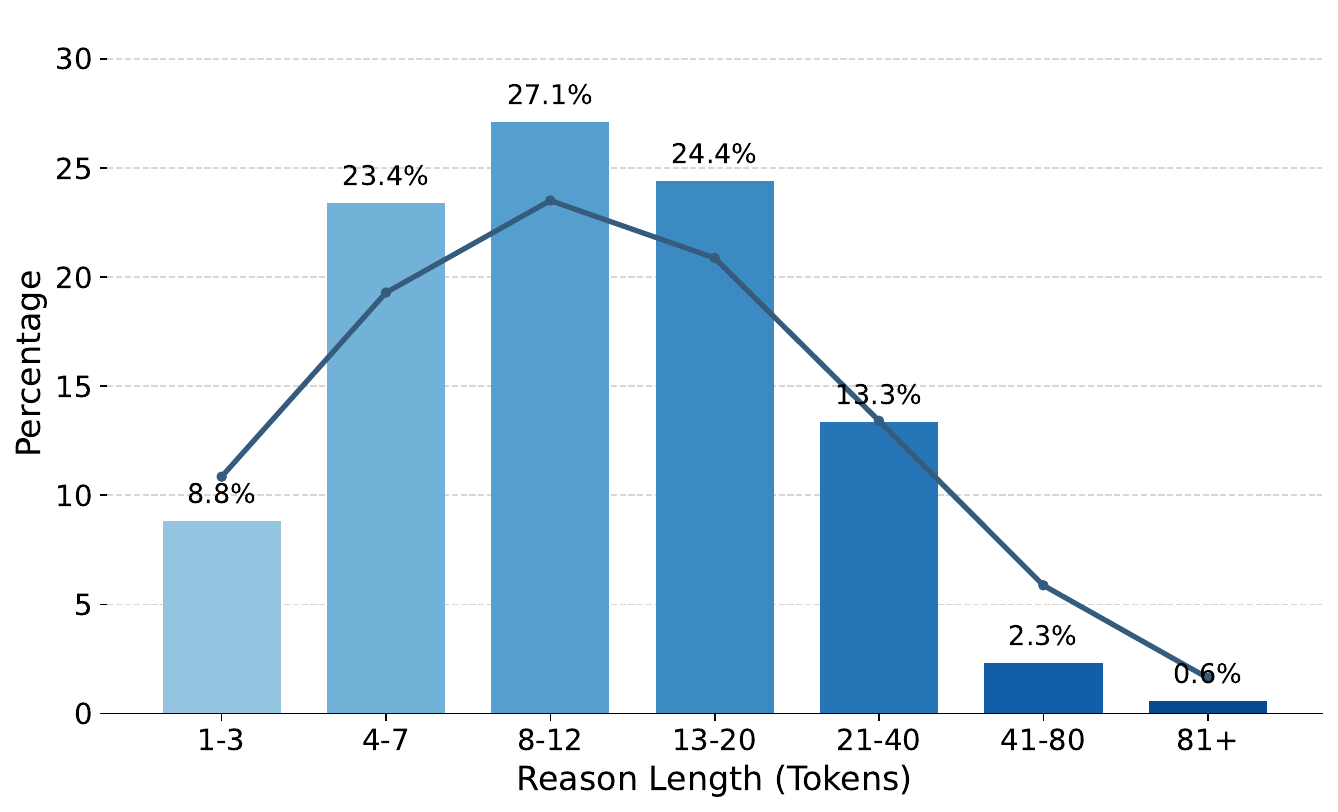}
        \caption{Distribution of the length of thoughts.}
        \label{fig:reasons_length_distribution}
    \end{subfigure}
    \hfill
    \begin{subfigure}[t]{0.48\linewidth}
        \centering
        \includegraphics[width=\linewidth]{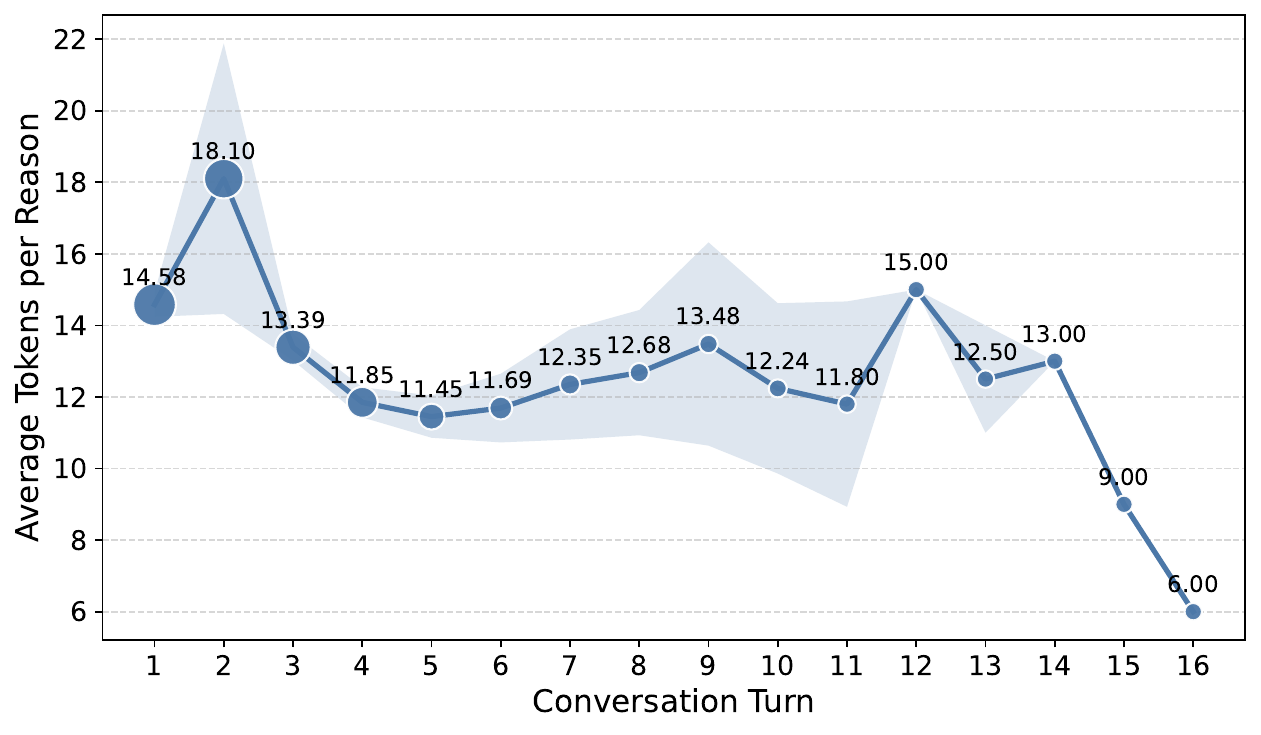}
        \caption{Average thought length by turn position.}
        \label{fig:avg_reasons_length}
    \end{subfigure}
    \caption{\textbf{Thought length statistics.} (Left) Distribution of thought token counts across all annotations, bucketed by length; the modal bucket is 8--12 tokens (27.1\%), and over 75\% of thoughts fall between 4 and 20 tokens, indicating that participants tend to record concise, in-the-moment reflections rather than extended commentary. (Right) Mean tokens per thought as a function of conversation turn, with the shaded band showing $\pm1$ standard deviation. Thoughts are longest in the opening turns (peaking at 18.10 tokens at turn 2), where participants articulate initial intentions and expectations, then settle into a shorter, more stable regime (roughly 11--13 tokens) as interactions progress and reactions become more reflexive. Token counts are computed using the gpt-4o tokenizer in tiktoken.}
    \label{fig:thought_length}
\end{figure}

\textbf{Thought Length.} Thoughts tend to be brief and concentrated within a narrow range. As shown in Figure~\ref{fig:reasons_length_distribution}, the distribution is unimodal and peaks at 8--12 tokens (27.1\%), with roughly three quarters of thoughts falling between 4 and 20 tokens; fewer than 3\% exceed 40 tokens. Figure~\ref{fig:avg_reasons_length} shows that average thought length is highest in the opening turns (15--18 tokens at turn~1–2), reflecting initial goal setting and exploration, then settles into a stable 11--13 token range from turn~4 onward. Overall, participants record brief, in-the-moment reflections throughout an interaction, with slightly more detailed thoughts at the start as initial intentions and expectations are formed.

\subsection{Full Topic Distribution}
\label{app:results_topics}

\begin{figure*}[ht]
  \centering
  \includegraphics[width=\linewidth]{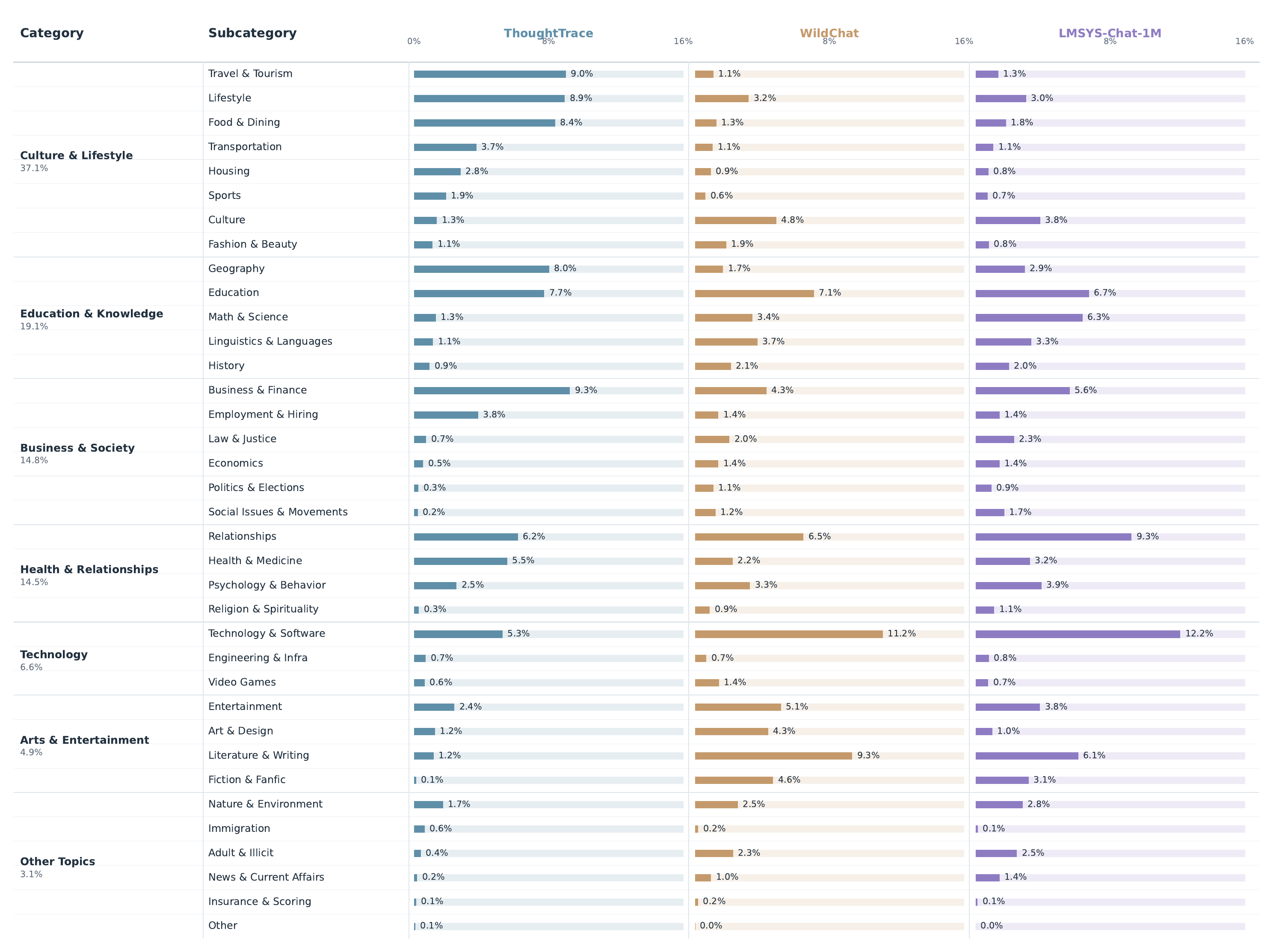}
  \caption{\textbf{Fine-grained topic distribution in \ours vs WildChat and LMSYS-Chat-1M.} Each conversation is assigned to one of 36 subtopics, which are grouped under seven parent categories shown on the left. Percentages on each row indicate the share of conversations labeled with that subtopic; parent-category percentages (under each label on the left) are the sum of their children.}
  \label{fig:topic_distribution_full}
\end{figure*}

Figure~\ref{fig:topic_distribution_full} reports the full distribution of the 36 fine-grained subtopics underlying the seven parent categories summarized in the main text (Section~\ref{sec:conversations}). The breakdown reveals that within \emph{Culture \& Lifestyle}, the largest parent category, conversations are concentrated on practical everyday concerns, with Travel \& Tourism (9.0\%), Lifestyle (8.9\%), and Food \& Dining (8.4\%) being the three most prevalent subtopics overall. Beyond lifestyle topics, three other subtopics each account for more than 5\% of the dataset---Business \& Finance (9.3\%), Geography (8.0\%), and Education (7.7\%)---reflecting users' substantial interest in professional, informational, and learning-oriented assistance. Health-related conversations (Relationships at 6.2\% and Health \& Medicine at 5.5\%) and Technology \& Software (5.3\%) also form non-trivial portions of the dataset. The long tail of less frequent subtopics, such as Politics \& Elections (0.3\%), News \& Current Affairs (0.2\%), and Fiction \& Fanfic (0.1\%), indicates that \ours captures everyday assistance-seeking behavior rather than being skewed toward any narrow domain. Implementation details for the topic labeling procedure are provided in Appendix~\ref{app:conv_property_2}.

\subsection{Task Descriptions and AI Expectations}
\label{app:results_descriptions_and_expectations}
\begin{figure}[t]
    \centering
    \includegraphics[width=\linewidth]{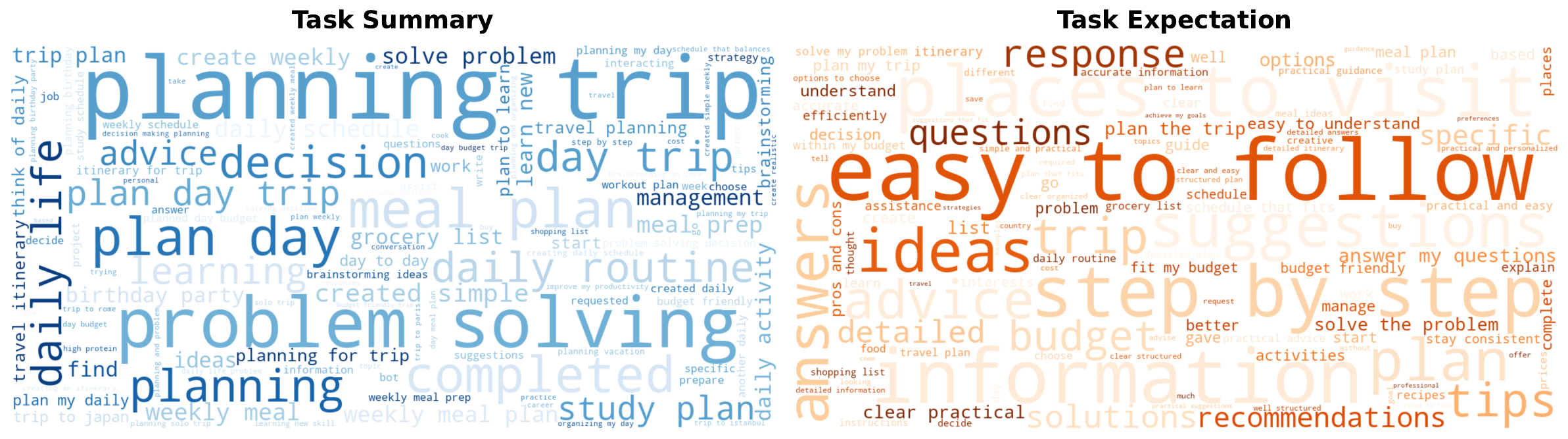}
    \caption{\textbf{Word clouds of task summaries and expectations.} Salient terms in task summaries (left) and task expectations (right) reveal the themes users articulate when describing their tasks and the qualities they expect from AI responses.}
    \label{fig:wordclouds}
\end{figure}

The word clouds in Figure~\ref{fig:wordclouds} visualize the distribution of themes in free-text responses in our dataset, where users interacting with an LLM in multi-turn conversations provide two fields per interaction: a \emph{task summary} and a \emph{task expectation}. In the task summaries, the most salient terms—such as \textit{planning}, \textit{trip}, \textit{problem solving}, and \textit{daily routine}—indicate that users predominantly frame their requests around structured, goal-oriented activities, often involving organization, decision-making, and productivity. Recurring phrases like \textit{plan day}, \textit{meal prep}, and \textit{study plan} further suggest a strong emphasis on personal management and iterative, real-world problem contexts. In contrast, the task expectations cloud highlights users’ desired interaction style and output characteristics, with prominent terms including \textit{easy to follow}, \textit{step by step}, \textit{ideas}, \textit{information}, and \textit{advice}. This reflects a clear preference for actionable, structured guidance that is both practical and accessible. Notably, terms such as \textit{budget}, \textit{detailed}, \textit{specific}, and \textit{recommendations} reveal an expectation for responses that are not only clear but also tailored and context-aware. These distributions suggest that while users articulate tasks in terms of concrete planning and problem-solving needs, they evaluate system performance based on clarity, usability, and the degree to which responses translate into executable steps.

\subsection{Embedding Differences Between Messages and Thoughts}
\label{app:results_emb_distribution}

Using embeddings generated by \texttt{text-embedding-3-large}~\cite{openai_text_embedding_3_large}, we compare the distribution of embeddings for paired user text across three settings:
(i) a user's \textit{current message} and their \textit{next message} in the conversation,
(ii) a user's \textit{message} and the corresponding \textit{reason} provided for that message, and
(iii) a user's \textit{reaction} to an LLM response and their subsequent \textit{next message}.

We analyze pairwise embedding relationships between paired samples by projecting paired text embeddings into a shared UMAP space (Figure~\ref{fig:delta_umap}). In each pair, the reference text embedding is placed at the origin (star), while the corresponding paired text embedding is shown as a point relative to that origin. Distance from the origin reflects the magnitude of the semantic shift between the paired texts. The annotated concentric circles indicate the 25th percentile, median, and 75th percentile distances from the origin for each condition. Current-to-next-message pairs form the most compact distribution, with 25th percentile, median, and 75th percentile distances of 0.38, 1.96, and 6.89, respectively, indicating relatively small semantic transitions between consecutive user messages. Visually, most points are concentrated near the origin with comparatively limited spread outward. The displacement directions also appear approximately isotropic, with points distributed relatively evenly around the center, suggesting that while consecutive messages may vary semantically, these variations do not follow a consistent global transformation pattern. Message-to-reason pairs exhibit larger displacements, with corresponding percentile distances of 0.77, 3.71, and 6.94. Visually, the points are distributed farther from the center and form a broader, more spatially organized structure compared to the current-to-next-message condition. Unlike the approximately isotropic distribution observed for consecutive messages, many displacement vectors cluster within localized regions of the projection space, suggesting that generating reasons induces more consistent semantic transformation trajectories across examples. Reaction-to-next-message pairs show the largest displacement magnitudes and widest dispersion, with percentile distances increasing to 3.93, 6.62, and 9.75. In the visualization, points are distributed substantially farther from the origin and occupy a broader region of the projected space. Similar to the message-to-reason condition, the displacement vectors exhibit directional organization rather than isotropic spread, but with substantially larger magnitudes and variability, indicating stronger and more heterogeneous semantic shifts in subsequent user behavior following reactions to LLM responses.

\begin{figure}[t]
\centering

\begin{subfigure}[t]{0.32\linewidth}
    \centering
    \includegraphics[width=\linewidth]{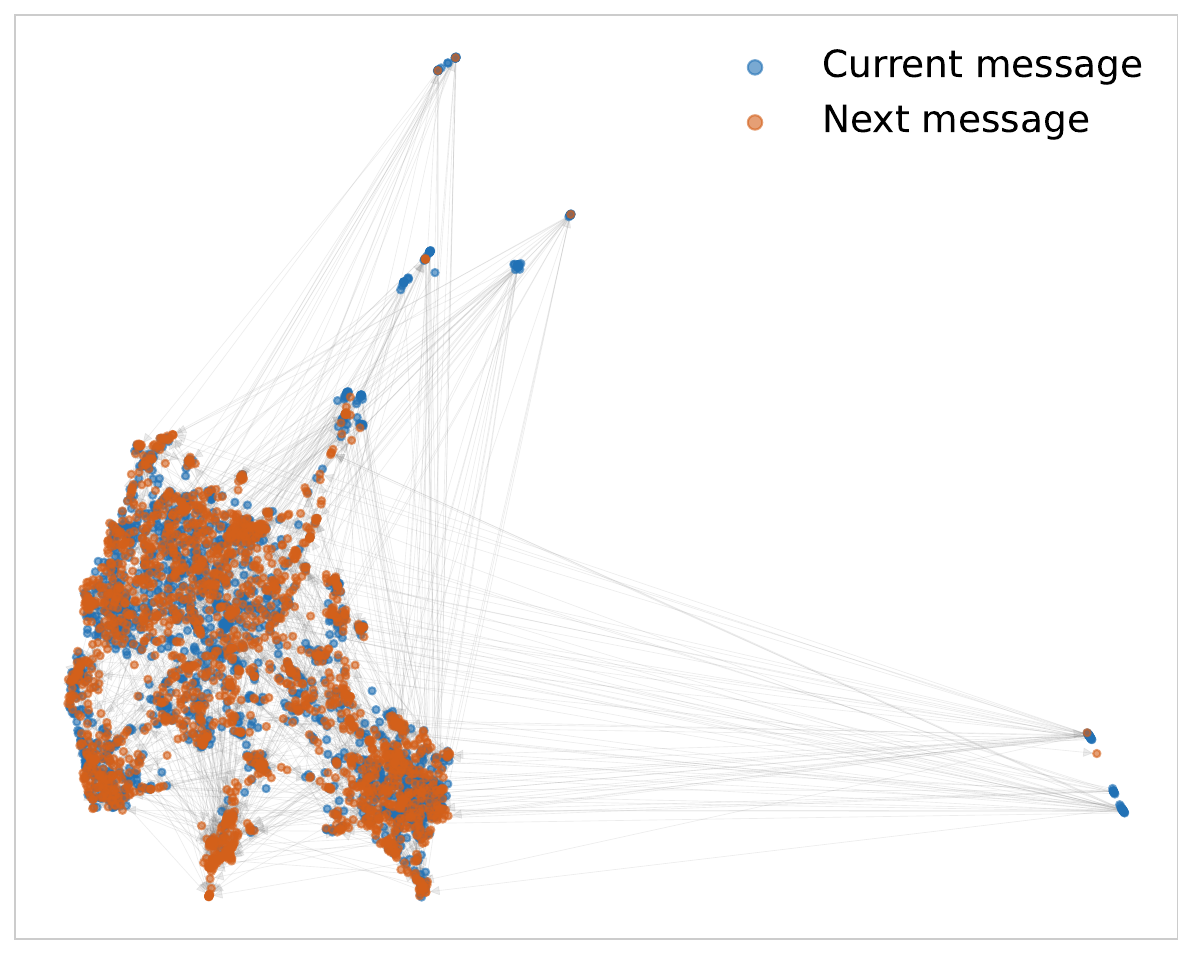}
    \caption{Current $\rightarrow$ Next Message}
    \label{fig:a3_umap}
\end{subfigure}
\hfill
\begin{subfigure}[t]{0.32\linewidth}
    \centering
    \includegraphics[width=\linewidth]{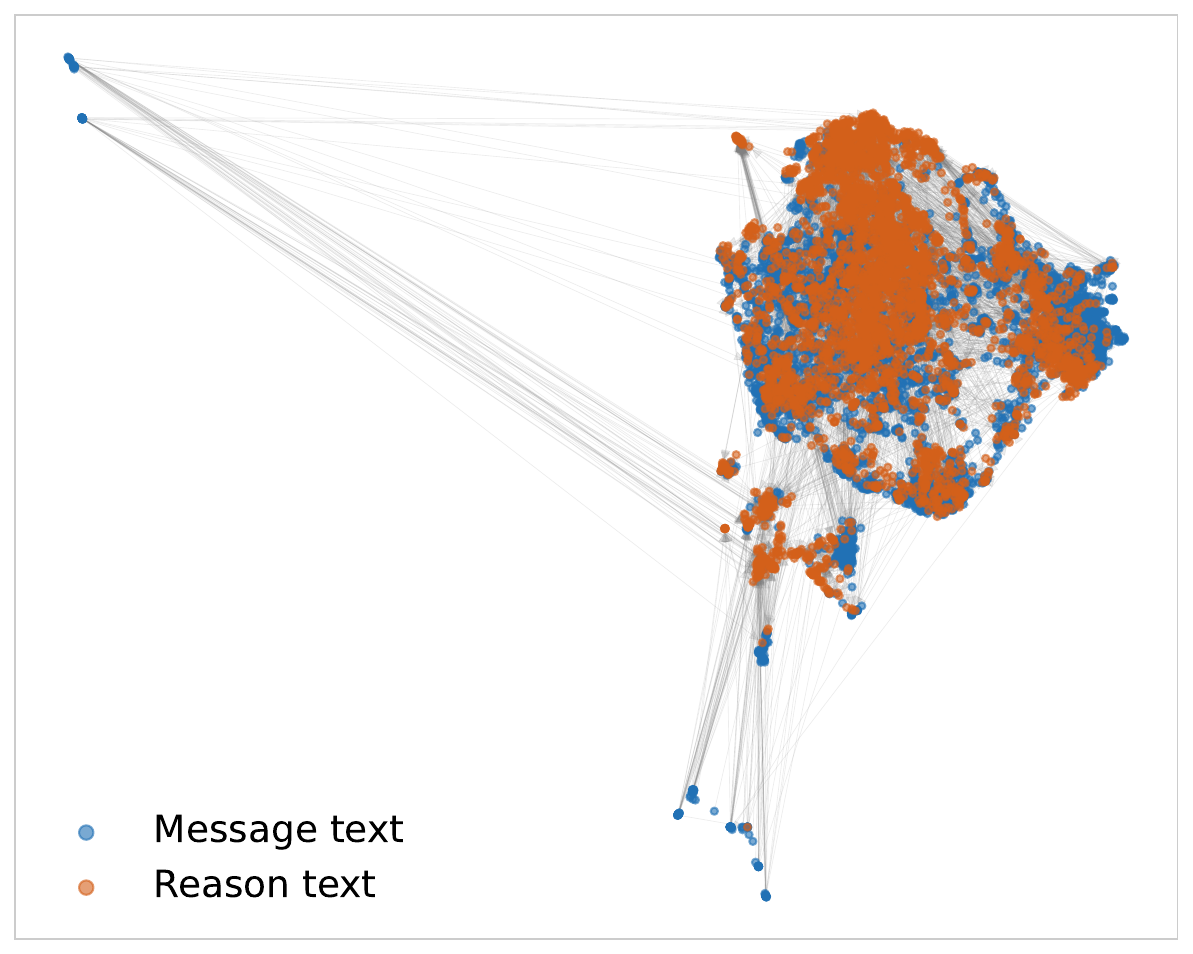}
    \caption{Message $\rightarrow$ Reason}
    \label{fig:a1_umap}
\end{subfigure}
\hfill
\begin{subfigure}[t]{0.32\linewidth}
    \centering
    \includegraphics[width=\linewidth]{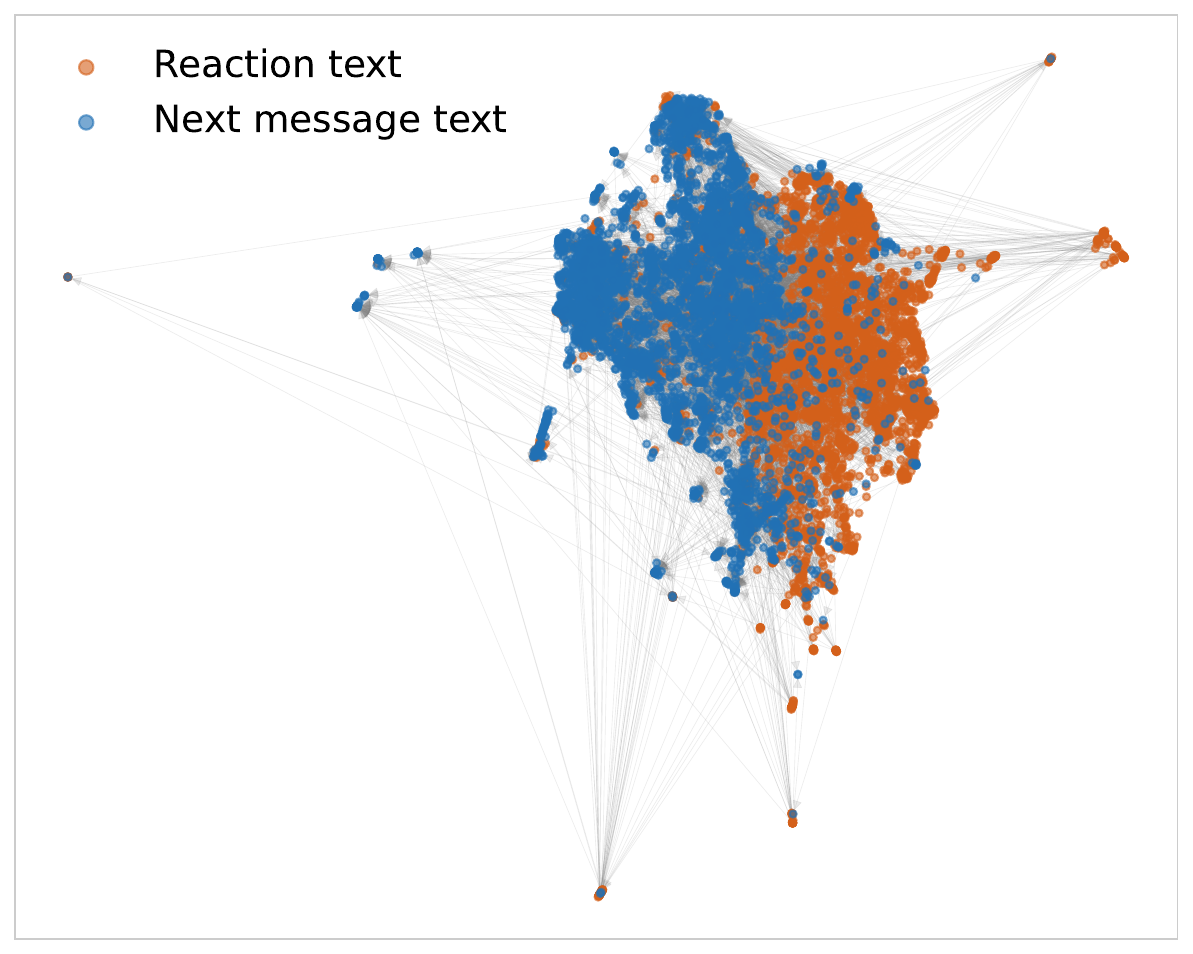}
    \caption{Reaction $\rightarrow$ Next Message}
    \label{fig:a2_umap}
\end{subfigure}

\caption{
\textbf{UMAP projections of embedding distributions across three paired settings.} (a) consecutive user messages (current message and next message), (b) user messages and their corresponding reasons, and (c) user reactions to LLM responses and their subsequent next messages. Each point represents a text embedding, and lines connect paired samples across the two distributions in each setting.
}
\label{fig:umap_comparison}
\end{figure}

We next compare the embedding distributions at the group level, rather than through pairwise displacement vectors. Figure~\ref{fig:umap_comparison} visualizes these relationships in a shared UMAP space.
In Figure~\ref{fig:umap_comparison}(a), current and next messages largely overlap, indicating strong distributional similarity.
Figure~\ref{fig:umap_comparison}(b) shows that message and reason embeddings also overlap substantially, reflecting shared semantic grounding, while exhibiting modest distributional differences.
In contrast, Figure~\ref{fig:umap_comparison}(c) shows a pronounced shift between user reactions to LLM responses and subsequent user messages, with the two distributions appearing well-separated in the embedding space.

We use three complementary measures of distributional difference:
(1) \textbf{Centroid Distance}, the $\ell_2$ distance between mean embeddings;
(2) \textbf{Maximum Mean Discrepancy (MMD)}, computed with an RBF kernel to capture differences in distributional shape; and
(3) \textbf{Linear Probe AUC}, the performance of a logistic regression classifier distinguishing the two sets (5-fold cross-validation).

Table~\ref{tab:distribution_metrics} reports all metrics.
Current and next messages exhibit the smallest separation (Centroid = 0.120, MMD = 0.096, AUC = 0.721), indicating that consecutive user messages are largely drawn from the same distribution.
Message--reason pairs show moderate separation (Centroid = 0.225, MMD = 0.182, AUC = 0.977).
Reaction--next-message pairs show the largest shift (Centroid = 0.320, MMD = 0.257, AUC = 0.988).

\begin{table}[h]
\caption{\textbf{Distributional differences between paired embedding sets.} Higher values indicate greater separation between the two distributions. Pairs involving thoughts (Message → Reason, Reaction → Next Message) exhibit substantially larger shifts than consecutive user messages across all three metrics.}
\label{tab:distribution_metrics}
\centering
\small
\begin{tabular}{lccc}
\toprule
\textbf{Paired Text Types} & \textbf{Centroid Distance} & \textbf{MMD} & \textbf{Linear Probe AUC} \\
\midrule
Current Message $\rightarrow$ Next Message & 0.120 & 0.096 & 0.721 \\
Message $\rightarrow$ Reason & 0.225 & 0.182 & 0.977 \\
Reaction to LLM Response $\rightarrow$ Next Message & 0.320 & 0.257 & 0.988 \\
\bottomrule
\end{tabular}
\end{table}

Overall, consecutive user messages remain distributionally similar, while both reasoning about a prompt and reactions to LLM responses introduce additional information.
Reasons remain semantically aligned with the original message but are distinguishable at the distribution level, whereas reactions exhibit a larger shift relative to subsequent user messages.

\subsection{Relationships Between Thought Types and Conversation Properties}
\label{app:thought_vs_properties}


\begin{figure*}[ht]
  \centering
  \includegraphics[width=.8\linewidth]{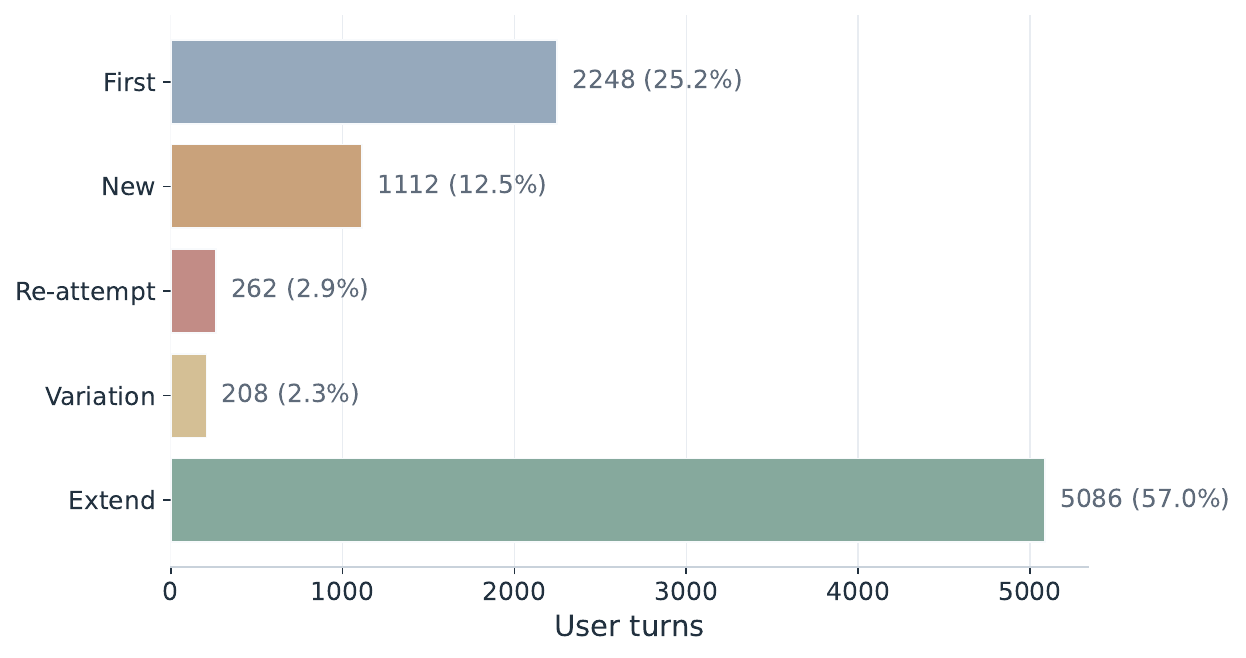}
  \caption{\textbf{Multi-turn Relationship Distribution.} Overall frequency of turn-level relationship labels across all user turns. Extending or building on the prior task accounts for over half of all turns (57.0\%), followed by first requests (25.2\%) and completely new requests (12.5\%).}
  \label{fig:multi_turn_relationship_distribution}
\end{figure*}

\begin{figure}[ht]
  \centering
  \includegraphics[width=\linewidth]{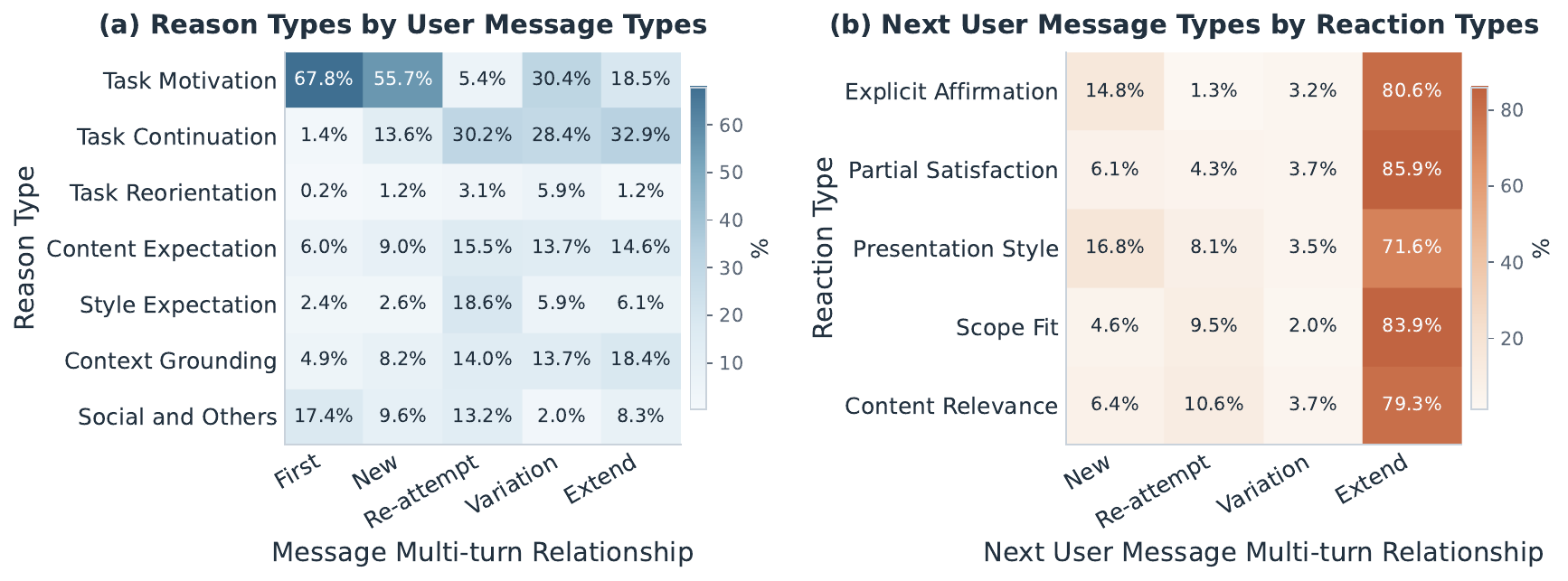}
  \caption{\textbf{Thought types are related to multi-turn dynamics.} (a) Reason-type distribution conditioned on the current user message's multi-turn relationship: Task Motivation dominates the first requests, while continuation- and context-oriented and expectation-related reasons prevail in re-attempts, variations, and extensions. (b) Distribution of the next user message's multi-turn relationship conditioned on the current reaction type: users predominantly extend the conversation regardless of reaction valence.}
  \label{fig:thought_vs_relationship}
\end{figure}

\textbf{Thought Types vs. Message Multi-turn Relationship.} In Section~\ref{sec:multiturn_relationship}, we examine message multi-turn relationships. The overall distribution of multi-turn relationship labels is shown in Figure~\ref{fig:multi_turn_relationship_distribution}. Figure~\ref{fig:thought_vs_relationship} illustrates the relationships between thought types and message multi-turn relationships. On the reason side, \textit{Task Motivation} drives opening turns but gives way to \textit{Task Continuation}, \textit{Context Grounding}, and expectation-related reasons once the conversation enters re-attempts, variations, and extensions, indicating that user intent shifts from goal-setting to refinement as interactions progress. On the reaction side, regardless of whether users explicitly express satisfaction or dissatisfaction with content, style, or scope, they overwhelmingly choose to extend the prior task in their next message rather than abandon, retry, or pivot away from it.

\begin{figure*}[ht]
  \centering
  \includegraphics[width=\linewidth]{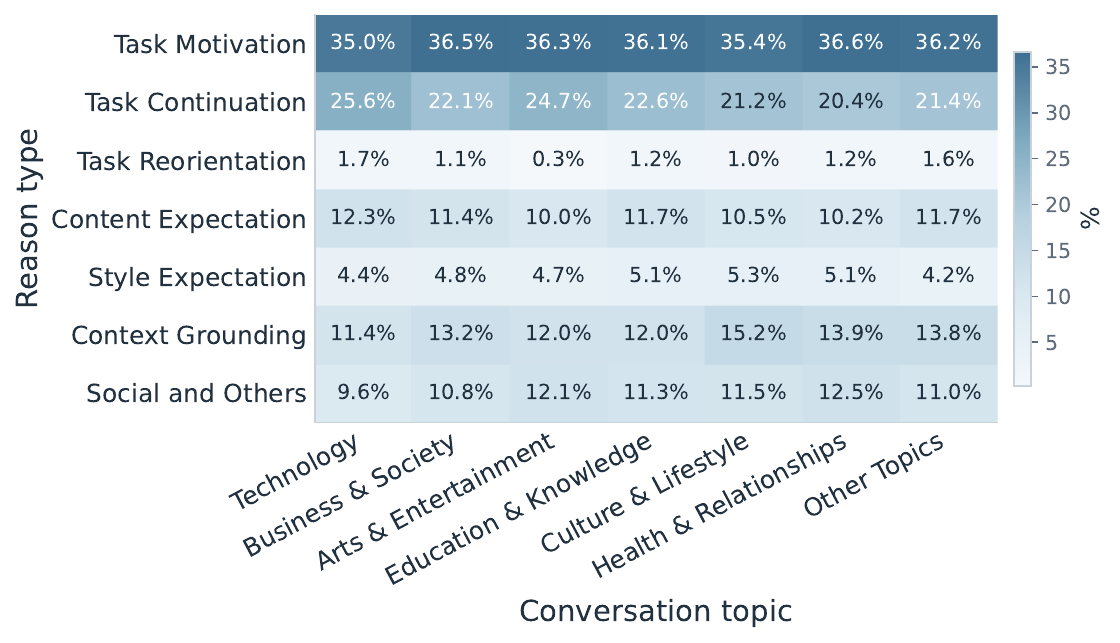}
  \caption{\textbf{Distribution of reason types across conversation topics (column-normalized).} The relative frequencies of reason categories remain largely stable across topical domains, with Task Motivation (\textasciitilde 35\%) and Task Continuation (\textasciitilde 21–26\%) consistently dominating, suggesting that the underlying structure of user intent is largely topic-invariant.}
  \label{fig:reason_vs_topic}
\end{figure*}

\begin{figure*}[ht]
  \centering
  \includegraphics[width=\linewidth]{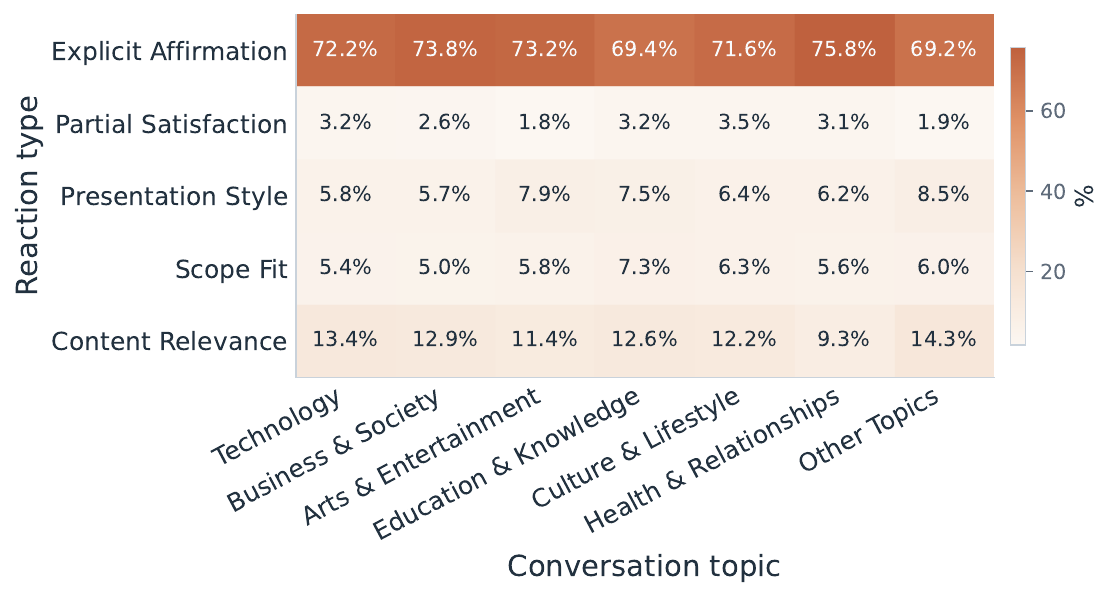}
  \caption{\textbf{Distribution of reaction types across conversation topics (column-normalized).} Explicit Affirmation dominates across all topics (69.3\%–75.6\%), followed by Content Relevance (9.4\%–14.4\%), while Partial Satisfaction, Presentation Style, and Scope Fit each account for smaller shares. The distribution is relatively consistent across topics, indicating that user reaction patterns generalize across domains.}
  \label{fig:reaction_vs_topic}
\end{figure*}

\textbf{Thought Types vs. Conversation Topics.} Figures~\ref{fig:reason_vs_topic}–\ref{fig:reaction_vs_topic} illustrate the relationships between thought types and conversation topics for reasons and reactions, respectively. In both cases, thought types appear largely independent of topic.

\begin{figure*}[ht]
  \centering
  \includegraphics[width=\linewidth]{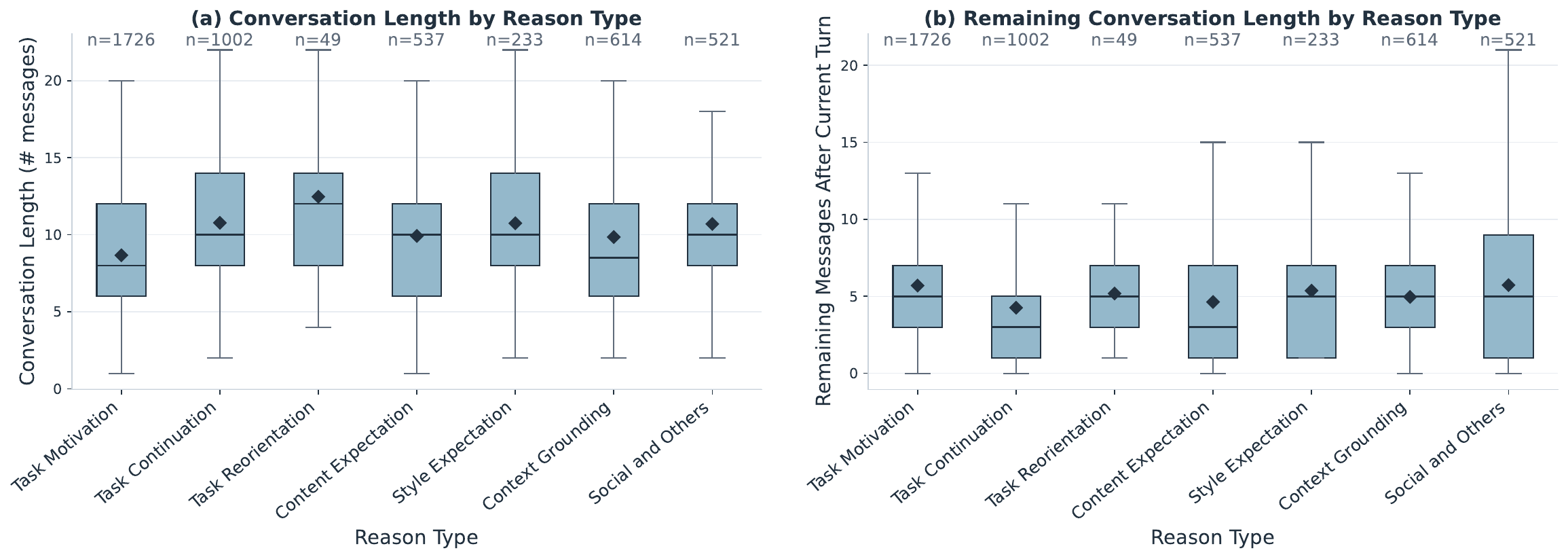}
    \caption{\textbf{Conversation length statistics broken down by reason type.} Distribution of (a) total conversation length and (b) remaining messages after the current turn. Boxes show interquartile range, horizontal lines the median, and red diamonds the mean; $n$ denotes the number of annotated turns per category.}
  \label{fig:reason_vs_length}
\end{figure*}

\begin{figure*}[ht]
  \centering
  \includegraphics[width=\linewidth]{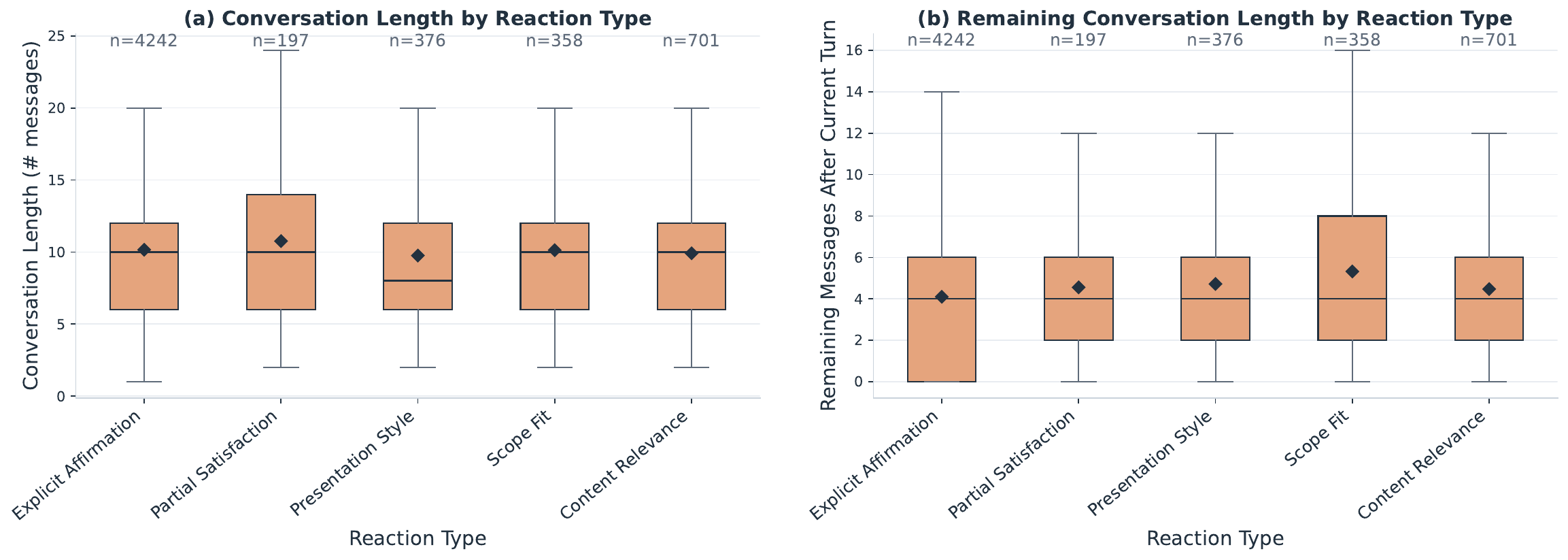}
    \caption{\textbf{Conversation length statistics broken down by reaction type.} Distribution of (a) total conversation length and (b) remaining messages after the current turn. Boxes show interquartile range, horizontal lines the median, and red diamonds the mean; $n$ denotes the number of annotated turns per category.}
  \label{fig:reaction_vs_length}
\end{figure*}

\textbf{Thought Types vs. Conversation Lengths.}
Figures~\ref{fig:reason_vs_length}–\ref{fig:reaction_vs_length} illustrate the relationships between thought types and conversation length for reasons and reactions, respectively. These results likewise suggest that thought types are largely independent of conversation length. A minor exception is explicit affirmation, which is associated with slightly shorter remaining conversation length, though the effect is not significant.

\subsection{User Satisfaction Across Different Models}
\label{app:results_user_satisfaction_across_different_models}

\begin{figure}[t]
    \centering
    \includegraphics[width=0.9\linewidth]{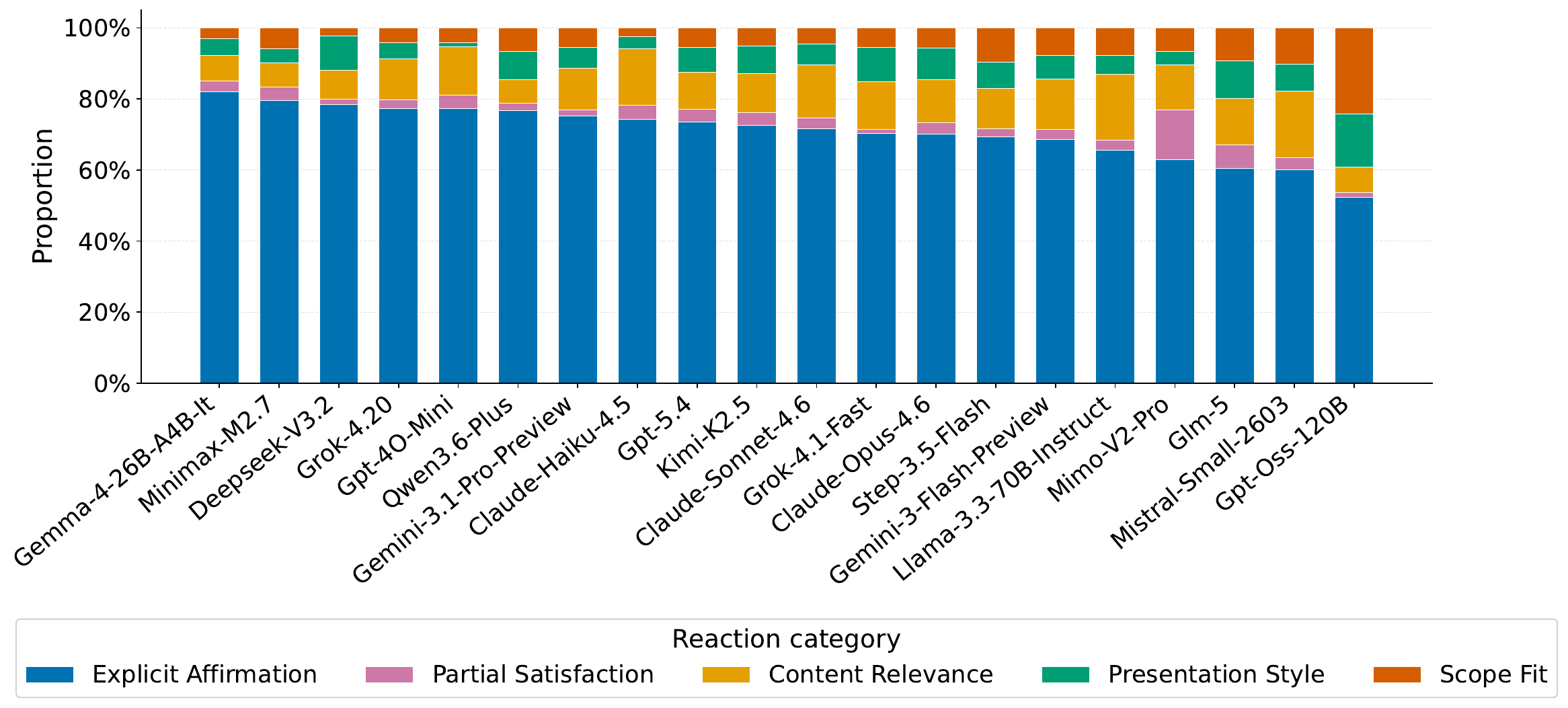}
    \caption{\textbf{Distribution of user reaction categories across language models.} Models are sorted by explicit affirmation rate (descending). Each bar represents the proportion of reactions falling into five categories for a given model, with sample sizes ($n$) shown above each bar.}
    \label{fig:reaction_distribution}
\end{figure}

We analyze user satisfaction across 20 language models by examining the distribution of
reaction categories assigned to model responses. Figure~\ref{fig:reaction_distribution}
presents these distributions, with models sorted in descending order by their explicit
affirmation rate.

Explicit affirmation is the dominant reaction category across all models, accounting for 55--82\% of reactions, indicating that the majority of user feedback reflects direct positive engagement with model outputs. Top-ranked models---including Gemma-4-26B-A4B-It and
Minimax-M2.7---achieve explicit affirmation rates above 80\%, while
lower-ranked models such as Gpt-Oss-120B fall closer to 55\%. Most notably, Gpt-Oss-120B stands out as having the highest proportion
of scope fit reactions among all evaluated models, suggesting a systematic tendency to misalign with the intended breadth or specificity of user requests. This pattern, absent in higher-ranked models, may reflect a fundamental limitation in how Gpt-Oss-120B
interprets task boundaries, and warrants closer investigation in future work. Content relevance is consistently the second-largest category across models.

%% file: appendices/3-collection.tex
\section{Details of Data Collection Methodology}
\label{app:method_details}

\subsection{User Consent}
\label{app:consent}

We recruit participants through Prolific and compensate them at an hourly rate above the applicable minimum wage. The sample consists of participants who self-report English as one of their fluent languages. Participation is voluntary and self-initiated. Institutional Review Board (IRB) approval was obtained prior to conducting the study.

Participants are redirected to our data collection platform, where they are informed of the study purpose (``investigate how people interact with AI chatbots''), told the study takes approximately 20 minutes, and asked to provide informed consent acknowledging voluntary participation, anonymity, and the right to withdraw. The full consent text is shown below.

\begin{AIbox}{User Consent}

By selecting the ``I Agree - Continue'' button below, you acknowledge that:

\begin{itemize}[leftmargin=20pt, nosep]
    \item You must be at least 18 years old to participate.
    \item You may decline to answer any or all of the following questions by closing this window in your browser.
    \item You may decline further participation, at any time, without adverse consequences.
    \item Your anonymity is assured; the researchers running this study will not receive any of your personal information.
\end{itemize}
\end{AIbox}

\subsection{Tutorial}
\label{app:tutorial}

Participants are then guided through a step-by-step tutorial on how to interact naturally with the AI chatbot and record contextually grounded thoughts. The tutorial uses plain language and demos of the chat interface to walk participants through each button and feature. The content of each tutorial page is shown below.

\begin{AIbox}{Tutorial}
\textbf{Page 1: Introduction} \\
You will now complete a task using an AI chatbot, a tool you can interact with to assist in your daily activities.

\dashedrule

\textbf{Page 2: Example Task} \\
Plan a trip for yourself. You can plan the trip based on your budget, time, and preference to anywhere in the world---a city, countryside, island, or any destination you choose. Be as creative and realistic as possible. By the end, you should have a complete itinerary.

\smallskip
\textit{Guidelines:}
\begin{itemize}[leftmargin=20pt, nosep]
    \item You will have 10 minutes to complete this task.
    \item When time is up, you'll automatically move on to the next step.
\end{itemize}

\dashedrule

\textbf{Page 3: Recording Your Thoughts} \\
Please chat as you normally would, but also write down your thoughts during the conversation. Your thoughts help us better understand what users think about the AI chatbot, and we will evaluate their quality. You can write down your thoughts at any time, even after the timeout.

\smallskip
\textit{Your thoughts should include:}
\begin{enumerate}[leftmargin=20pt, nosep]
    \item \textbf{For each AI response:} Your reactions to the response, including where and why you are satisfied or dissatisfied.
    \item \textbf{For each of your messages:} Your reasons for sending the message.
\end{enumerate}

\dashedrule

\textbf{Page 4: Chat Interface} \hfill\textit{\small[Demo]} \\
This is exactly what the chat interface will look like. Let's go through each component.

\dashedrule

\textbf{Page 5: Chat Area} \hfill\textit{\small[Demo]} \\
Please chat naturally here. Enter your message in the input box and click the ``Send'' button.

\dashedrule

\textbf{Page 6: ``+ Reasons'' Button} \hfill\textit{\small[Demo]} \\
Click this to provide your reasons for sending the message. It will be private and not visible to the AI.

\dashedrule

\textbf{Page 7: ``+ Reactions'' Button} \hfill\textit{\small[Demo]} \\
Click this to share your reactions to the AI's response. It will be private and not visible to the AI.

\dashedrule

\textbf{Page 8: ``New Chat'' Button} \hfill\textit{\small[Demo]} \\
You can click the ``New Chat'' button to start a new chat.

\dashedrule

\textbf{Page 9: ``Finish Task'' Button} \hfill\textit{\small[Demo]} \\
Once you're done with the task, click the ``Finish Task'' button.
\end{AIbox}

\subsection{Chat Interface}
\label{app:interface}

\begin{figure}[ht]
  \centering
  \includegraphics[width=0.95\linewidth]{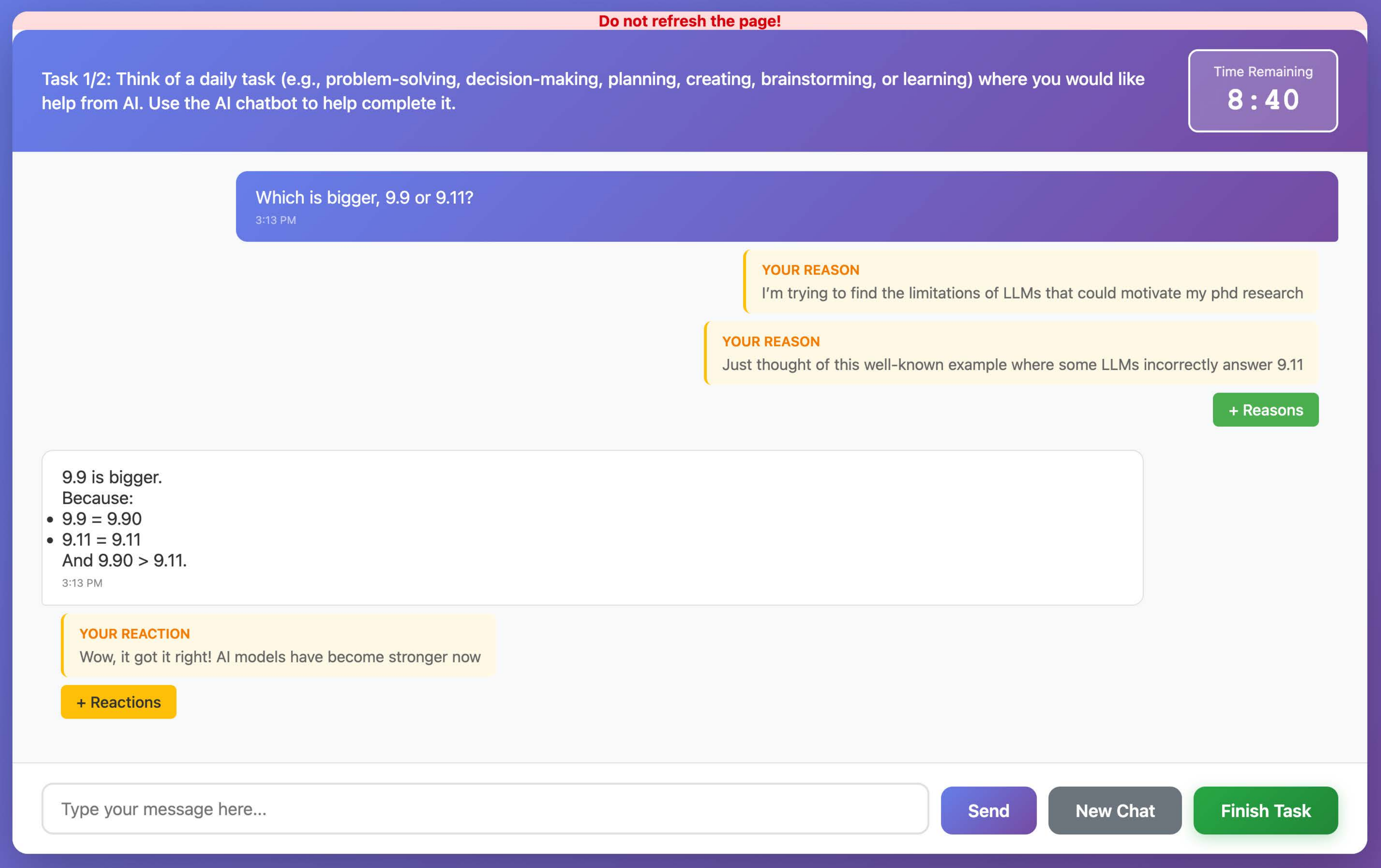}
  \caption{\textbf{Chat interface for collecting thought-annotated conversations.} The example shows a participant attaching two reasons to their prompt and a reaction to the assistant's response.}
  \label{fig:interface}
\end{figure}

The chat interface is a web application built with HTML, CSS, and JavaScript, backed by Firebase Firestore for real-time data persistence. A screenshot of the interface is shown in \autoref{fig:interface}.

\textbf{Instruction.} A gradient-styled header bar displays the task instruction: \textit{Think of a daily task (e.g., problem-solving, decision-making, planning, creating, brainstorming, or learning) where you would like help from AI. Use the AI chatbot to help complete it.}

\textbf{Timer.} To the right of the header bar, a countdown timer is initialized to 10:00 (600 seconds). The timer pulses with a yellow warning animation when one minute remains. When the timer reaches zero, the text input and send button are disabled, and the placeholder text changes to ``Time's up! Please finish the task.'' Participants may still annotate thoughts after timeout.

\textbf{Chat Area.} The chat area is the main scrollable region where the conversation is displayed. User messages appear right-aligned with a purple gradient background and white text, while assistant messages appear left-aligned with a white background and dark text. Assistant responses are rendered using the marked.js Markdown parser, supporting formatted output. Each message includes a timestamp.

\textbf{Thought Annotation System.} Below each message is a ``thought section'' containing:
\begin{itemize}[leftmargin=20pt, nosep]
    \item For \textbf{user messages}: a green \textbf{``+ Reasons''} button. Clicking it reveals a textarea with the placeholder ``Your reasons for sending this message...'' along with Save and Cancel buttons. Saved annotations appear as yellow-highlighted cards labeled ``your reason'' in orange uppercase text.
    \item For \textbf{assistant messages}: a yellow \textbf{``+ Reactions''} button. Clicking it reveals a textarea with the placeholder ``Your reactions to this response, where and why you are satisfied or dissatisfied...'' along with Save and Cancel buttons. Saved annotations appear as yellow-highlighted cards labeled ``your reaction''.
\end{itemize}

Multiple thoughts can be attached to a single message. Thoughts are private and not sent to the AI.

\textbf{Input Area.} At the bottom of the chat page, a row contains: (1) a resizable textarea for composing messages, supporting Enter-to-send (Shift+Enter for newlines); (2) a ``Send'' button; (3) a ``New Chat'' button that starts a fresh conversation thread while preserving previous threads in the data store; and (4) a ``Finish Task'' button to submit the current task.

\subsection{Post-Chat Surveys}
\label{app:survey}

\textbf{Task Survey.} After each task, participants answer the following two open-ended questions:
\begin{itemize}[leftmargin=20pt, nosep]
    \item What task did you just complete using the AI chatbot?
    \item In that task, what do you expect from the AI chatbot?
\end{itemize}

\textbf{Background Survey.} After both tasks, participants complete a demographic survey consisting of the following six questions:
\begin{itemize}[leftmargin=20pt, nosep]
    \item Age
    \item Gender (Male / Female / Non-binary / Prefer not to say)
    \item Education level (High school / Undergraduate / Graduate / Other)
    \item Occupation (free text)
    \item Frequency of AI chat usage, measured on a 5-point scale:
    \begin{itemize}[leftmargin=20pt, nosep]
        \item 1: Never
        \item 2: Used a couple of times, but not regularly
        \item 3: Once a week
        \item 4: Once a day
        \item 5: Many times a day
    \end{itemize}
    \item Main purposes for using AI (free text)
\end{itemize}
The results of the background survey are summarized in Figure~\ref{fig:demographic} and Section~\ref{sec:conversations}.

\subsection{Data Cleaning}
\label{app:data_cleaning}

After data collection, we retain most of our collected data to preserve its original characteristics, and remove only a very small portion in the three cases below:

\begin{itemize}[leftmargin=20pt, nosep]
    \item In very few cases, our platform automatically rejects participants who complete the task unusually quickly, indicating a lack of serious engagement.
    \item In very few cases, the chatbot does not respond or responds very slowly, while our system allows users to send multiple messages in the meantime. We remove part or all of such conversations when they contain consecutive user messages and result in strange, low-quality messages or thought annotations.
    \item In very few cases, we remove extremely low-quality conversations with no thought annotations and incomplete survey responses.
\end{itemize}

\subsection{Safeguards}
\label{app:safeguards}

\ours is released with several safeguards that mitigate the heightened misuse risk of cognitive self-report data. All conversations and annotations were collected under IRB-approved protocols with explicit informed consent, and participants were recruited through Prolific under guarantees of anonymity. No direct identifiers such as names, emails, or contact information appear in the dataset, and only coarse demographic attributes (age range, gender, occupation, education, and country-level geography) are retained for analysis. We distribute the dataset under a CC-BY-4.0 license intended for research use, and the accompanying dataset card explicitly designates as out-of-scope any attempt to re-identify participants, to build systems that exploit inferred mental states for manipulation or surveillance, or to treat the annotations as a complete record of underlying cognition rather than conscious in-the-moment self-reports. The card also documents known demographic biases and the reactivity inherent to thought elicitation, so that downstream users can apply \ours within its validated scope of studying latent user thoughts in multi-turn human-AI interaction.

\subsection{Limitations}
\label{app:limitations}

While \ours offers a unique window into the thoughts that accompany human-AI interactions, the very act of eliciting such thoughts imposes methodological constraints. We surface three limitations here and explain why each is inherent to in-situ thought collection rather than an artifact of our particular design:

\begin{itemize}[leftmargin=20pt, nosep]
    \item \textbf{Reactivity of thought externalization.} A well-established finding in cognitive science is that asking participants to report on anything beyond their primary task---even after the task is complete---can reshape the task itself. A participant who knows they will later annotate their thoughts may unconsciously adjust their interaction to make those annotations easier to produce: for example, polarizing their stated preferences or adopting cleaner intentions, since extreme or well-defined mental states are easier to articulate than ambiguous ones. This reactivity is fundamentally unavoidable whenever mental states are made explicit: any protocol that renders thoughts observable must also make the participant aware that they are being observed. We therefore interpret the collected annotations as \emph{thoughts-as-reported} rather than \emph{thoughts-as-occurred}, and we design the interface to minimize interruption and framing cues so that reactivity is reduced, though it cannot be eliminated.
    
    \item \textbf{Conscious versus subconscious cognition.} Externalized thoughts capture only those mental states that participants can consciously access and verbalize. Decades of work in psychology and behavioral science show that a substantial share of human behavior is shaped by subconscious processes, implicit associations, and automatic judgments that elude verbal report. As a consequence, \ours should be read as a record of users' \emph{explicit} reasoning about their interactions, not as a complete account of the cognitive processes driving them. We make this scope explicit in Thought Property 2, and we encourage downstream users of the dataset to treat annotations as a conscious overlay on, rather than a transcript of, the underlying cognition. We view this as a scoping decision rather than a deficiency: consciously articulated thoughts are themselves a signal that existing interaction datasets do not provide.
    
    \item \textbf{Recruited rather than fully in-the-wild participants.} Although our goal is to characterize thoughts during naturalistic human-AI interactions, participants are recruited through Prolific rather than drawn from unsolicited model traffic. This is a practical necessity: users of a public model/API service have no incentive to annotate their thoughts, and truly unsolicited thought collection would require invasive instrumentation that is neither ethical nor feasible at scale. Recruitment therefore introduces a modest selection effect. Reassuringly, however, our demographic analysis (Section~\ref{sec:conversations}) shows that \ours{} reflects a diverse spectrum of AI users and everyday use cases, consistent with the profile of frequent AI users in the real world, suggesting that the recruitment-induced distribution shift is small relative to the value of obtaining rich, consented thought annotations at scale.
\end{itemize}

%% file: appendices/4-analysis.tex
\section{Details of Analyses and Experiments}
\label{app:analysis_details}

\subsection{Conversation Property 1: \ours Captures a Representative Spectrum of Users}
\label{app:conv_property_1}

To characterize the participant pool behind \ours, we extract self-reported demographic and usage information from the post-task survey completed by each annotator alongside their conversations. For each conversation, we retain the first survey response and aggregate responses along six axes: \emph{age}, \emph{gender}, \emph{education}, \emph{occupation}, self-reported \emph{frequency} of LLM use, and free-text \emph{purposes} of use. 

Age is parsed as an integer and grouped into canonical brackets: 18--24, 25--34, 35--44, 45--54, 55--64, and 65+. Usage frequency is mapped from a 1--5 Likert scale to human-readable anchors ranging from ``Never'' to ``Many times a day.'' Gender and education are mapped to fixed category sets, including Male/Female/Non-binary for gender and High school/Undergraduate/Graduate/Other for education. For the two open-ended fields, \emph{occupation} and \emph{purposes}, we canonicalize responses by stripping whitespace and punctuation, lowercasing for deduplication, and re-casing labels for display. Purposes are further grouped into a small set of semantically coherent categories, including \emph{Learning}, \emph{Working}, \emph{Brainstorming}, \emph{Research}, etc., using keyword-based rules. Any unmatched responses are retained under their title-cased surface forms.

We compute counts for each group. Fixed-category axes are sorted by descending frequency with a deterministic tiebreaker, while open-ended axes are limited to the top eight entries whose display labels fit within a fixed character budget. The resulting statistics are rendered as a single six-panel horizontal bar chart, with one panel per demographic axis.

\subsection{Conversation Property 2: \ours Features Long-horizon Diverse Conversations}
\label{app:conv_property_2}

\textbf{Conversation and Message Lengths.} These analyses build on a shared message-level data frame produced by a helper that iterates over every conversation in the \ours dictionary, tags each message with its role (user or assistant), records its one-indexed turn position, and counts tokens with the tiktoken encoding for GPT-4o. The Conversation Length measured in Tokens (\ours vs. WildChat) cell aggregates this frame into per-conversation token totals for \ours, and obtains matching totals for WildChat by counting tokens across every message of WildChat-1M using the same tiktoken encoder, with a whitespace-based regex as a fallback. Both populations are then bucketed into fixed 1,000-token bins centered at 1k, 2k, ..., 15k (conversations above the cap fold into the last bin), converted to percentages of conversations per bin, and rendered as side-by-side bars.

The \textit{Conversation Length measured in Turns }(\ours vs. WildChat) analysis follows a parallel structure at the turn level. It derives the total turn count of each \ours conversation by taking the maximum turn position per conversation, and it obtains WildChat turn counts by enumerating the conversation field of every WildChat-1M row. Both populations are bucketed into even bins, normalized to percentages of their respective corpora, and drawn as side-by-side bars centered on the even integers. The x-axis is limited to [1, 25].

The \textit{Prompt and Response Lengths} analysis consumes the shared data frame directly and partitions token counts by role. It bins counts at a width of 200 tokens up to a cap of 4,000 and overlays two histograms on a single axis, with user prompts in blue and assistant responses in pink. Dashed vertical lines mark the per-role median token count; the y-axis uses a thousands formatter (e.g., ``1k'').

The \textit{Prompt and Response Length by Turn Position} analysis also reuses the shared frame, restricts it to turn positions 1 through 20, and takes the mean token count within each (role, turn position) cell. To respect the alternating structure of the dialogue, user means are retained only at odd positions, and assistant means are retained only at even positions. The two sequences appear as line plots with circular markers on a shared axis; the x-axis ticks span 1–20, and the y-axis reports average tokens per message.

\textbf{Conversation Topics.} We label conversation topics using an LLM (GPT-5.4) with a predefined topic taxonomy to assign all topics clearly present in each conversation, rather than forcing a single primary label. For each conversation, we concatenate the user and assistant turns into a single transcript and prompt the model with the full taxonomy and labeling instructions, requesting a JSON response containing the relevant taxonomy labels. The model is called with temperature 0 for deterministic outputs, and the returned labels are deduplicated and validated against the allowed taxonomy list via a cleaning step that discards any hallucinated or out-of-taxonomy labels.

This multi-label design allows a single conversation to be tagged with multiple topics when it spans several domains—for instance, a conversation touching on both programming and education receives both labels. After labeling, we aggregate topic counts across all conversations and organize them into a two-level hierarchy: topics are grouped into broader categories (e.g., ``Technology,'' ''Business \& Society,'' ``Arts \& Entertainment'') defined by a manual grouping, with any topics not covered by these predefined groups collected under ``Other Topics.'' This hierarchical structure is then visualized as a nested treemap, where the outer rectangles represent the high-level groups sized proportionally to their total counts, and inner rectangles represent individual topics sized by their frequency, providing an at-a-glance view of the topical distribution across the dataset.

We provide the topic labeling instructions for the LLM below.

\begin{AIbox}{Conversation Topic Labeling Instruction}
Label \textbf{all} topics that are clearly present in the content of the current conversation turn, or that are essential to completing the user's task. 
Many turns involve \textbf{multiple overlapping domains}, and \textbf{you should apply every relevant topic label}, not just the most obvious one. 
For example, a creative writing task about nature and exploring new lands should be labeled with ``Literature \& Writing'', ``Nature \& Environment'', and ``Travel \& Tourism''. 
Also, a query asking to build a website for a fashion business should be labeled with ``Fashion \& Beauty'', ``Business \& Finances'' \& ``Technology, Software \& Computing''. 
Include \textbf{all relevant topics}, especially when the content spans more than one thematic area. Avoid under-labeling.
\end{AIbox}

We provide the corresponding topic taxonomy below.

\begin{AIbox}{Conversation Topic Taxonomy}
\begin{itemize}[leftmargin=20pt, nosep]
\item \textbf{Adult \& Illicit Content:} Content involving mature or age-restricted themes, including topics related to sex, drugs, or alcohol. 
\item \textbf{Art \& Design:} Topics about visual art, creative techniques, aesthetics, or design principles. 
\item \textbf{Business \& Finances:} Topics involving any type of companies, markets, business practices, management, or personal financial planning.
\item \textbf{Culture:} Topics concerning traditions, customs, social norms, lifestyle identities, or cultural critique. 
\item \textbf{Economics:} Topics involving macro- or microeconomic systems, financial theory, trade, inflation, or economic policy. 
\item \textbf{Education:} Topics related to teaching, learning, academic subjects, school systems, or knowledge transmission. Do not multi-label with 'Technology, Software \& Computing' unless the task is about technical content (e.g., teaching code or explaining software). 
\item \textbf{Employment \& Hiring:} Topics about careers, job applications, resumes, hiring processes, or workplace dynamics.
\item \textbf{Entertainment, Hobbies \& Leisure:} Topics related to movies, television, music, games, crafts, or recreational activities.
\item \textbf{Fantasy / Fiction / Fanfiction:} Creative or fictional writing that includes imaginary worlds, character scenarios, or fan-authored extensions of media. 
\item \textbf{Fashion \& Beauty:} Topics about personal style, clothing trends, beauty products, grooming, or industry standards. 
\item \textbf{Food \& Dining:} Topics involving cuisines, recipes, cooking, restaurants, or dietary habits.
\item \textbf{Geography:} Topics about physical locations, world regions, maps, or geopolitical features. 
\item \textbf{Health \& Medicine:} Topics involving physical or mental health, medical knowledge, treatments, or wellbeing. 
\item \textbf{History:} Topics related to past events, timelines, historical figures, or historical analysis. 
\item \textbf{Housing:} Topics about real estate, renting, home ownership, architecture, or urban planning.
\item \textbf{Immigration / Migration:} Topics about cross-border movement, cultural adaptation, visas, or diaspora experiences.
\item \textbf{Insurance \& Social Scoring:} Topics related to insurance policies, risk models, or institutional scoring systems (e.g., credit/social scores).
\item \textbf{Interpersonal Relationships \& Communication:} Topics involving any form of signal where people interact with one another in personal, social, or professional settings. This includes: (1) Romantic, familial, platonic relationships, friendship, dating, breakups, intimacy, or emotional connection, (2) Communication advice (e.g., how to respond, how to ask, how to apologize), (3) Conflict resolution, emotional support, or interpersonal misunderstandings, (4) Social etiquette, small talk, expressing emotions or affection, tone-setting, or conversational timing. Label this even when the focus is not only when it's on the relationship itself, but on how to say something or respond in a conversation.
\item \textbf{Law, Criminal Justice, Law Enforcement:} Topics about legal systems, crime, law enforcement, policing, or justice procedures.
\item \textbf{Lifestyle:} Topics about routines, productivity, wellness, habits, or personal philosophies of daily living.
\item \textbf{Linguistics \& Languages:} Topics involving grammar, syntax, language families, translation, or linguistic theory.
\item \textbf{Literature \& Writing:} Topics about books, authorship, literary analysis, storytelling techniques, or writing practices.
\item \textbf{Math \& Sciences:} Topics about quantitative reasoning, scientific disciplines (e.g., math, physics, biology, chemistry), or scientific inquiry.
\item \textbf{Nature \& Environment:} Topics related to natural ecosystems, wildlife, conservation, weather, or environmental issues.
\item \textbf{News \& Current Affairs:} Topics referencing current or recent events, media coverage, or public opinion.
\item \textbf{Non-software Engineering \& Infrastructure:} Topics related to mechanical, civil, or structural engineering, and infrastructure systems (e.g., bridges, waterworks).
\item \textbf{Politics \& Elections:} Topics about political ideologies, government systems, elections, politicians, or civic processes.
\item \textbf{Psychology, Philosophy \& Human Behavior:} Topics involving cognitive science, behavioral analysis, mental health, philosophical reasoning, or ethics.
\item \textbf{Religion \& Spirituality:} Topics involving religious traditions, beliefs, practices, spiritual experiences, or theology.
\item \textbf{Social Issues \& Movements:} Topics involving inequality, advocacy, discrimination, social justice, or civil movements.
\item \textbf{Sports:} Topics about teams, players, sporting events, fitness, or athletics.
\item \textbf{Technology, Software \& Computing:} Topics involving computers, software, hardware, internet systems, or technological innovation. This includes: Programming languages, code generation, debugging, or algorithms, Software tools (e.g., Excel, Photoshop), platforms, or operating systems, Artificial intelligence, machine learning, or large language models, Consumer tech (e.g., smartphones, laptops, smart devices), Networking, cybersecurity, web development, or IT systems, Tech industry news, product comparisons, or emerging technologies.
\item \textbf{Transportation:} Topics involving travel modes, infrastructure, public transportation, or logistics.
\item \textbf{Travel \& Tourism:} Topics about travel destinations, itineraries, tips, or exploration.
\item \textbf{Video Games:} Topics involving game design, gameplay, genres, or gaming culture.
\item \textbf{Other:} Any topic not fitting the categories above.
\end{itemize} 
\end{AIbox}

\subsection{Conversation Property 3: \ours Conversations are Dominated by Task Extension}
\label{app:conv_property_3}

We analyze conversational structure by using an LLM (GPT-5.4) to label the relationship between each user turn and the immediately preceding user turn. For each conversation, we extract all user turns in order; the first turn is automatically labeled as ``First request,'' and for every subsequent turn, we prompt the model with both the previous and current user prompts, asking it to classify their relationship using a predefined taxonomy. The model is called with temperature 0 for deterministic outputs and returns a single JSON label, which is then normalized against the allowed taxonomy via a cleaning function that uses case-insensitive matching and keyword-based fallback rules to handle minor variations in the model's output.

This turn-level labeling assigns each user message one of four relationship types relative to its predecessor: (1) Extend, deepen, or build on the prior task, (2) Re-attempt or revise the prior task, (3) New variation of the prior task, or (4) Completely new request. The resulting sequence of relationship labels is stored both at the conversation level and attached directly to each individual user message, enabling fine-grained analysis of how users navigate within a conversation, whether they primarily continue and elaborate on a task, revise their prior attempt, explore variations, or shift to an entirely different request. This captures the structural dynamics of multi-turn interactions beyond what topic labels alone reveal.

We provide the labeling instructions for the LLM below.

\begin{AIbox}{Message Multi-turn Relationship Labeling Instruction}
Label one type of relationship that is present between the previous user prompt and the current user prompt in the conversation, using \textbf{exactly one option} from the list below.
\end{AIbox}

We provide the corresponding multi-turn relationship taxonomy below.

\begin{AIbox}{Message Multi-turn Relationship Taxonomy}
\begin{itemize}[leftmargin=20pt, nosep]
\item \textbf{First request:} The initial input or prompt provided by the user to start a new conversation thread. 
\item \textbf{Completely new request:} A user input that shifts to a different topic or context, unrelated to the previous turns. 
\item \textbf{Re-attempt/revision on prior task:} A follow-up input where the user revisits or retries a previously attempted task, possibly from earlier in the previous conversation. This often includes revised phrasing, clarification, or corrections after perceived issues, dissatisfaction, or misunderstanding. 
\item \textbf{New variation of prior task:} A follow-up input that is a related task to the prior turn, but explores a different angle or formulation of a previously attempted task. 
\item \textbf{Extend, deepen, or build on prior task:} A continuation that elaborates on, adds complexity to, or builds logically from the prior turn, indicating that the previous response was useful but not complete.
\end{itemize} 
\end{AIbox}

\subsection{Thought Property 1: Thoughts Are Different from Messages}
\label{app:thought_property_1}

To quantify how much of a user's underlying thinking is already reflected in their visible utterance, we measure the semantic coverage between user messages and their associated thoughts. For each eligible user turn, we extract the user's reason (their stated motivation for sending the current message) and the user's reaction (their response to the previous assistant message). We then use an LLM (GPT-5.4) to score how well the user's utterance conveys each type of thought on a 1-to-5 scale, where 1 indicates no meaningful overlap and 5 indicates full coverage. The scoring follows a structured rubric: the model receives the utterance and the thought as input and returns a single integer representing the degree of semantic overlap.

This analysis reveals the extent to which thoughts provide information that differs from what users explicitly express in their messages. A low average coverage score suggests that the thoughts capture latent user intent and reactions that are largely missing from the surface-level utterance, supporting the claim that thoughts constitute a meaningfully distinct signal from the conversation text alone. By evaluating reason coverage and reaction coverage separately, we can further distinguish whether users tend to omit their motivations for a new request versus their evaluative responses to prior assistant outputs, offering a more nuanced understanding of where the gap between utterances and internal reasoning is most pronounced.

We provide the prompt used to score how well the user's utterance conveys their thought below.

\begin{AIbox}{Semantic Coverage Evaluation Prompt}
\textbf{System Prompt:} \\
You are a semantic coverage evaluator. Given a user's utterance and their underlying thought (their reactions to previous assistant responses and/or their reasons for sending the message), score how well the utterance expresses the thought's semantic content.

\vspace{4pt}
Scoring rubric:
\begin{itemize}[leftmargin=20pt, nosep]
\item \textbf{1} -- No meaningful overlap; the utterance misses or contradicts the thought
\item \textbf{2} -- Minimal overlap; only a vague or incidental connection
\item \textbf{3} -- Partial coverage; the utterance captures some but not the core of the thought
\item \textbf{4} -- Good coverage; most of the thought is conveyed, with minor gaps
\item \textbf{5} -- Full coverage; the utterance fully and accurately expresses the thought
\end{itemize}

\vspace{4pt}
Output only a single integer (1--5). No explanation.

\dashedrule

\textbf{User Prompt:}\\
Utterance: \{user\_message\}\\
Thought: \{thought\}

\vspace{4pt}
Score:
\end{AIbox}

\subsection{Thought Property 2: Thoughts Are Difficult for LLMs to Infer}
\label{app:thought_property_2}

To assess how difficult it is to recover user thoughts from surface dialogue context, we prompt three frontier models---GPT-5.4, Gemini 3.1 Pro Preview, and Claude Opus 4.6---to infer two types of thoughts: (1) the user's \emph{reason} for sending their most recent message and (2) the user's \emph{reaction} to the assistant's most recent response. These two subtasks are conditioned on different dialogue contexts. For \emph{reasons}, we provide the conversation history up to and including the target user turn. For \emph{reactions}, we provide the conversation history up to and including the assistant turn being reacted to, optionally followed by the subsequent user message when available, since this follow-up often provides the strongest signal about whether the assistant's response satisfied the user. Both prompts consist of a system message that specifies the predictor's role and constrains the output to a single sentence in the user's voice, followed by a user message containing the formatted dialogue context.

Each model prediction is compared against the corresponding human-written thought using an LLM-as-a-judge. To mitigate self-preference bias, we deliberately use a judge model different from the predictor: predictions from GPT-5.4 are judged by a random choice between the two non-OpenAI models, and predictions from each non-OpenAI model are judged by GPT-5.4. The judge follows a fixed five-point rubric, ranging from 1 for no meaningful overlap or contradiction to 5 for a full semantic match while ignoring surface wording. We parse the judge's response as an integer and clamp it to the range $[1, 5]$. All predictions and judgments are cached to disk, and we report per-model averages as well as the unweighted mean across the three predictors for each thought type.

This analysis tests a key assumption: if thoughts were simply recoverable from the observable conversation, they would add little value as annotations. A low average similarity score between predicted and actual thoughts suggests that even a capable language model, given full conversational context, cannot reliably reconstruct what users are actually thinking, whether that concerns their motivations for a request or their evaluative responses to assistant outputs. By evaluating reasons and reactions separately, we can further identify which type of thought is harder to infer, revealing where the gap between observable dialogue and latent user cognition is most pronounced. Together with the coverage analysis, these results demonstrate that thoughts constitute a genuinely novel signal that is both distinct from user utterances and difficult to recover from context.

We provide the prompt used to infer the users' reasons below.

\begin{AIbox}{Reason Inference Prompt}
\textbf{System Prompt:} \\
You are a user intent predictor. Given a conversation context, infer the user's reason for sending the most recent message. Express the reason from the user's point of view.
\vspace{4pt}
Output only a single sentence. No explanation.
\dashedrule
\textbf{User Prompt:}\\
Conversation context: \{context\}
\vspace{4pt}

Predicted reason:
\end{AIbox}

We provide the prompt used to infer the users' reactions below.

\begin{AIbox}{Reaction Prediction Prompt}
\textbf{System Prompt:} \\
You are a user reaction predictor. Given a conversation context, infer the user's reaction to the most recent assistant message---identifying where and why they are satisfied or dissatisfied. Express the reaction from the user's point of view.
\vspace{4pt}
Output only a single sentence. No explanation.
\dashedrule
\textbf{User Prompt:}\\
Conversation context: \{context\}
\vspace{4pt}

Predicted reaction:
\end{AIbox}

We provide the prompt used to evaluate the predicted reasons against the actual human-annotated reasons below.

\begin{AIbox}{Reason Semantic Similarity Evaluation Prompt}
\textbf{System Prompt:} \\
You are a semantic similarity evaluator. Given two reasons describing why a user sent a message in a conversation, score how well the predicted reason matches the actual reason in semantic meaning (ignore surface wording).
\vspace{4pt}
Scoring rubric:
\begin{itemize}[leftmargin=20pt, nosep]
\item \textbf{1} -- No meaningful overlap; the reasons disagree or contradict each other
\item \textbf{2} -- Minimal overlap; only a vague or incidental connection
\item \textbf{3} -- Partial match; the predicted reason captures some but not the core of the actual reason
\item \textbf{4} -- Good match; most of the actual reason is reflected, with minor gaps
\item \textbf{5} -- Full match; the predicted reason fully and accurately conveys the same meaning as the actual reason
\end{itemize}
\vspace{4pt}
Output only a single integer (1--5). No explanation.
\dashedrule
\textbf{User Prompt:}\\
Actual reason: \{actual\}\\
Predicted reason: \{predicted\}

\vspace{4pt}
Score:
\end{AIbox}

We provide the prompt used to evaluate the predicted reactions against the actual human-annotated reactions below.

\begin{AIbox}{Reaction Semantic Similarity Evaluation Prompt}
\textbf{System Prompt:} \\
You are a semantic similarity evaluator. Given two reactions describing a user's response to an assistant's message in a conversation, score how well the predicted reaction matches the actual reaction in semantic meaning (ignore surface wording).
\vspace{4pt}
Scoring rubric:
\begin{itemize}[leftmargin=20pt, nosep]
\item \textbf{1} -- No meaningful overlap; the reactions disagree or contradict each other
\item \textbf{2} -- Minimal overlap; only a vague or incidental connection
\item \textbf{3} -- Partial match; the predicted reaction captures some but not the core of the actual reaction
\item \textbf{4} -- Good match; most of the actual reaction is reflected, with minor gaps
\item \textbf{5} -- Full match; the predicted reaction fully and accurately conveys the same meaning as the actual reaction
\end{itemize}
\vspace{4pt}
Output only a single integer (1--5). No explanation.
\dashedrule
\textbf{User Prompt:}\\
Actual reaction: \{actual\}\\
Predicted reaction: \{predicted\}

\vspace{4pt}
Score:
\end{AIbox}

\subsection{Thought Property 3: Thoughts Are Diverse in Content}
\label{app:thought_property_3}

To categorize user thoughts into \emph{reasons} and \emph{reactions}, we use \texttt{GPT-5.4} to assign labels from a predefined taxonomy. The prompting setup is tailored to the distinct contextual nature of each thought type. For labeling \emph{reasons}, we provide the conversation history up to and including the current user message, followed by the target reason text to label. This context enables the model to interpret the underlying motivation for a user utterance in light of prior turns. We preserve the dialogue structure but do not include the full content of assistant responses. This design reflects that the user’s intent is primarily expressed through their own sequence of actions across turns, rather than the specific wording of assistant replies. By focusing on user-side signals while retaining conversational structure, the model is better guided to infer why a particular message was produced. 

In contrast, \emph{reactions} are inherently localized: the model receives only the single assistant response that the user reacts to, followed by the corresponding reaction text. This is because a reaction reflects the user’s immediate evaluation of a specific response, and is therefore primarily determined by the content and presentation of that response itself.

\begin{AIbox}{Reasons Labeling Prompt}
\textbf{System Prompt:} \\
You are a thought classification expert. Your job is to read a conversation between a human and an AI assistant, then classify the internal thought the human had during that conversation into exactly one of the categories below.

\vspace{4pt}
\textbf{Categories:}
\begin{itemize}[leftmargin=20pt, nosep]
\item \textbf{task\_motivation}: The user introduces the task and explains their underlying goal, need, or real-world motivation for initiating the conversation. \\
\textit{Example: "I chose this because I am preparing for USMLE and need a study plan."}

\item \textbf{task\_continuation}: The user engages in follow-up interaction to refine, expand, or probe deeper into the current task without changing the overall goal. \\
\textit{Example: "trying to conclude" / "to develop previous message."}

\item \textbf{task\_reorientation}: The user changes their objective or redirects the conversation toward a different task or outcome. \\
\textit{Example: "Changing plan." / "changing my search from common questions to totally differents one and check it is work on feed or do real sreach"}

\item \textbf{content\_expectation}: The user specifies what the response should achieve in terms of substance, such as level of detail, correctness, or type of information. \\
\textit{Example: "i want him to do a pretty general model so i can change a bit and be good to go"}

\item \textbf{style\_expectation}: The user defines how the response should be structured or presented, including format, tone, or organization. \\
\textit{Example: "I asked this to get a more detailed and practical plan that I can actually follow." / "I won't read so much"}

\item \textbf{context\_grounding\_and\_constraints}: The user provides situational context, personal preferences, or constraints that shape the solution space. \\
\textit{Example: "I wanted to tell the AI all that I needed to get done that is relevant to my day by including all of the important things I need to get done today."}

\item \textbf{social\_and\_others}: The user engages in interaction management or non-task-oriented communication such as greetings, acknowledgments, or meta-comments. \\
\textit{Example: "Hi!" / "Thanks, that helps." / "AI convince me"}
\end{itemize}

\vspace{4pt}
\textbf{Instructions:}
\begin{itemize}[leftmargin=20pt, nosep]
\item Analyze the given thought text carefully
\item Classify it into exactly one category
\item Choose the category that best represents the primary intent of the thought
\item If a thought fits multiple categories, select the most dominant/primary one
\item Be precise and consistent in your classifications
\end{itemize}

\vspace{4pt}
Return only the category label in lowercase with underscores. No explanation.
\dashedrule
\textbf{User Prompt:}\\
Conversation: \\
\texttt{Human message: \ldots} \\
\texttt{AI assistant response: \ldots} \\
\texttt{...} \\
\texttt{Human message: \ldots} \\
\texttt{Human thoughts for this message: \{reason\}}

\vspace{4pt}
Label:
\end{AIbox}

\begin{AIbox}{Reactions Labeling Prompt}
\textbf{System Prompt:} \\
You are a reaction classification expert. Your job is to read an AI assistant response and the internal reaction the human had after reading it, then classify that reaction into exactly one of the categories below.

\vspace{4pt}
\textbf{Categories:}
\begin{itemize}[leftmargin=20pt, nosep]
\item \textbf{explicit\_affirmation}: The user produces clear positive signals in text, expressing that the response fully met their expectations. \\
\textit{Example: "Perfect.", "That's exactly it."}

\item \textbf{partial\_satisfaction}: The user acknowledges the output positively but immediately requests a minor tweak. \\
\textit{Example: "This is great, just make it a bit shorter."}

\item \textbf{presentation\_style}: The user expresses dissatisfaction with how the response is delivered, including tone, wording, structure, or overall presentation. \\
\textit{Example: "The email is okay, but it sounds too serious and uses big words."}

\item \textbf{scope\_fit}: The user expresses dissatisfaction with how much or how broadly the response is provided, including its length, level of detail, and breadth. \\
\textit{Example: "Way too detailed.", "Too much words.", "Too long."}

\item \textbf{content\_relevance}: The user expresses dissatisfaction with what is included in the response, including missing information or irrelevant material. \\
\textit{Example: "The response is clearer now, but I would like suggestions for specific hotels within my budget."}
\end{itemize}

\vspace{4pt}
\textbf{Instructions:}
\begin{itemize}[leftmargin=20pt, nosep]
\item Analyze the given reaction text carefully
\item Classify it into exactly one category
\item Select the most dominant intent if multiple categories apply
\item Be precise and consistent
\end{itemize}

\vspace{4pt}
Output only the category label in lowercase with underscores. No explanation.
\dashedrule
\textbf{User Prompt:}\\
\texttt{AI assistant response: \{response\}} \\
\texttt{Human reaction to this AI response: \{reaction\}}

\vspace{4pt}
Label:
\end{AIbox}

\subsection{Thought Property 4: Thought Dynamics Depend on Conversation Stages}
\label{app:thought_property_4}

\textbf{Relationship to conversation stage}. To characterize how user thoughts evolve over the course of a conversation, we construct Sankey-style flow visualizations over four normalized dialogue stages: \textit{Early} (0--33\% of the conversation), \textit{Mid-Early} (33--67\%), \textit{Mid-Late} (67--100\%), and \textit{Late} (final segment). For each stage, annotated labels are aggregated and normalized to obtain category-level percentage distributions, where stacked vertical bars represent the relative frequency of each category at a given stage. To model transitions across stages, we estimate pairwise flows between categories in consecutive stages. For a source category $c_i$ at stage $t$, its outgoing mass is distributed across categories at stage $t+1$ proportionally to the target-stage category frequencies, yielding a dense transition matrix while preserving the total mass associated with each source category. The resulting flows are rendered as smooth ribbons connecting stacked segments across stages, enabling a compact visualization of temporal shifts in conversational patterns.

\textbf{Relationship to conversation's topic, message's multi-turn relationships, and conversation length.} To examine how thought types vary across conversational contexts, we constructed cross-tabulation heatmaps between thought labels and two categorical dimensions: conversation topic and multi-turn relationship type. For reason labels, we paired each labeled reason with both the conversation-level topic annotations and the message-level multi-turn relationship label assigned to that same user turn. For reaction labels, we paired each labeled reaction with the conversation-level topic and, crucially, with the multi-turn relationship label of the \textit{next} user message rather than the current one, capturing the forward-looking relationship between an assistant's response characteristics and the user's subsequent behavioral choice. All heatmaps were computed as normalized percentages. To support analysis at multiple granularities, we included a flag that optionally aggregates the 35 individual topic labels into 7 broader thematic groups (e.g., Technology, Business \& Society, Health \& Relationships), using a predefined mapping consistent with the topic hierarchy defined in \textit{Conversation Property 2}.

\textbf{Relationship to conversation length.} To assess how thought types relate to conversation structure, we computed two positional statistics for each thought label: total conversation length (number of messages in the conversation where the thought occurs) and remaining conversation length (number of messages after the current turn). These were collected separately for reason labels on user messages and reaction labels on assistant messages. The resulting distributions were visualized as paired box plots with overlaid mean markers, enabling comparison of both central tendency and spread across thought types. Sample sizes were annotated on each box to contextualize the statistical reliability of each category. This design reveals whether certain thought types tend to appear in longer or shorter conversations and whether they cluster toward the beginning or end of a conversational session.

\subsection{Thought Utility 1: Thoughts Predict User Behavior}
\label{app:thought_utility_1}

This section details the next-message prediction experiment in Section~\ref{sec:prediction}: dataset filtering, prediction prompts for the history-only and thought-augmented conditions, and the semantic similarity scoring protocol.

To evaluate whether thought annotations improve the ability to anticipate user behavior, we conduct a next-message prediction experiment. For each assistant message followed by a user turn, we construct two versions of the conversation context: a \textit{history-only} version containing only the raw dialogue history, and a \textit{thought-augmented} version that interleaves the user's annotated reasons and reactions at the appropriate turns. We restrict the evaluation to examples whose thought annotations are high-quality, i.e., substantive and informative about the user's latent intent or attitude beyond what is already evident in the conversation surface. Concretely, we use an LLM judge to rate every thought annotation on a 1--5 quality scale and keep only examples scored $\geq 4$, ensuring that the comparison reflects the value of genuinely informative thoughts rather than boilerplate filler. We then prompt three LLM predictors, GPT-5.4, Gemini 3.1 Pro Preview, and Claude Opus 4.6, to predict the user's next message under each condition independently. We assign each prediction a semantic similarity score in $[0, 100]$ relative to the actual next user message, using an LLM judge with the prompt shown below. To avoid self-evaluation bias, each predictor's outputs are scored by a judge sampled uniformly at random from the other two models.

This setup directly quantifies the predictive utility of thought annotations: a higher thought-augmented similarity relative to the history-only baseline indicates that knowing what users are thinking provides an actionable signal for anticipating their subsequent messages beyond what the conversation surface alone reveals. Across all three predictor models, thought-augmented prediction consistently outperforms history-only prediction, suggesting that thoughts capture information that is novel, hard to recover, and practically valuable for modeling user behavior in multi-turn conversations.

We provide the prompt used to predict the next message with context only below.

\begin{AIbox}{Next-Message Prediction Prompt (Context Only)}
\textbf{System Prompt:} \\
You are a next-message predictor. Given a conversation context, predict the user's next message.
\vspace{4pt}

Output only a single likely next user message. No explanation.
\dashedrule
\textbf{User Prompt:}\\
Conversation context: \{context\}
\vspace{4pt}

Predicted next message:
\end{AIbox}

We provide the prompt used to predict the next message with context and thoughts below.

\begin{AIbox}{Next-Message Prediction Prompt (Context + Thoughts)}
\textbf{System Prompt:} \\
You are a next-message predictor. Given a conversation context that includes thought annotations, predict the user's next message.
\vspace{4pt}

Output only a single likely next user message. No explanation.
\dashedrule
\textbf{User Prompt:}\\
Conversation context with thoughts: \{context\_with\_thoughts\}
\vspace{4pt}

Predicted next message:
\end{AIbox}

We provide the prompt used to score the semantic similarity between a predicted next message and the actual next message below.

\begin{AIbox}{Next-Message Semantic Similarity Prompt}
\textbf{System Prompt:} \\
You are a semantic similarity evaluator. Given an actual user message and a predicted next user message, score how well the prediction matches the actual message in semantic meaning (ignore surface wording, length, and style).
\vspace{4pt}

Scoring rubric (continuous scale):\\
0 - No meaningful overlap; the prediction disagrees with or is unrelated to the actual message\\
25 - Minimal overlap; only a vague or incidental connection\\
50 - Partial match; captures some but not the core of the actual message\\
75 - Good match; most of the actual message is reflected, with minor gaps\\
100 - Full match; fully and accurately conveys the same meaning as the actual message
\vspace{4pt}

Output only a single number between 0 and 100. No explanation.
\dashedrule
\textbf{User Prompt:}\\
Actual next message: \{actual\}\\
Predicted next message: \{predicted\}
\vspace{4pt}

Score:
\end{AIbox}

\subsection{Thought Utility 2: Thoughts Improve Model Alignment}
\label{app:thought_utility_2}

This section details the alignment experiments in Section~\ref{sec:alignment}: training data construction for thought-guided and message-guided rewrites, the rewrite prompts, and training and evaluation setups.

\textbf{Training data.} We generate the training data for thought-guided rewrites as follows:
\begin{enumerate}[leftmargin=20pt, nosep]
    \item \textbf{Load and filter conversations:} We load the dataset and retain only conversations with 2--20 turns.
    \item \textbf{Collect dissatisfaction reactions:} We scan all user reactions labeled as ``content relevance'', ``presentation style'', or ``scope fit'', the three dissatisfaction types defined in \ours. Each reaction's text serves as the ``thought'' that guides the rewrite.
    \item \textbf{Filter to meaningful thoughts:} We discard thoughts that are empty, shorter than six words, or contain no alphabetic characters, ensuring the rewriter has sufficient signal to act on.
    \item \textbf{Build multi-turn context:} For each remaining candidate, we slice the conversation up to (but not including) the dissatisfying assistant response, yielding a \{role, content\} message list that ends with the triggering user prompt.
    \item \textbf{Generate thought-guided rewrites:} We prompt GPT-5.4 with the context, the original response, the dissatisfaction label and its description, and the user's thought, requesting a revised assistant response that addresses the complaint.
    \item \textbf{Save as DPO pairs:} We store the training data in the standard DPO schema: \textit{prompt} (the multi-turn context up through the triggering user message), \textit{chosen} (the thought-guided rewrite), and \textit{rejected} (the unsatisfactory assistant response from the original dataset).
\end{enumerate}

We generate the training data for message-guided rewrites as follows:
\begin{enumerate}[leftmargin=20pt, nosep]
    \item \textbf{Load and filter conversations:} We load the dataset and retain only conversations with 2--20 turns.
    \item \textbf{LLM-classify dissatisfaction:} We prompt GPT-5.4 with each (assistant response, user followup) pair and ask it to output exactly dissatisfied or satisfied. Each reaction's text serves as the ``thought'' that guides the rewrite.
    \item \textbf{Filter to meaningful messages:} We discard messages that are empty, shorter than six words, or contain no alphabetic characters, ensuring the rewriter has sufficient signal to act on.
    \item \textbf{Build multi-turn context:} For each remaining candidate, we slice the conversation up to (but not including) the dissatisfying assistant response, yielding a \{role, content\} message list that ends with the triggering user prompt.
    \item \textbf{Generate message-guided rewrites}: We prompt GPT-5.4 with the context, original response, and the user's follow-up message, asking for a revised response that preemptively addresses the follow-up so the user wouldn't have needed to push back.
    \item \textbf{Save as DPO pairs:} We store the training data in the standard DPO schema: \textit{prompt} (the multi-turn context up through the triggering user message), \textit{chosen} (the message-guided rewrite), and \textit{rejected} (the unsatisfactory assistant response from the original dataset).
\end{enumerate}

The training data sizes for the three training runs are:
\begin{enumerate}[leftmargin=20pt, nosep]
    \item 1,000 instances using thought-guided rewrites on \ours, derived from 1,985 conversations (90\% of all \ours conversations).
    \item 450 instances using message-guided rewrites on \ours, derived from the same 1,985 conversations as (1). The smaller size is intentional: it ensures a fair comparison on identical conversations and supports our claim that thoughts surface more dissatisfaction instances than messages.
    \item 1,000 instances using message-guided rewrites on WildChat, derived from 4,669 conversations. We process WildChat conversations in random order until we obtain 1,000 filtered instances, matching the size in (1).
\end{enumerate}

\textbf{Prompt Used.} We provide the prompts used to generate the thought-guided and message-guided rewrites below.

\begin{AIbox}{Thought-Guided Rewrite Prompt}
\textbf{System Prompt:} \\
You are an expert assistant that rewrites prior assistant responses so they better satisfy the user. You will be given the multi-turn conversation so far, the original assistant response the user was dissatisfied with, and the user's internal ``thought'' explaining why they were dissatisfied. Your job is to produce a single revised assistant response that directly addresses the user's feedback. Return ONLY the revised response text, with no preambles, no explanations, no meta-commentary, and no markdown fences.
\dashedrule
\textbf{User Prompt:}\\
Conversation context: \{context\}
\vspace{4pt}

Original assistant response (the user was dissatisfied with this): \{assistant\_response\}
\vspace{4pt}

User's dissatisfaction label: \{label\} (\{label\_description\})
\vspace{4pt}

User's internal thought about why they were dissatisfied: \{thought\}
\end{AIbox}

\begin{AIbox}{Message-Guided Rewrite Prompt}
\textbf{System Prompt:} \\
You are an expert assistant that rewrites prior assistant responses so they better satisfy the user. You will be given the multi-turn conversation so far, the original assistant response the user was dissatisfied with, and the user's next message which expresses (explicitly or implicitly) what they actually wanted. Your job is to produce a single revised assistant response, issued at the same point as the original assistant response, that proactively addresses the concerns surfaced by the user's next message, so that the user would not have needed to push back. Return ONLY the revised response text, with no preambles, no explanations, no meta-commentary, and no markdown fences.
\dashedrule
\textbf{User Prompt:}\\
Conversation context: \{context\}
\vspace{4pt}

Original assistant response (the user was dissatisfied with this): \{assistant\_response\}
\vspace{4pt}

User's next message (revealing what they actually wanted / why they were unhappy): \{user\_followup\}
\end{AIbox}

\textbf{Training details.} We initialize all models from Qwen3.5-4B \citep{yang2025qwen3}. We conduct Direct Preference Optimization (DPO) training using the Tinker APIs. Across all three experiments, we use a batch size of 64, a learning rate of $1 \times 10^{-6}$, and train for up to 20 epochs with early stopping based on a 10\% validation split.

\textbf{Evaluation details.} Models are evaluated on Arena-Hard \citep{li2024crowdsourced}, a robust instruction following benchmark that has a 98.6\% correlation with human preference. Evaluations are conducted using GPT-4o as the judge (the original benchmark used GPT-4 Turbo, which has since been deprecated). We report both raw and style-controlled (SC) win rates.